\documentclass{article}

\usepackage{PRIMEarxiv}
\usepackage[square, numbers]{natbib}
\usepackage{subfigure}
\usepackage{multirow}
\usepackage{multicol}
\usepackage{amsmath}
\usepackage{todonotes}
\usepackage{pifont}
\usepackage[utf8]{inputenc} 
\usepackage[T1]{fontenc}    
\usepackage{xcolor}         
\usepackage[ruled,linesnumbered]{algorithm2e}
\newtheorem{theorem}{Definition}
\usepackage{wrapfig}
\usepackage{hyperref}       
\usepackage{url}            
\usepackage{booktabs}       
\usepackage{amsfonts}       
\usepackage{nicefrac}       
\usepackage{microtype}      
\usepackage{lipsum}
\usepackage{fancyhdr}       
\usepackage{graphicx}       
\newcommand{\ours}{\texttt{CC-MIA}\xspace}
\newcommand{\oursLong}{\texttt{Cross-Client Membership Inference Attack}\xspace}
\graphicspath{{media/}}     

\title{Who Owns This Sample: Cross-Client Membership Inference Attack in Federated Graph Neural Networks}
\author{
  Kunhao Li$^{\dagger}$ \\
  South China University of Technology \\
  Guang Zhou \\
  \And
  Di Wu$^{\dagger}$ \\
  University of Southern Queensland \\
  Queensland \\
  \AND
  Jun Bai \\
  McGill University \\
  Montreal\\
  \And
  Jing Xu \\
  CISPA Helmholtz Center for Information Security \\
  Saarbrücken\\
  \And
  Lei Yang \\
  South China University of Technology \\
  Guang Zhou\\
  \And
  Ziyi Zhang \\
  South China University of Technology \\
  Guang Zhou\\
  \And
  Yiliao Song \\
  University of Adelaide \\
  Adelaide\\
  \And
  Wencheng Yang \\
  University of Southern Queensland \\
  Queensland \\
  \And
  Taotao Cai \\
  University of Southern Queensland \\
  Queensland \\
  \And
  Yan Li \\
  University of Southern Queensland \\
  Queensland \\
}

\begin{document}
\maketitle

\begingroup
\renewcommand\thefootnote{$^{\dagger}$}
\footnotetext{These authors contributed equally to this work.}
\endgroup

\begin{abstract}
Graph-structured data is prevalent in many real-world applications, including social networks, financial systems, and molecular biology. Graph Neural Networks (GNNs) have become the de facto standard for learning from such data due to their strong representation capabilities. As GNNs are increasingly deployed in federated learning (FL) settings to preserve data locality and privacy, new privacy threats arise from the interaction between graph structures and decentralized training. 
In this paper, we present the first systematic study of cross-client membership inference attacks (\ours) against node classification tasks of federated GNNs (FedGNNs), where a malicious client aims to infer which client owns the given data. Unlike prior centralized-focused work that focuses on whether a sample was included in training, our attack targets sample-to-client attribution, a finer-grained privacy risk unique to federated settings. We design a general attack framework that exploits FedGNNs' aggregation behaviors, gradient updates, and embedding proximity to link samples to their source clients across training rounds.
We evaluate our attack across multiple graph datasets under realistic FL setups. Results show that our method achieves high performance on both membership inference and ownership identification. Our findings highlight a new privacy threat in federated graph learning—client identity leakage through structural and model-level cues, motivating the need for attribution-robust GNN design.
\end{abstract}


\section{Introduction}
\label{introduction}
Graphs are a fundamental data structure used to model pairwise relationships between entities, and they naturally arise in a wide range of real-world domains such as social networks, citation graphs, molecular structures, recommendation systems, and financial transaction networks. To effectively learn from graph-structured data, Graph Neural Networks (GNNs) have emerged as a powerful class of models that extend deep learning to non-Euclidean domains \cite{wu2020comprehensive}. Despite their effectiveness, the training of GNNs often involves sensitive data, such as user profiles, social interactions, or molecular properties, raising significant privacy concerns. It becomes crucial to understand and mitigate their vulnerabilities to inference attacks and other forms of information leakage \cite{sajadmanesh2021locally}.

Existing privacy attacks in GNNs have primarily focused on models trained in centralized settings, where a single server has access to the entire graph. Under this paradigm, adversaries have demonstrated the feasibility of various inference attacks, such as membership inference \cite{wu2021adapting}, property inference \cite{he2020roadtagger}, and link reconstruction \cite{zhang2022inference}, aiming to extract sensitive node or edge information from the trained model. These attacks exploit the relational inductive bias of GNNs, where node representations are heavily influenced by their neighbors, making it possible to infer private attributes or determine whether specific nodes or connections were part of the training data. 

To mitigate the privacy risks in graph representation learning, Federated Learning (FL), as a proven distributed training scheme, has been widely used in different areas \cite{bai2025non}. It has also been integrated into GNNs, resulting in Federated Graph Neural Networks (FedGNNs) that support decentralized training over sensitive graph data. \cite{liu2024federated, schmierer2025advancing}. However, despite its privacy-preserving designs, FL itself remains vulnerable to emerging threats \cite{rao2024privacy}, such as poisoning \cite{zhao2020pdgan}, backdoor \cite{zhang2021badfss}, membership inference \cite{xie2021defending}, and model inversion attacks \cite{wu2024fedinverse}.

This distributed paradigm fundamentally shifts the adversary’s perspective compared to centralized settings. In FedGNNs, no single party has full access to the entire graph; instead, attackers can only observe aggregated updates or gradients. Nevertheless, studies have demonstrated that even these limited observations pose significant risks \cite{bai2024membership, liu2025piafgnn}. For instance, gradients can be exploited to infer training examples or determine the presence of specific data points. These unique vulnerabilities underscore the need to systematically address novel attack surfaces that arise in FedGNNs.

To this end, we introduce \textbf{\ours}, the pioneer membership inference attack framework targeting the node classification task of FedGNNs, which is designed to enable both cross-client membership inference on training datasets and client identification for specific nodes. \ours shows that publicly available graph datasets with similar structures can be used as shadow datasets to launch membership inference attacks on other clients, with a maximum improvement of 72.16\%. Furthermore, \ours can reverse local graph features through eavesdropping on uploaded client gradients that acquire the ability to identify the ownership of target data across clients via prototype matching. 

Our study offers the following contributions:
\ding{182} We initiate the study of node-level membership inference attacks in FedGNNs by positioning the attacker as one of the participating clients. Leveraging only malicious client data and public structurally similar shadow datasets, our framework allows inferring whether specific nodes from other clients were part of their training datasets.
\ding{183} We investigate the challenges of determining data ownership identification of clients in FedGNNs. Our approach shows that an attacker, relying solely on eavesdropped gradients, can reconstruct client subgraph data and leverage prototype construction to infer data ownership.
\ding{184} Our comprehensive experiments validate the effectiveness of \ours from both membership inference perspectives. The results indicate that \ours significantly surpasses centralized MIA methods within the federated learning context, providing valuable insights into potential defense strategies for FedGNNs.

\section{Related Works}
\label{related_works}

Federated learning (FL), despite its privacy-preserving design, remains vulnerable to membership inference attacks (MIAs), which aim to determine whether specific data samples were used during training \cite{nasr2019comprehensive}. These attacks pose serious privacy risks—such as revealing medical diagnoses—and can support compliance auditing or serve as precursors to more advanced threats like model extraction \cite{zhu2019deep, melis2019exploiting, wang2019beyond}. In FL, MIAs have evolved from gradient-based approaches to more advanced methods involving shadow training \cite{zhang2022label}, hyperparameter stealing \cite{li2022auditing}, and feature-based inference \cite{liu2023subject, yan2022membership}. Recent works have extended MIAs to graph neural networks (GNNs), including node-level attacks using posterior distributions \cite{he2021node}, label-only attacks \cite{conti2022label}, and subgraph- or graph-level inference \cite{olatunji2021membership, wu2021adapting, zhang2022inference, liu2022membership}. Detailed related works are stated in Appendix~\ref{detailed_related_works}.

However, FedGNNs introduce new and largely unexplored privacy risks. Most GNN-specific attacks are limited to centralized settings with full graph access, while in FedGNNs, clients hold disjoint and heterogeneous subgraphs, and their relationships are implicitly captured through shared model updates, making traditional attacks less effective. Additionally, no prior work has explored cross-client MIAs, where a malicious client aims to infer not just whether a node was used in training, but also which client it belongs to. We identify this overlooked threat and present the first systematic study of membership and ownership inference in FedGNNs from a client-side adversary perspective.


\section{Problem Definition}
\label{problem_definition}

\subsection{Threat Model}
Following existing client-side attacks in federated learning~\cite{2023feddefender, 2023watermarking, 2024client, xu2024robust}, we consider a scenario where one of the participating clients acts as an attacker. This adversarial client has access to its own local graph data, denoted as $G_{a}(X_{a}, A_{a})$, where $X_{a}$ represents the node features and $A_{a}$ denotes the adjacency of the attacker's subgraph.

The attacker is capable of obtaining the following information during the federated learning process:

\textbf{\ding{182} Global Model Updates}: The attacker has access to the global model parameters $\Theta$ after each communication round~\cite{bagdasaryan2020backdoor, bhagoji2019analyzing}.
\textbf{\ding{183} Subgraph Dataset Category}: Following~\cite{2021nodemia, 2021gnnmia}, the attacker is assumed to be aware of the subgraph dataset's category (\textit{e.g.}, citation, product, social, \textit{etc.}). This enables the attacker to match a shadow dataset $\tilde{G}(\tilde{X}, \tilde{A})$ that closely approximates the data distribution of the target category.
\textbf{\ding{184} Gradient Eavesdropping}: Based on~\cite{wu2025secret, hu2021federated}, the attacker possesses eavesdropping capabilities, allowing it to intercept the gradient updates $\mathcal{G} = \{\mathcal{G}_1, \cdots, \mathcal{G}_K\}$ uploaded by other clients to the server.
\textbf{\ding{185} Why Upload Gradients?}~Under our threat model, uploading gradients to the server for aggregation is safer than directly uploading local updated parameters. If an attacker can eavesdrop on the upload link of clients, exposing model parameters would enable the attacker to infer the posterior probabilities of the local GNN directly~\cite{2021gnnmia}. Such exposure significantly increases the risk of MIAs on the training-set. 
\subsection{Attack Taxonomy}
Building upon the threat model assumptions, we concentrate on training-set membership inference and client-data identification within federated GNN settings and provide formal definitions.

\begin{theorem}[Membership Inference Attack]
\label{def_training_set_mia}
Let $G(X,A)$ be the target graph with nodes $X = \{x_1,\dots,x_N\}$ and let $\mathcal{F}_\Theta$ denote the global GNN embedding function parametrized by $\Theta$. Suppose an attacker possesses a shadow dataset $\tilde G(\tilde X,\tilde A)$ drawn from the same category as $G$.  By extracting embeddings 
\begin{equation}
    \tilde{E} = \mathcal{F}_\Theta(\tilde{X},\tilde{A})\,,
\end{equation}
the attacker trains a binary classifier $f_w\colon \mathbb{R}^D \to \{0,1\}$ with parameters $w$ on $(\tilde E,\tilde m)$, where $\tilde m_i=1$ if $\tilde x_i$ is a training member and $\tilde m_i=0$ otherwise. Then for any node $x_i \in X$, the membership inference attack is realized by
\begin{equation}
    f_w\bigl(\mathcal{F}_\Theta(x_i)\bigr)
    \;\approx\;
    \begin{cases}
    1, & \text{if } x_i \text{ is used in training},\\
    0, & \text{otherwise}.
    \end{cases}
\end{equation}
    
\end{theorem}

\begin{theorem}[Client-data Identification]
\label{def_client_ownership_mia}
Let $\mathcal{G} = \{\mathcal{G}_1, \cdots, \mathcal{G}_K\}$ denote the gradients uploaded by $K$ participating clients during each communication round in a federated GNN framework. Suppose an attacker eavesdrops on $\mathcal{G}$ and employs a reconstruction function $f_{rec}(\cdot)$ to approximate the feature representations $r_k = f_{rec}(\mathcal{G}_k)$ of the subgraphs held by client $k \in \{1, \cdots, K\}$. For a target node $x_i \in X = \{x_1, \cdots, x_N\}$, the attacker infers the ownership by determining the client $k$ that minimizes a similarity measure $\mathcal{D}(\cdot)$. Formally, the ownership inference is given by:
\begin{equation}
    k^* = \arg \min_{k \in \{1, \cdots, K\}} \mathcal{D}(x_i, r_k),
\end{equation}
where $\mathcal{D}(\cdot)$ quantifies a similarity function.
\end{theorem}

\section{Methodology}
\label{methodology}
\begin{figure*}
  \centering
        \includegraphics[scale=0.65]{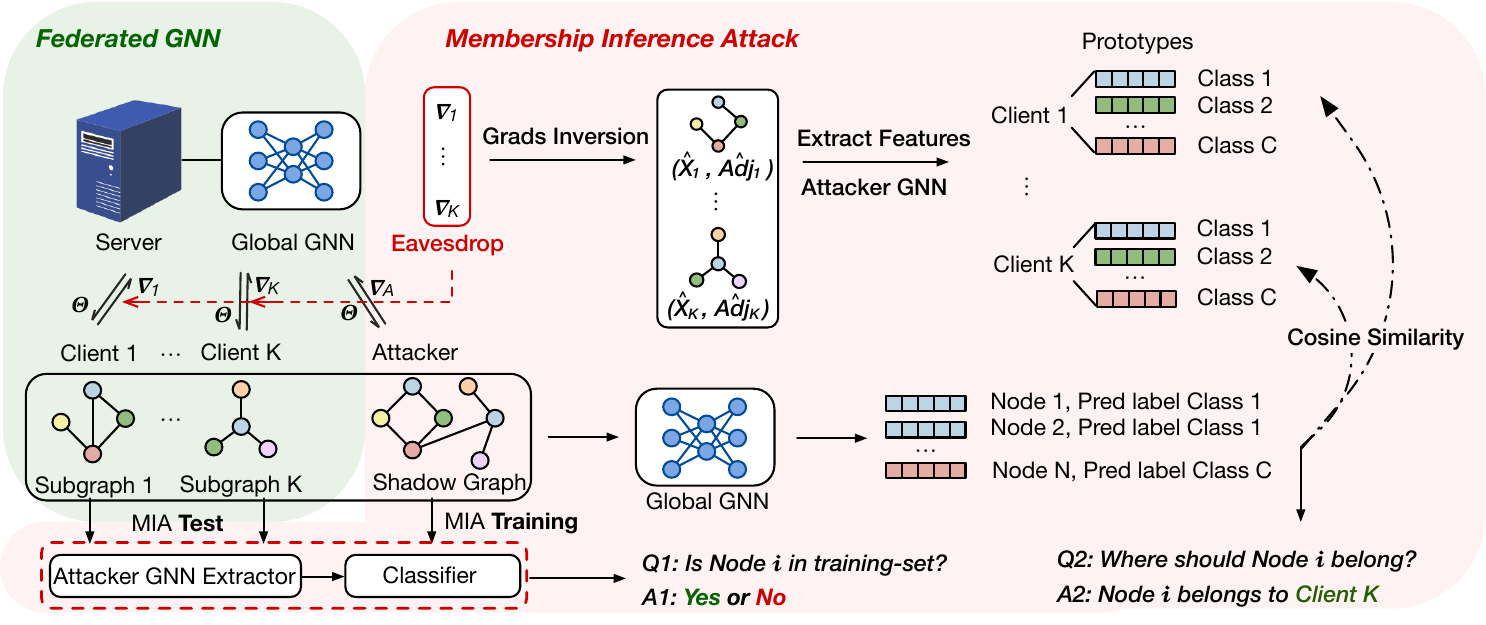}
  \caption{\textbf{Framework overview of \ours.} Our MIA is designed to address two objectives: (i) Membership inference, where the attacker employs a public shadow dataset with a similar structure to the target data to train a binary classifier, enabling inference of whether a given node belongs to the training set; and (ii) Client-data identification, which employs gradient inversion to reconstruct pseudo node features and adjacencies. The reconstructed features, extracted through the attacker GNN, generate class-specific prototypes for each client. By comparing the node features with these prototypes, \ours determines the client ownership of the nodes. The time and space complexities of \ours are discussed in Appendix~\ref{complexity}.}
  \label{framework}
\end{figure*}

Based on the motivation discussed in Appendix~\ref{motivation}, we extend the scope of MIAs on GNNs from centralized to federated scenarios. This section details the \ours for FedGNNs, decomposing the attack into two objectives: (i) training‑set membership inference and (ii) client‑data-ownership identification. The overall architecture is illustrated in Fig~\ref{framework}.  


\subsection{Membership Inference Attack}
\paragraph{Local GNN Training.}Given the threat model and Definition~\ref{def_training_set_mia}, the target graph is partitioned by METIS~\cite{1998metis} across $K$ clients as $ G(X, A)={\,G_1(X_1, A_1), \dots, G_k(X_K, A_K)}$, where client $k$ holds subgraph $G_k$ with node features $X_k$ and adjacency $A_k$. 

During each federated training round $t$, client $k$ computes its local gradient:
\begin{equation}
\mathcal{G}_k^t \;=\; \nabla_W \,\mathcal{L}\bigl(\mathcal{F}\bigl(X_k,A_k;W^t\bigr),\,Y_k\bigr),
\end{equation}
where $\mathcal{F}(\cdot;W)$ is the GNN forward pass and $\mathcal{L}$ the cross-entropy loss over client $k$’s labeled nodes $Y_k$. These gradients $\{\mathcal{G}_k^t\}_{k=1}^K$ are then uploaded to the server.

\paragraph{Attacker GNN Training} The attacker, acting as client $a$, similarly possesses its own local subgraph $G_a(X_a, A_a)$ and additionally acquires a shadow dataset $\tilde G(\tilde X,\tilde A)$ drawn from the similar structurally public datasets. Utilizing $\tilde G$, the attacker trains an attacker GNN $\tilde {\mathcal{F}}$ to extract features for a downstream classifier. The training process involves iterative gradient updates, formulated as:
\begin{equation}
    \tilde {\mathcal{G}}^t \;=\; \nabla_{\tilde W} \,\mathcal{L}\bigl(\tilde{\mathcal{F}}\bigl(\tilde X, \tilde A;\tilde W^t\bigr),\,\tilde Y\bigr), \quad
    \tilde W^{t + 1} = \tilde W^{t} - \eta \Delta \tilde W^t
\end{equation}

\paragraph{Federated Aggregation}The server incorporates weight decay and momentum into the gradient aggregation. Denoting the momentum buffer by $M^t$ and a weight‐decay coefficient $\lambda$, the update at round $t$ becomes:
\begin{equation}
    \Delta W^t 
    = \frac{1}{K}\sum_{k=1}^K \mathcal{G}_k^t
          \;+\;\lambda\,W^t,\\
    M^{t+1}
    = \mu\,M^t \;+\; \Delta W^t,\\
    W^{t+1}
    = W^t \;-\;\eta\,M^{t+1},
\end{equation}
where $\mu\in[0,1)$ is the momentum coefficient and $\eta$ the global learning rate.  

\paragraph{Feature Extraction.}$\text{Lap} = \tilde D^{-1/2}(\tilde D-\tilde A)\tilde D^{-1/2}$ is the normalized laplacian matrix with the degree matrix $\tilde D$ and a nonlinear activation $\sigma(\cdot)$. Specially, the attacker gets the 1st‐layer attacker GNN embeddings on its shadow graph $(\tilde X,\tilde A)$:
\begin{equation}
    \tilde{\mathcal{E}}
\;=\;
\sigma(\text{Lap}\,\tilde X\,W^{(1)} + \mathbf{1}\,b^{(1)\mathsf{T}}),
\end{equation}
where $W^{(1)} \in \mathbb{R}^{ \tilde{D} \times \tilde{D} }$ and $b^{(1)} \in \mathbb{R}^ {\tilde{D}}$ are the 1st‐layer weights and bias, and $\mathbf{1}\in\mathbb{R}^{|\tilde X|\times1}$ replicates the bias vector. By making the aggregation and embedding operations, we align the formulation with practical FedGNN implementations and clarify the attacker's embedding pipeline.

\paragraph{Training Classifier.}A classifier $f_w$ composed of multiple fully connected layers interleaved with batch normalization is employed to conduct membership inference, as follows:
\begin{equation}
    \begin{aligned}
    H^{(l)} = M^{(l)} \;\odot\;&\max(B^{(l)}\bigl(W^{(l)} H^{(l-1)} + b^{(l)}),\,0\bigr)
    \quad (l = 1, \cdots, L),\\
    \tilde{\mathcal{Z}} &= \mathrm{softmax}\!\bigl(W^{(L)} H^{(L-1)} + b^{(L)}\bigr),
    \end{aligned}
\end{equation}
we set $H^{(0)} = \tilde{\mathcal{E}}$, where for each layer $l$, $W^{(l)}, b^{(l)}$ are the weight matrix and bias vector. $B^{(l)}(\cdot)$ denotes the affine batch‐normalization operator, $M^{(l)}\in\{0,1\}^D$ is a random binary mask, and $\max(x,0)$ applies the rectified linear unit elementwise. $\odot$ denotes elementwise multiplication.

The classifier processes feature representations $\tilde{\mathcal{E}}$ that are derived through the shadow GNN, ensuring alignment with the target model’s inductive biases and data distribution. This alignment mitigates the risk of overfitting to raw node attributes, which is trained by optimizing the cross-entropy loss over the shadow dataset:
\begin{equation}
    \min_{w} \; \mathbb{E}_{i \sim \tilde G}\Bigl[
    -\tilde m_i \log \sigma\bigl(f_w(\tilde{\mathcal{E}}_i)\bigr)
    -(1-\tilde m_i)\log \bigl(1 - \sigma\bigl(f_w(\tilde{\mathcal{E}}_i)\bigr)\bigr)
    \Bigr],
\end{equation}
where $f_w(\tilde{\mathcal{E}}_i)$ is the predicted probability of class $c\in\{0,1\}$. $\tilde m_i\in\{0,1\}$ indicates whether node $i$ belongs to the shadow training-set.


\subsection{Client-data Identification}
\label{client-data_indentification}
Given the threat model and Definition~\ref{def_client_ownership_mia}, eavesdropping on gradients uploaded by clients to the server can infer the ownership of target nodes. This attack contains two stages: (1) gradient inversion, which reconstructs the subgraph associated with each client, and (2) prototype-based client matching, which assigns reconstructed nodes to their respective clients based on specific patterns.

\paragraph{Eavesdropping} At each round $t$, eavesdropping is modeled as $\mu_t \sim \text{Bernoulli}(\gamma)$ \cite{wu2025secret, hu2021federated}, where $\mu_t = 1$ means successful interception.  
The adversary observes the client update $\xi_t$ if $\mu_t = 1$; otherwise, it estimates $\xi_t$ and the proxy term $\zeta_t$.  It knows whether the client was selected ($\delta_t$), and $\mu_t$, $\delta_t$ are independent over time.  
A local proxy rule is used to simulate updates when interception fails.  
The estimate $x_t^a$ is updated by interpolating between observed and proxy signals. Estimation accuracy is measured by $\mathbb{E}[\|x_t^c - x_t^a\|^2]$.

\paragraph{Gradient Inversion for Client Subgraph.}
As discussed in the threat model, an attacker can intercept gradients from individual clients through eavesdropping. Specifically, each client's 1st-layer gradients are exploited to reconstruct the node features and the adjacency. Gradients are denoted as $ g \in \mathbb{R}^{d_\text{in} \times d_\text{hid}} = \nabla_W \mathcal{L} $, where $ W $ represents the parameter weights of the first GCN layer, and $ \mathcal{L} $ denotes the loss of each client GNN.

To reconstruct client data from gradients, it is necessary to generate synthetic gradients, denoted as $ \nabla_W \hat{\mathcal{L}} $, and iteratively optimize them to approximate the intercepted real gradients. As discussed in \cite{2024gnngradientinversion}, this approximation is achieved by minimizing the negative cosine similarity, formulated as:
\begin{equation}
    \mathbb{L} = 1 - \frac{\nabla_W \mathcal{L} \cdot \nabla_W \hat{\mathcal{L}}}{\left\| \nabla_W \mathcal{L} \right\| \left\| \nabla_W \hat{\mathcal{L}} \right\|}
\end{equation}

The synthetic gradient $ \nabla_W \hat{\mathcal{L}} $ is generated by feeding a pseudo-graph input into the GNN. In \ours, we assume that the input labels are known, as they can be easily inferred from the gradients in classification tasks with cross-entropy loss \cite{2020idlg}. Many real-world graphs, such as social networks, exhibit the property of feature smoothness, where connected nodes tend to have similar attributes. To ensure the reconstructed graph data adheres to this smoothness property, we adopt the smoothness loss proposed in \cite{2021graphmi} for gradient fitting:

\begin{equation}
\mathbb{L}_{\text{smooth}} = \text{tr}(X^T \hat{\text{Lap}} X) = \frac{1}{2} \sum_{i,j=1}^{N} A_{ij} \left( \frac{x_i}{\sqrt{d_i}} - \frac{x_j}{\sqrt{d_j}} \right)^2,
\end{equation}

where $ \hat{\text{Lap}} $ denotes the normalized laplacian matrix of $A$ and $D$ is the diagonal matrix of $A$, $ A_{ij} $ denotes elements of the adjacency, and $ d_i $ represents the degree of node $ i $.

To enhance the sparsity of the graph structure, the Frobenius norm of the adjacency matrix $ A $ is incorporated into the loss function. The final objective function is defined as:  

\begin{equation}
    \hat{\mathbb{L}} = \mathbb{L} + \alpha \mathbb{L}_{\text{smooth}} + \beta \| \hat{A} \|_F^2,
    \label{grad_loss_final}
\end{equation}

where $ \| \hat{A} \|_F^2 $ denotes the adjacency during optimization. The coefficients $ \alpha $ and $ \beta $ control the relative importance of the smoothness and sparsity constraints, respectively.

The features of the reconstructed nodes, $ \hat{X} $, can be used directly for the downstream task. However, the reconstructed adjacency requires further mapping and sampling. Specifically, the gradient descent step for the adjacency includes a projection step. During this step, each entry of the adjacency $ \tilde{A}_{ij} $ is updated using an entry-wise projection operator defined as follows:

\begin{equation}
    \hat{A}_{ij} = \text{proj}_{[0,1]}(\tilde{A}_{ij}) = 
    \begin{cases} 
        1, & \tilde{A}_{ij} > 1 \\ 
        0, & \tilde{A}_{ij} < 0 \\ 
        \tilde{A}_{ij}, & \text{otherwise} 
    \end{cases},
\end{equation}

To refine the adjacency further, we retain only the top $N_e$ edges with the largest weights, setting their values to 1, while all other entries are set to 0:

\begin{equation}
    \hat{A}_{ij} =
    \begin{cases}
    1, & \text{if } (i,j) \in E_\text{top}, \\
    0, & \text{otherwise,}
    \end{cases}
\label{sample_edges}
\end{equation}

where $E_\text{top}$ denotes the set of $N_e = \rho \cdot N$ edges with the highest weights in $\hat{A}$. $\rho$ represents the known edge density, and $N$ is the number of nodes.

Specifically, \ours utilizes Algorithm~\ref{grad_inversion_alg} to reconstruct the client’s node features and adjacency.

\paragraph{Prototype-based Client Matching.}
As an adversarial client within the FedGNNs, the attacker gains access to the global GNN $\mathcal{F}$ distributed by the server. By leveraging this model, the attacker feeds a reconstructed graph into $\mathcal{F}$ to generate pseudo-embeddings $\hat{\mathcal{E}}$ for nodes associated with different target clients. These pseudo-embeddings serve as the foundation for constructing class-specific prototypes for each targeted client:
\begin{equation}
    \label{obtain_prototype}
    \begin{aligned}
        &\hat{\mathcal{E}}_k = \mathcal{F}_1(\hat{X}, \hat{A}), \\
        \mathcal{P}_k =& \left\{ c \mapsto \frac{1}{|\mathcal{I}^k_c|} \sum_{j \in \mathcal{I}^k_c} \hat{\mathcal{E}}_k^j \,\middle|\, c \in \mathcal{C}_k \right\},
    \end{aligned}
\end{equation}
where $\mathcal{C}_k$ is the set of all classes of client $k$. $\mathcal{I}_c^k$ is the sample index set labeled as $c$ in client $k$.

Each prototype set of client $k$ is $\mathcal{P}_k = \left\{ c \mapsto \mu_c^{(k)} \mid c \in \mathcal{C}_k \right\}$, $ \mu_c^{(k)} \in \mathbb{R}^D $ represents the prototype vector of class $ c $ for client $ k $, and $ \mathcal{C}_k $ denotes the set of classes for client $ k $.

The real node information to be analyzed is passed to the adversary. By measuring node embeddings generated by the GNN 1-st layer for extraction with the class prototypes of each target client:
\begin{equation}
    \label{compute_distance}
    \begin{aligned}
        &\mathcal{E}= \mathcal{F}_1(X, A), \\
        d_{i,k} =& \left\{\begin{matrix}
                          1 - \frac{\mathcal{E}_k^i \cdot \mu^{(k)}_{y_i}}{\|\mathcal{E}_k^i\| \cdot \|\mu^{(k)}_{y_i}\|} & y_i \in \mathcal{C}_k\\
                          \infty & \text{otherwise}
                        \end{matrix}\right.,
    \end{aligned}
\end{equation}
the target node is assigned to the client with the closest prototype, determined by $\hat{k_i} = \arg \min_{k \in \{1,\cdots, K\}}d_{i,k}$. The complete matching process is detailed in Algorithm~\ref{prot_match_alg}.

\section{Experiments}
\subsection{Baselines}
\label{details}

In our experiments, we evaluate three representative GNN models, including GCN~\cite{2017gcn}, GAT~\cite{gat2018}, and GraphSAGE~\cite{2017graphsage} as global models within the FedGNN framework. To demonstrate the generalizability of \ours, results using GAT and GraphSAGE are presented in Appendix~\ref{cc-mia_on_other_gnn}. We further assess \ours under five widely adopted federated approaches: FedAvg~\cite{2017fedavg}, FedProx~\cite{2020fedprox}, SCAFFOLD~\cite{2020scaffold}, FedDF~\cite{2020feddf}, and FedNova~\cite{2020fednova}. Detailed descriptions can be found in Appendix~\ref{approaches}.

The evaluation is conducted on six benchmark datasets for node classification: Cora~\cite{2016plaintoid}, Citeseer~\cite{2016plaintoid}, PubMed~\cite{2016plaintoid}, CS, Physics~\cite{2018pitfall}, and DBLP~\cite{2018dblp}, with dataset details summarized inAppendix~\ref{dataset_statistics}. For comparison, we include three strong MIA baselines: HP-MIA~\cite{2024hpmia}, GAN-Based Data Enhancement~\cite{2023ganenhance}, and CS-MIA~\cite{2022csmia}, described in Appendix~\ref{baselines}. Lastly, the computational cost of \ours and explore potential defense strategies in Appendix~\ref{cost} and Appendix~\ref{defense}, respectively.

\subsection{Experimental Settings}
\label{settings}

In our experiments, the target dataset is partitioned into $ k $ subgraphs using the METIS partitioning~\cite{1998metis}, with each subgraph assigned to one of $ k $ clients, which is widely applied in real-world applications for subgraphs~\cite{wang2025fast, zhu2025simplifying, chen2025leapgnn, ma2025adaptive}. Due to the graph structure, nodes of similar classes are clustered closely within subgraphs, leading to a non-i.i.d. distribution of nodes across clients. A statistical analysis of the node class distributions for each client is provided in Fig~\ref{fig_client_class_distribution}. Specifically, 40\% of the target dataset is allocated as the training-set, while the entire shadow dataset is used for the training-set MI attack model. Both the GNNs and the attacker models are trained with a learning rate of $ 1e-3 $. The number of hidden neurons in the GNN and the attack model is set to 128. For the hyperparameters in Eq.~\ref{grad_loss_final}, $ \alpha $ and $ \beta $ are set to $ 1e-3 $ and $ 1e-4 $, respectively. We evaluate membership inference and client-data-ownership using \textbf{AUC} and \textbf{Accuracy}, respectively, while \textbf{AUC} and \textbf{RNMSE} assess gradient inversion quality. Details are in Appendix~\ref{metrics}.

\subsection{Membership Inference Attack Results}
\label{training_set_mia}
\begin{table*}
\centering
\setlength{\tabcolsep}{0.6pt}
\small
\caption{Performance of member inference attacks and client-data identification. Clts: Clients.}
\begin{tabular}{ccccccc|cccccccc}
\toprule
Dataset & Approach & Shadow & HP-MIA & GAN-Based & CS-MIA & \ours & 3-Clts & 4-Clts & 5-Clts & 6-Clts & 7-Clts & 8-Clts & 9-Clts & 10-Clts \\\midrule
\multicolumn{2}{c}{\textbf{Client-uniform Probability}} & - & - & - & - & - & \textbf{33.33} & \textbf{25.00} & \textbf{20.00} & \textbf{16.66} & \textbf{14.29} & \textbf{12.50} & \textbf{11.11} & \textbf{10.00} \\
\midrule

\multirow{5}{*}{Cora} & FedAvg & DBLP & 51.05 & 53.15 & 57.87 & \textbf{82.04} & 59.27 & 55.61 & 56.31 & 38.92 & 34.08 & 33.20 & 27.18 & 27.95 \\
 & FedProx & DBLP & 50.79 & 50.42 & 74.72 & \textbf{83.31} & 55.80 & 47.56 & 40.66 & 36.15 & 40.47 & 28.66 & 23.67 & 19.13 \\
 & SCAFFOLD & DBLP & 52.04 & 52.09 & 69.04 & \textbf{83.38} & 64.99 & 42.32 & 34.12 & 35.75 & 41.99 & 35.56 & 25.59 & 22.01 \\
 & FedDF & DBLP & 52.29 & 51.89 & 72.29 & \textbf{82.71} & 36.74 & 39.73 & 40.07 & 29.87 & 25.70 & 25.48 & 23.49 & 19.46 \\
 & FedNova & DBLP & 54.08 & 56.16 & 74.51 & \textbf{84.37} & 39.00 & 30.80 & 23.67 & 18.65 & 17.91 & 18.50 & 14.55 & 12.11 \\
\midrule

\multirow{5}{*}{Citeseer} & FedAvg & PubMed & 51.69 & 50.76 & 72.68 & \textbf{86.04} & 51.85 & 48.21 & 44.00 & 36.55 & 35.32 & 32.04 & 31.89 & 22.00 \\
 & FedProx & PubMed & 52.57 & 53.56 & 71.10 & \textbf{85.98} & 52.15 & 44.66 & 44.18 & 42.44 & 34.39 & 30.36 & 28.13 & 26.45 \\
 & SCAFFOLD & PubMed & 49.89 & 51.23 & 59.70 & \textbf{85.89} & 55.64 & 45.15 & 42.11 & 40.49 & 39.92 & 33.78 & 29.13 & 23.65 \\
 & FedDF & PubMed & 54.32 & 52.67 & 74.61 & \textbf{86.83} & 51.61 & 51.19 & 44.39 & 37.60 & 29.37 & 28.70 & 26.39 & 23.35 \\
 & FedNova & PubMed & 53.49 & 58.14 & 74.42 & \textbf{85.97} & 42.23 & 38.35 & 34.48 & 30.42 & 28.46 & 27.35 & 25.28 & 22.78 \\
\midrule

\multirow{5}{*}{PubMed} & FedAvg & DBLP & 55.74 & 50.23 & 58.14 & \textbf{71.73} & 49.56 & 41.56 & 32.68 & 28.47 & 25.55 & 19.16 & 17.32 & 14.52 \\
 & FedProx & DBLP & 54.98 & 54.12 & 58.79 & \textbf{71.84} & 46.24 & 32.91 & 26.03 & 23.24 & 22.15 & 21.41 & 15.67 & 12.23 \\
 & SCAFFOLD & Physics & 53.75 & 54.78 & 52.68 & \textbf{71.25} & 44.12 & 38.83 & 26.10 & 25.15 & 24.34 & 20.64 & 18.94 & 17.85 \\
 & FedDF & DBLP & 56.09 & 55.01 & 59.24 & \textbf{72.03} & 43.55 & 39.62 & 24.55 & 24.22 & 24.10 & 22.68 & 20.64 & 18.37 \\
 & FedNova & DBLP & 55.48 & 52.34 & 57.56 & \textbf{65.23} & 45.81 & 40.66 & 34.50 & 27.08 & 24.50 & 23.44 & 23.69 & 22.56 \\
\midrule
\multicolumn{2}{c}{Max Improve \%} & - & - & - & - & - & 95.0 & 122.4 & 181.6 & 154.7 & 277.0 & 184.5 & 187.0 & 179.5 \\
\bottomrule
\end{tabular}
\label{tab_cc-mia}
\end{table*}

To evaluate the attack performance, we first initialize \ours on the GCN as the FL global model.
Specifically, we report the shadow dataset for each target dataset that achieves the best attack performance when the number of clients is set to 5. \textbf{AUC} is employed to evaluate attack performance, ensuring that biases introduced by training-set proportion are effectively mitigated. The MIA results obtained using the optimal shadow data set are reported on the left side of Table~\ref{tab_cc-mia}. The proposed \ours consistently outperforms all MIA baselines across different federated settings. On Citeseer with SCAFFOLD, it achieves up to 72.16\% improvement. In contrast, HP-MIA and GAN-based methods, designed for centralized models, perform only slightly above random guessing (AUCs of 56.09\% and 58.14\% on PubMed and Citeseer, respectively). While CS-MIA performs better in federated settings, it still falls short due to its lack of GNN-specific design. Note that the superior performance of \ours with Physics as the shadow dataset under SCAFFOLD stems from its correction mechanism, enhancing alignment between Physics and PubMed for better inference.

More results on other datasets (CS, Physics, and DBLP) are reported in Appendix Table~\ref{tab_training_set_mia_appendix}. For generalization purposes, we also proposed the MIA attack results on GAT and GraphSAGE in Table~\ref{tab_training_set_mia_gat} and \ref{tab_training_set_mia_sage}, respectively.

\subsection{Client-data Identification Results}
\label{client-data-ownership_results}
We evaluate the performance of \ours's client-data identification attack within the framework of federated GCNs. Specifically, we test with the number of clients ranging from 3 to 10. A larger number of clients increases the difficulty of achieving a successful attack. For prototype computation, we use node features extracted from the 1-st GCN layer of the trained global GNN, the results are presented on the right side of Table~\ref{tab_cc-mia}. For comparison, we provide the baseline probability for $k$-clients, referred to as \textbf{Client-uniform Probability}. For the convenience of comparison, we attach the probabilities of random selection. The optimal performance of $\ours$ across different clients consistently exceeds client-uniform probability by at least 95\%. Notably, the classification accuracy shows a significant improvement as the number of clients increases, particularly beyond 7 clients, highlighting the method's scalability in more complex federated settings. More datasets can be referred to Table~\ref{tab_client_ownership_mia_appendix}. 


\subsection{Ablation Studies and Variants}
We conducted ablation studies using GCN to evaluate the impact of \ours components. As shown in Table~\ref{tab_ablation_study_trainingset}, removing the shadow dataset (\ours \textit{no shadow}) lowers AUC to below 60\% on Cora, PubMed, and Physics, though it still outperforms baselines. Using subgraph-based shadow datasets further reduces performance, as full shadow datasets can better capture data distributions.

For client data ownership identification, removing the prototype strategy (\ours \textit{no prot}) weakens class differentiation, reducing AUC from 27.95\% to 20.82\% for 10 clients. The \ours (\textit{norm}) variant performs near random (AUC 9.12\%), emphasizing cosine similarity’s effectiveness in high-dimensional feature alignment.
The details of the analysis of the variants implementation on \ours (subgraph) are discussed in Appendix~\ref{var_imp}.

\begin{table}[ht]
    \setlength{\tabcolsep}{1pt}
    \centering
    \small
    \begin{minipage}{.5\textwidth}
        \centering
        \caption{Ablations and Variants of member inference.}
        \begin{tabular}{ccccccc}
        \toprule
        Dataset                     & Cora         & Citeseer     & PubMed       & CS           & Physics      & DBLP        \\\midrule
        \ours                     &\textbf{82.04}&\textbf{86.04}&\textbf{71.73}&\textbf{81.60}&\textbf{73.00}&\textbf{83.42}\\
        \ours (\textit{no shadow})     & 57.86        & 67.01        & 59.57        & 64.66        & 57.58        & 61.63        \\
        \ours (\textit{subgraph}) & 62.92        & 79.69        & 67.76        & 76.62        & 70.89        & 78.09        \\\bottomrule
        \end{tabular}
        \label{tab_ablation_study_trainingset}
    \end{minipage}%
    \hspace{0.05\textwidth}
    \begin{minipage}{.4\textwidth}
        \centering
        \caption{Ablations and Variants of client-data identification.}
        \begin{tabular}{ccccccc}
        \toprule
        Client Number               & 3-Client     & 5-Client     & 7-Client     & 10-Client    \\\midrule
        \ours                     &\textbf{59.27}&\textbf{56.31}&\textbf{34.08}&\textbf{27.95}\\
        \ours (\textit{no prot})       & 51.99        & 43.46        & 31.68        & 20.82       \\
        \ours (\textit{norm}) & 35.16        & 18.76        & 12.90        & 9.12         \\\bottomrule
        \end{tabular}
        \label{tab_ablation_study_client}
    \end{minipage}
\end{table}

\subsection{Inversion Quality}

\begin{figure}[h]
  \centering
  \subfigure[Clt1-GT]{
    \includegraphics[scale=0.15]{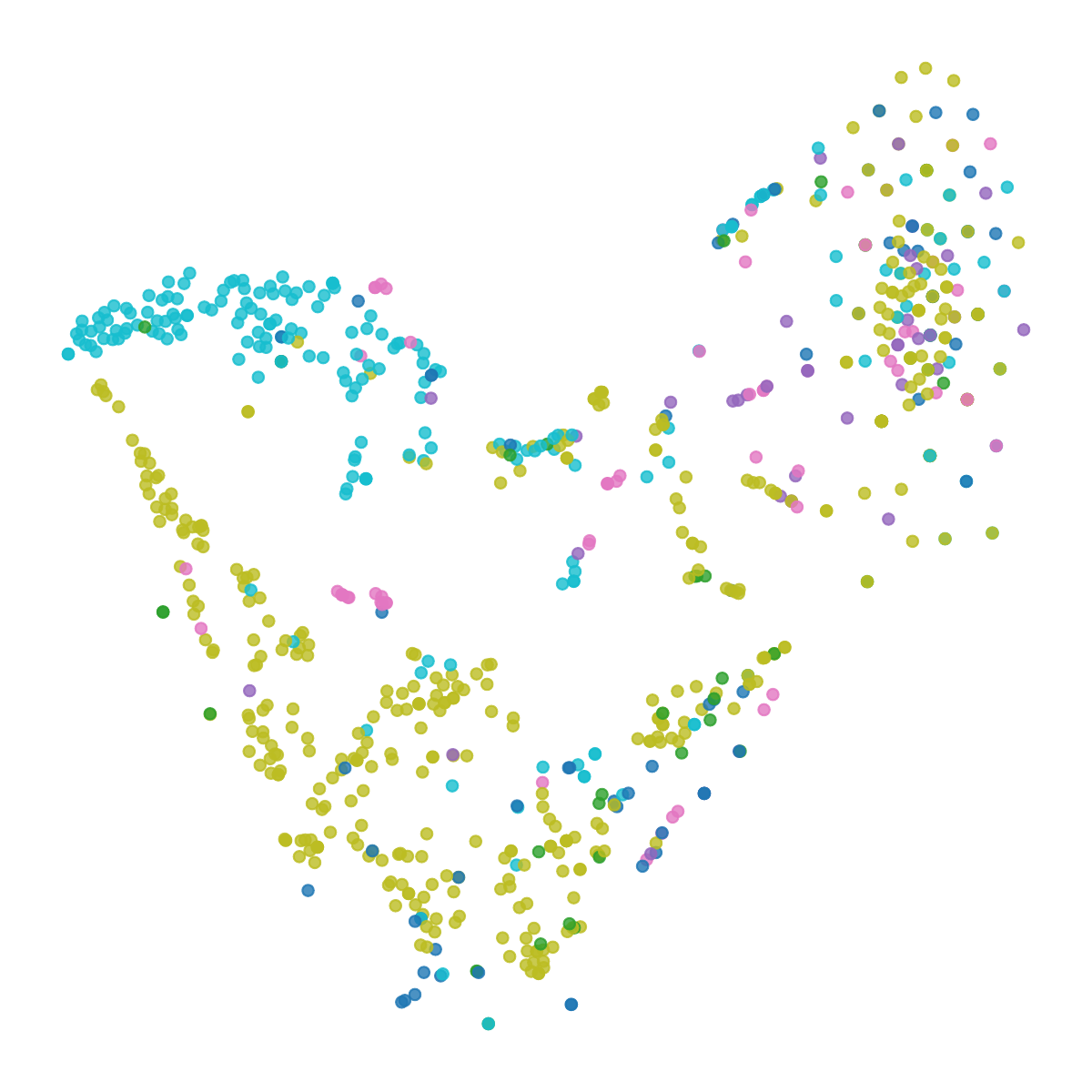}
    \label{citeseer_subgraph_0_true}
  }
  \subfigure[Clt2-GT]{
    \includegraphics[scale=0.17]{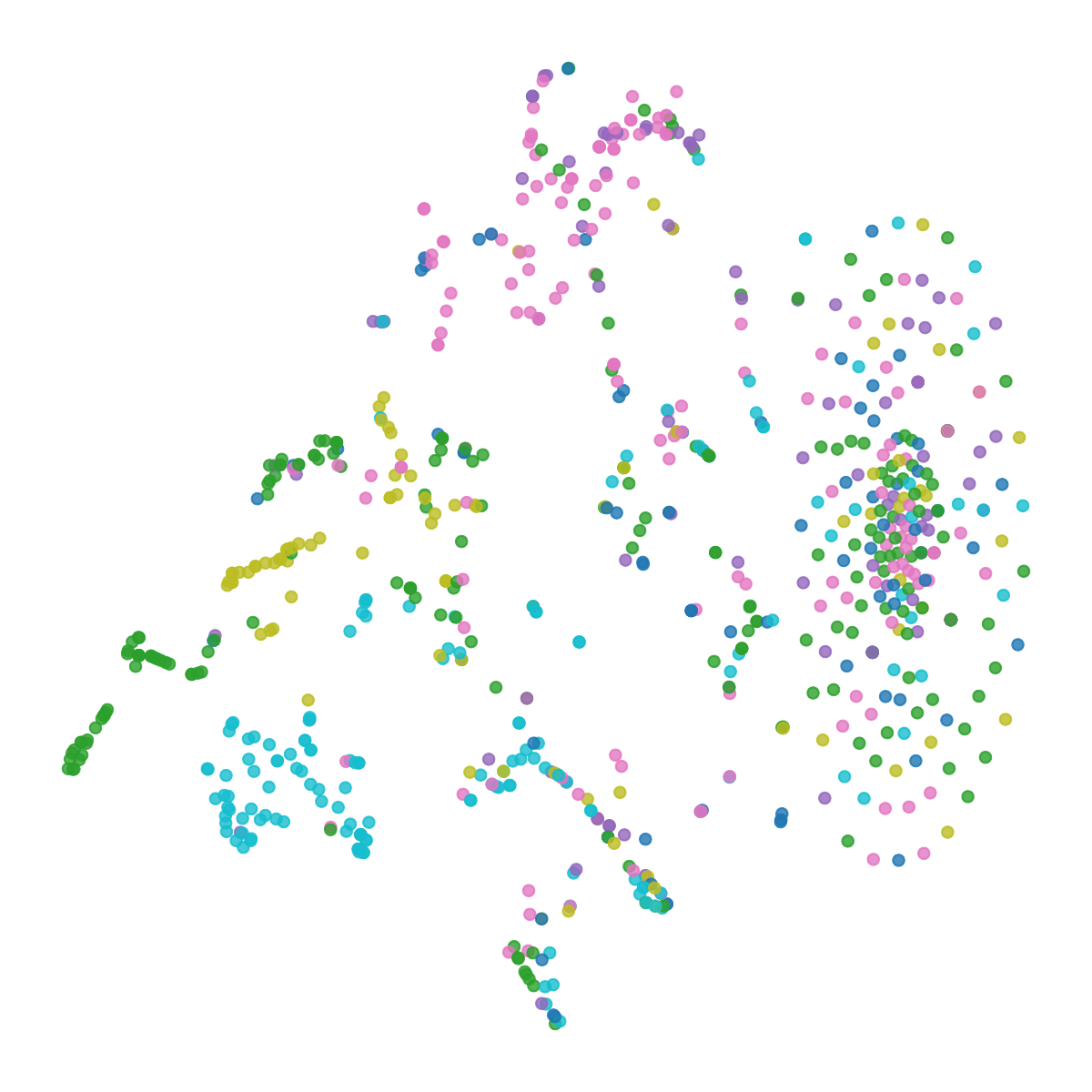}
    \label{citeseer_subgraph_1_true}
  }
  \subfigure[Clt3-GT]{
    \includegraphics[scale=0.17]{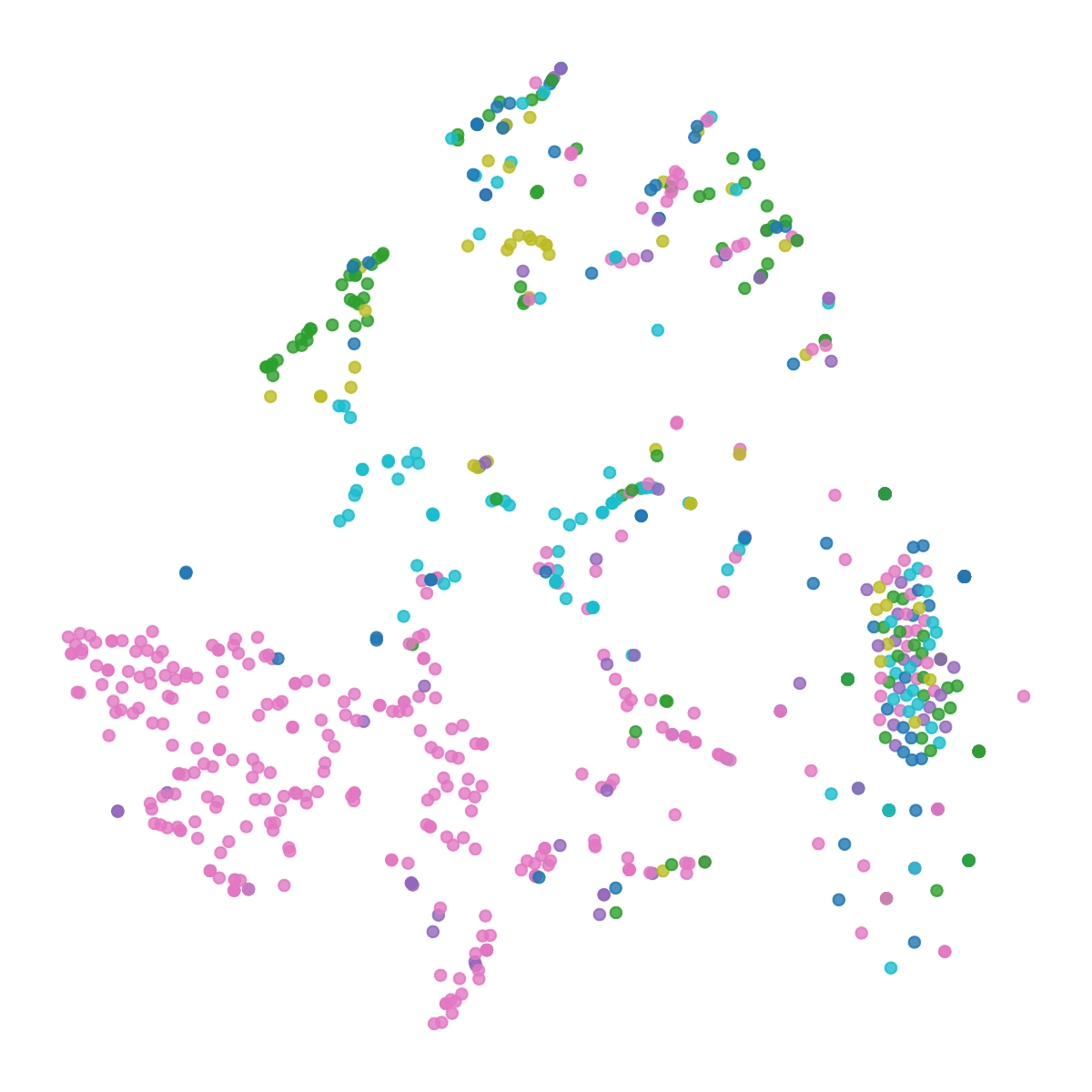}
  }
  \subfigure[Clt4-GT]{
    \includegraphics[scale=0.17]{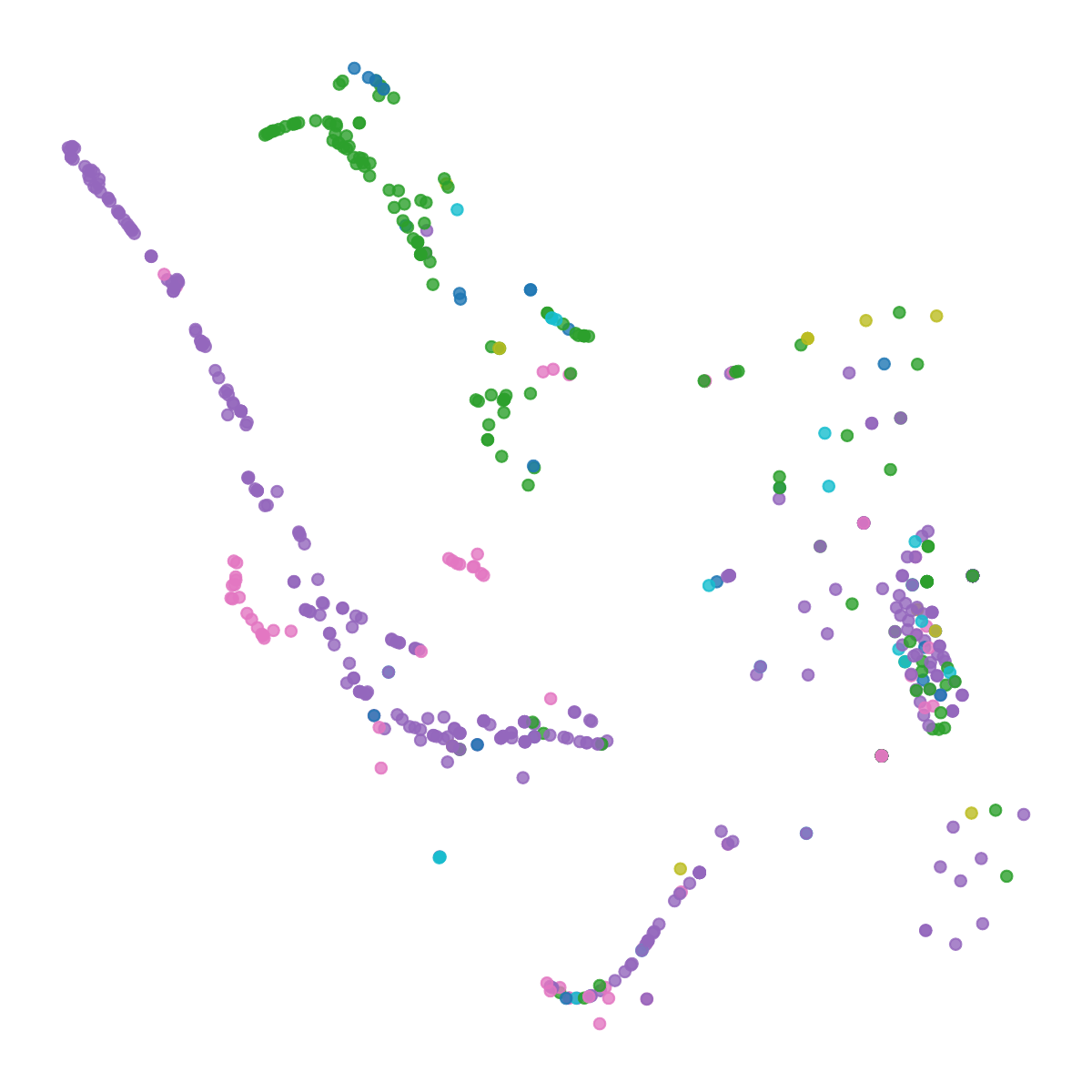}
  }
  \subfigure[Clt1-Inv]{
    \includegraphics[scale=0.17]{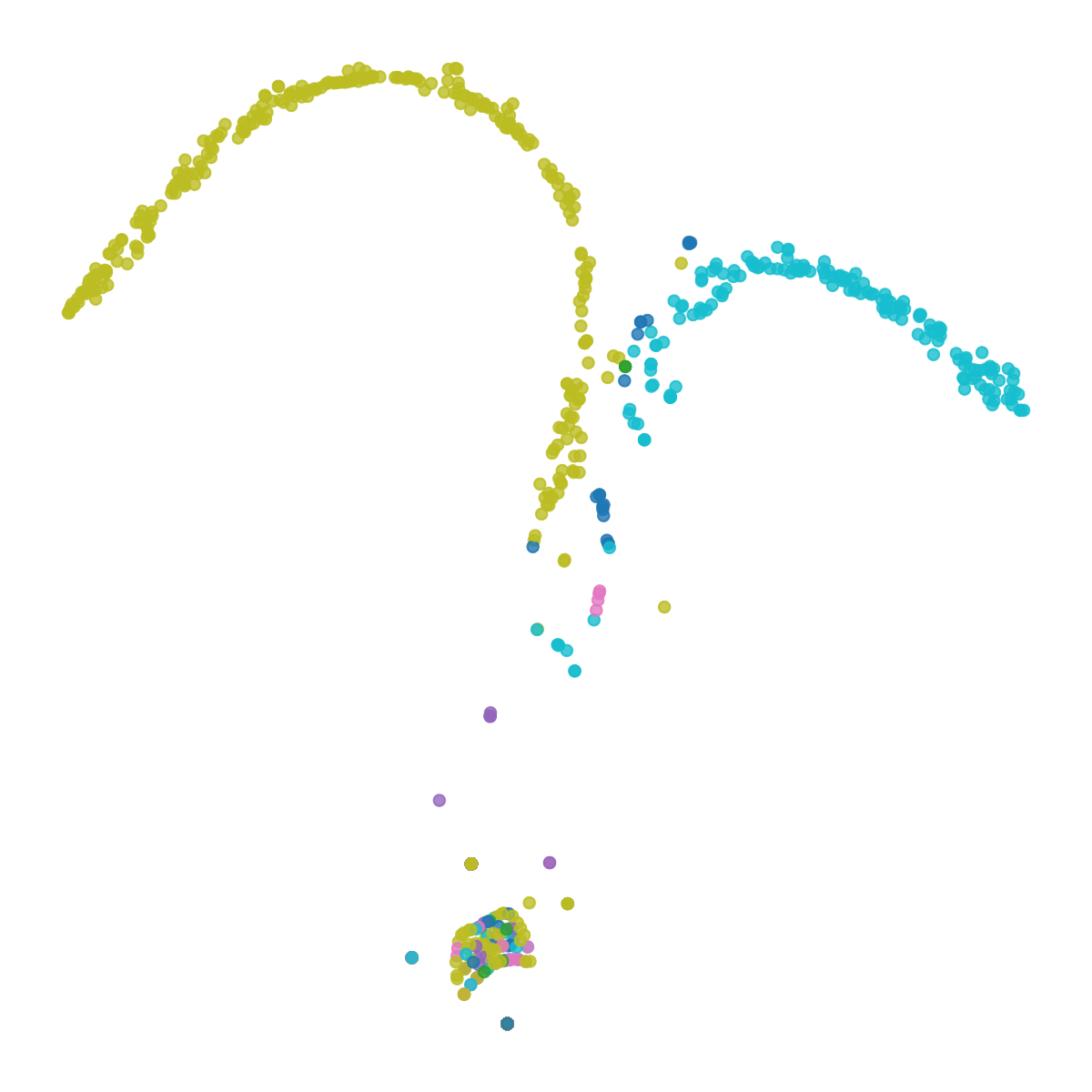}
  }
  \subfigure[Clt2-Inv]{
    \includegraphics[scale=0.17]{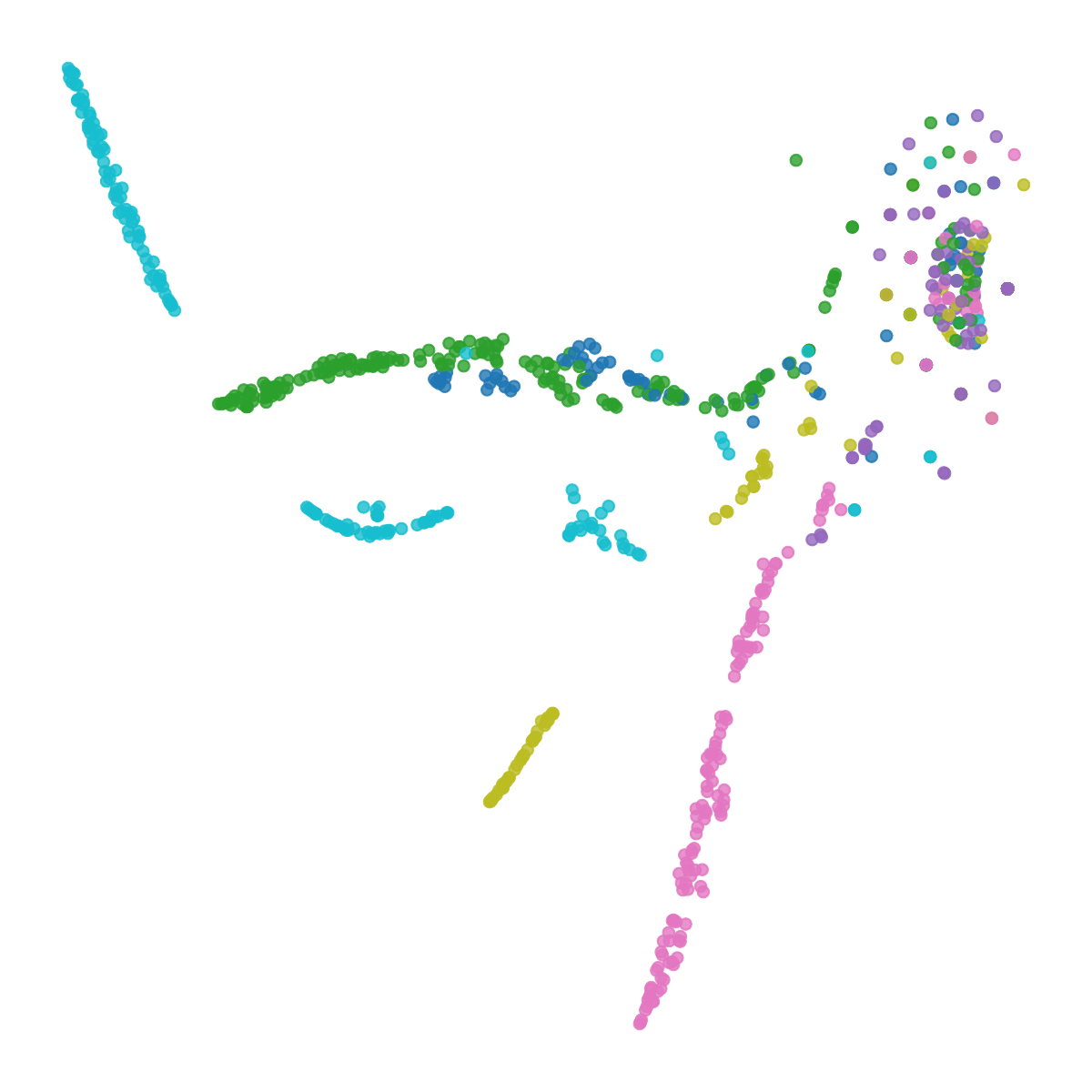}
  }
  \subfigure[Clt3-Inv]{
    \includegraphics[scale=0.17]{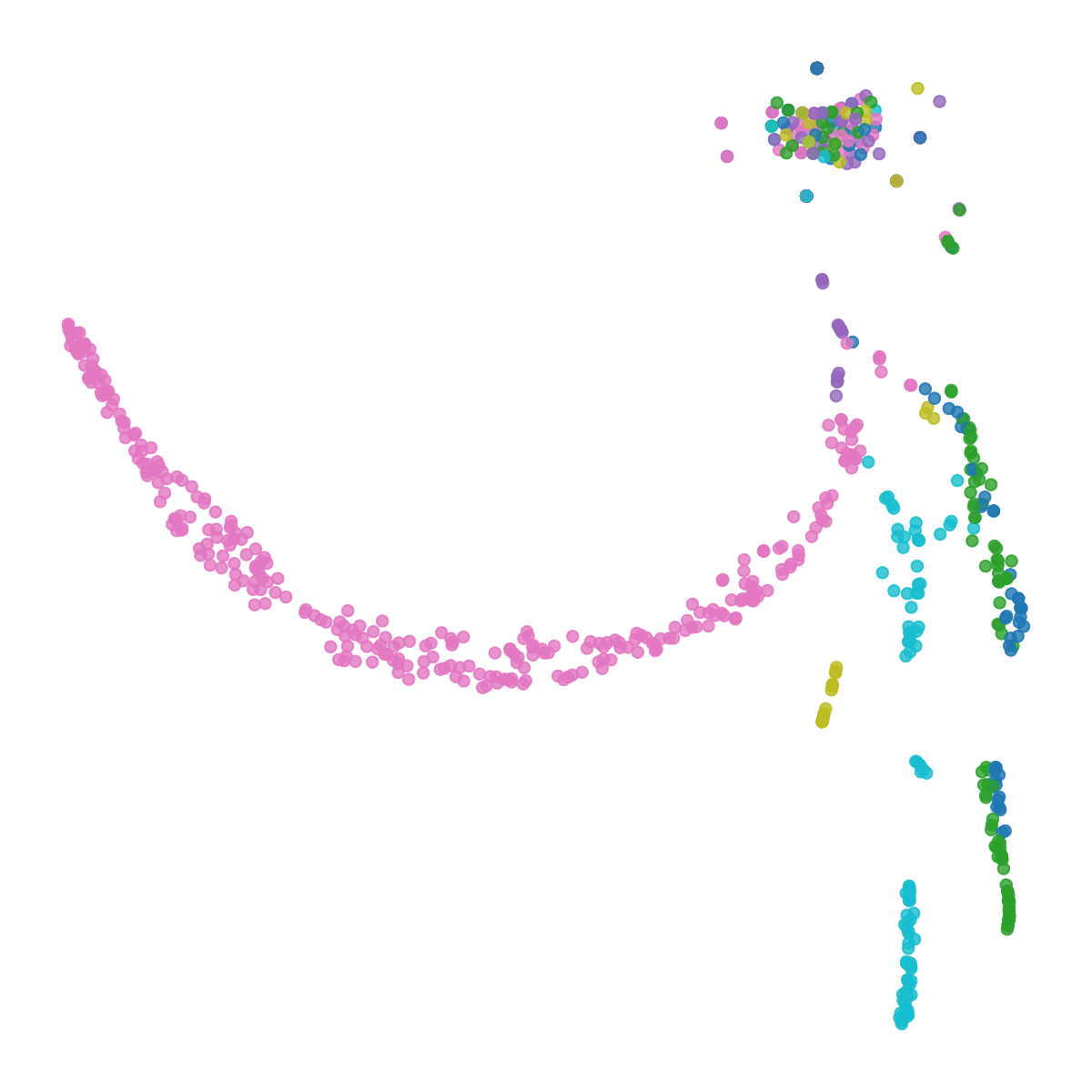}
  }
  \subfigure[Clt4-Inv]{
    \includegraphics[scale=0.17]{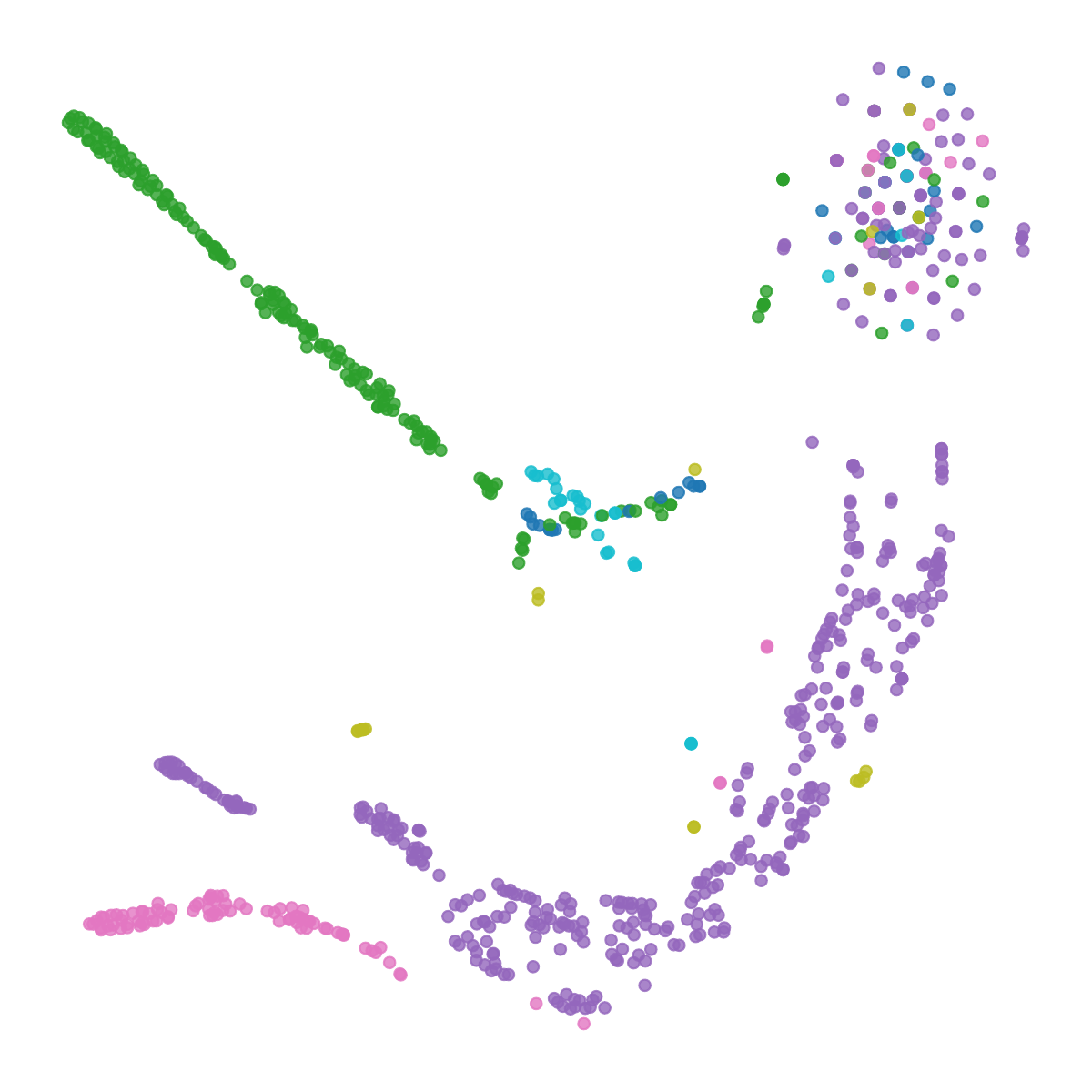}
  }
  \caption{Inverse comparison of node features of each client on Citeseer. Clt: Client; GT: Ground-truth; Inv: Inverse.}
  \label{fig_inversion_quality}
\end{figure}

\begin{wraptable}{r}{.47\textwidth}
\setlength{\tabcolsep}{2pt}
\centering
\small
\caption{Performance of \ours in the gradient inversion of the federated GCN}
\begin{tabular}{cccccc}
\toprule
\multicolumn{2}{c}{Client Number}                        & 3-Client & 5-Client & 7-Client & 10-Client \\\midrule
\multirow{2}{*}{Cora}     & AUC $\uparrow$   & 76.52    & 69.78    & 68.48    & 69.59     \\
                          & RNMSE $\downarrow$& 0.0027   & 0.0028   & 0.0041   & 0.0044    \\\hline
\multirow{2}{*}{Citeseer} & AUC $\uparrow$   & 76.09    & 74.23    & 74.78    & 64.89     \\
                          & RNMSE $\downarrow$& 0.0017   & 0.0021   & 0.0024   & 0.0026    \\\hline
\multirow{2}{*}{PubMed}   & AUC $\uparrow$   & 65.98    & 62.67    & 61.22    & 60.87     \\
                          & RNMSE $\downarrow$& 0.0018   & 0.0022   & 0.0021   & 0.0030    \\\hline
\multirow{2}{*}{CS}       & AUC $\uparrow$   & 79.07    & 72.71    & 72.44    & 67.72     \\
                          & RNMSE $\downarrow$& 0.0002   & 0.0003   & 0.0003   & 0.0005    \\\hline
\multirow{2}{*}{Physics}  & AUC $\uparrow$   & 68.67    & 65.43    & 62.34    & 62.53     \\
                          & RNMSE $\downarrow$& 0.0002   & 0.0002   & 0.0002   & 0.0003    \\\hline
\multirow{2}{*}{DBLP}     & AUC $\uparrow$   & 69.52    & 65.12    & 64.48    & 65.01     \\
                          & RNMSE $\downarrow$& 0.0013   & 0.0017   & 0.0019   & 0.0019    \\\bottomrule
\end{tabular}
\label{tab_inversion_quality}
\end{wraptable}

We measure the fidelity of gradient‐inverted subgraphs using edge‐AUC (higher is better) and feature‐RNMSE (lower is better) from~\cite{2021graphmi,2024gnngradientinversion} (Appendix~\ref{metrics}). Table~\ref{tab_inversion_quality} shows reconstruction quality declining as client count increases—for example, PubMed’s adjacency AUC falls from 65.98\% (3 clients) to 60.87\% (10 clients), while RNMSE rises from 0.0018 to 0.0030. Larger, more complex graphs like PubMed and Physics maintain relatively low RNMSE even with many clients, whereas small graphs (\textit{e.g.}, CS) achieve the highest AUC (79.07\%) and lowest RNMSE (0.0002) under three‐client settings. 
Fig~\ref{fig_inversion_quality} visualizes reconstructed versus true subgraph embeddings on Citeseer (4 clients) using t-SNE on first‐layer GCN features. Reconstructed clusters are more compact, class-consistent, and often exhibit greater linear separability.  This reduces ambiguity in dense regions and enhances the preservation of class-specific structures.

\subsection{Prototype Visualization}
\begin{figure}
  \centering
  \subfigure[Client3]{
    \includegraphics[scale=0.23]{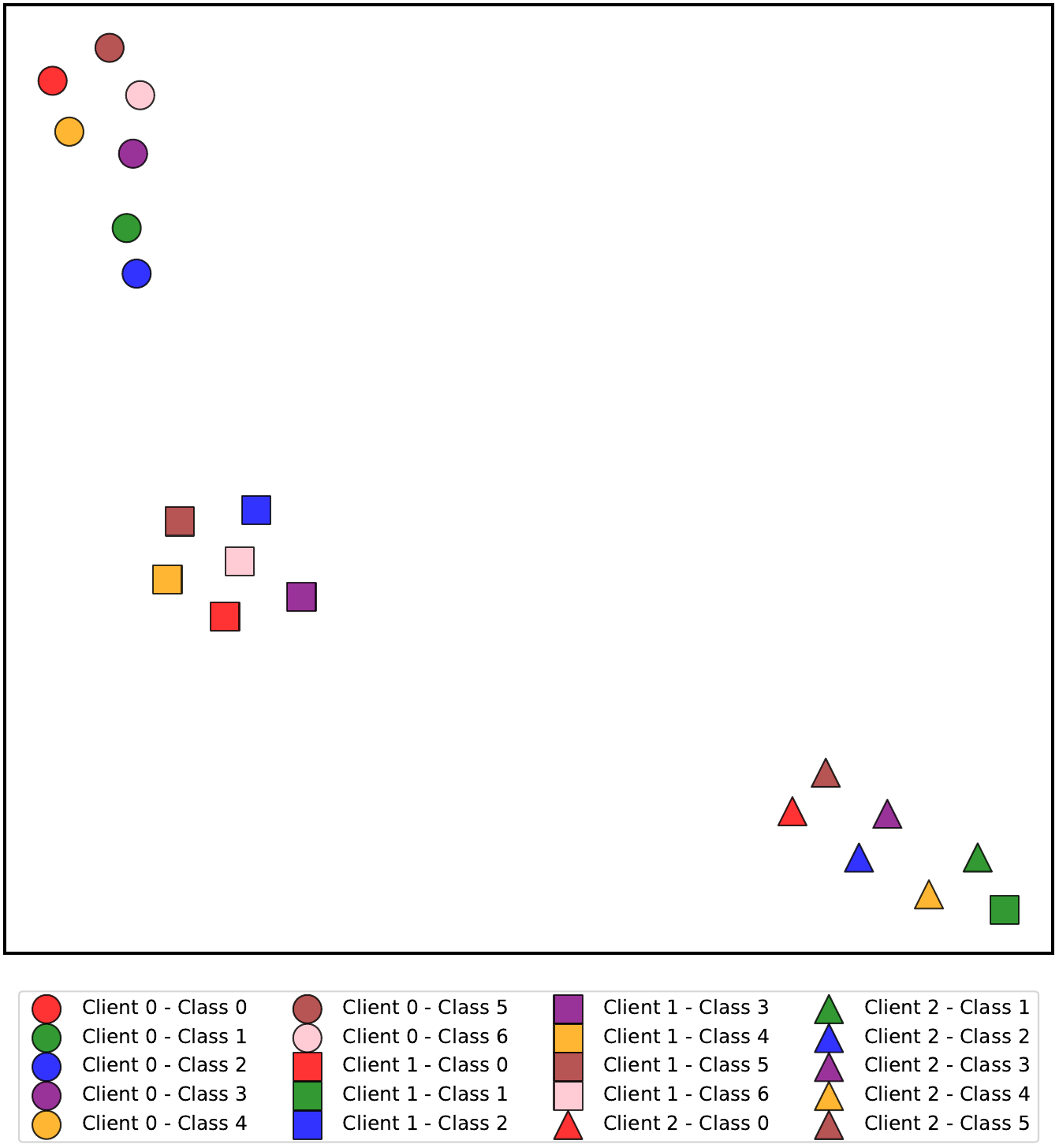}
    \label{fig_prototype_visualization_a}
  }
  \subfigure[Client4]{
    \includegraphics[scale=0.23]{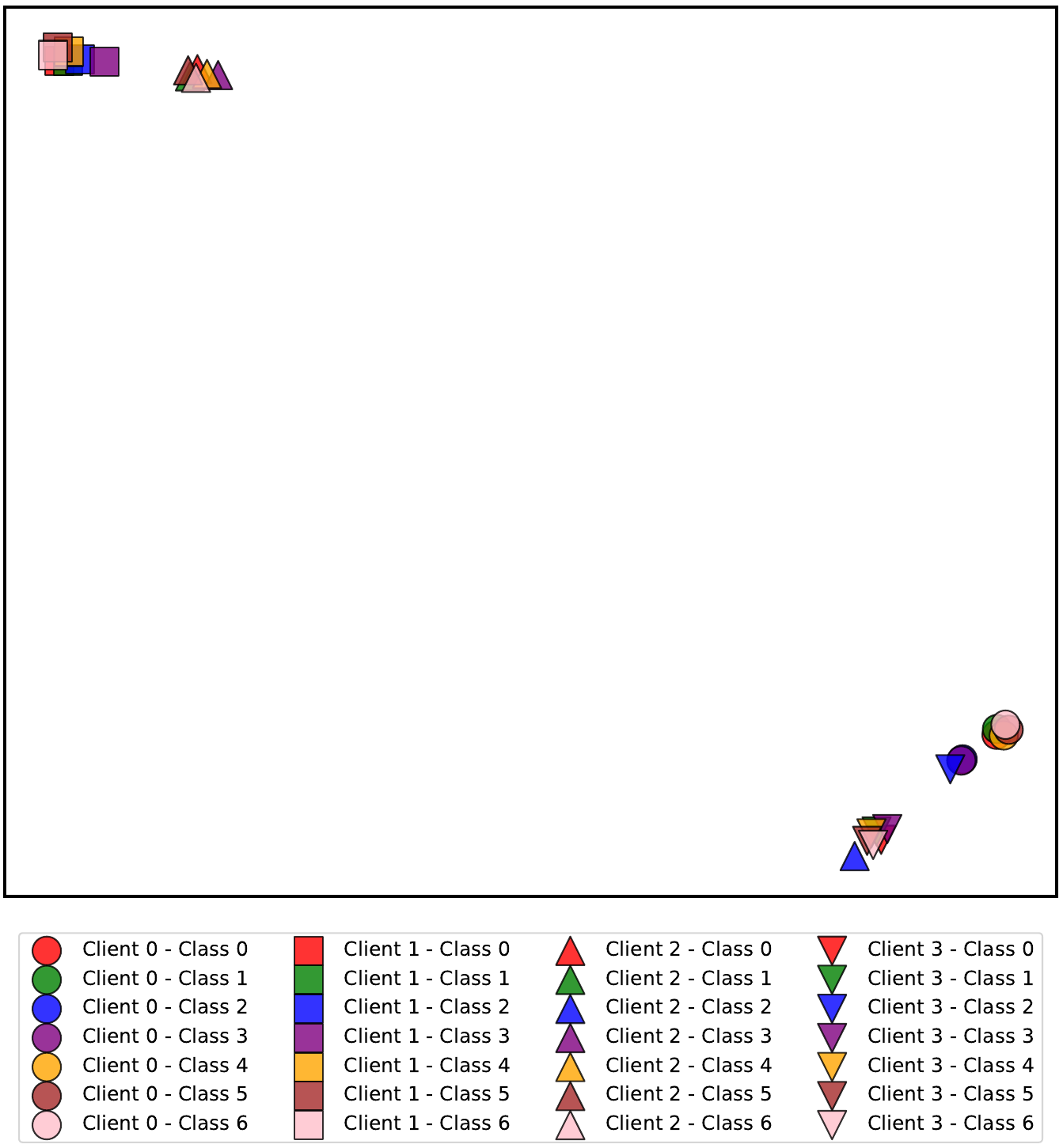}
  }
  \subfigure[Client5]{
    \includegraphics[scale=0.23]{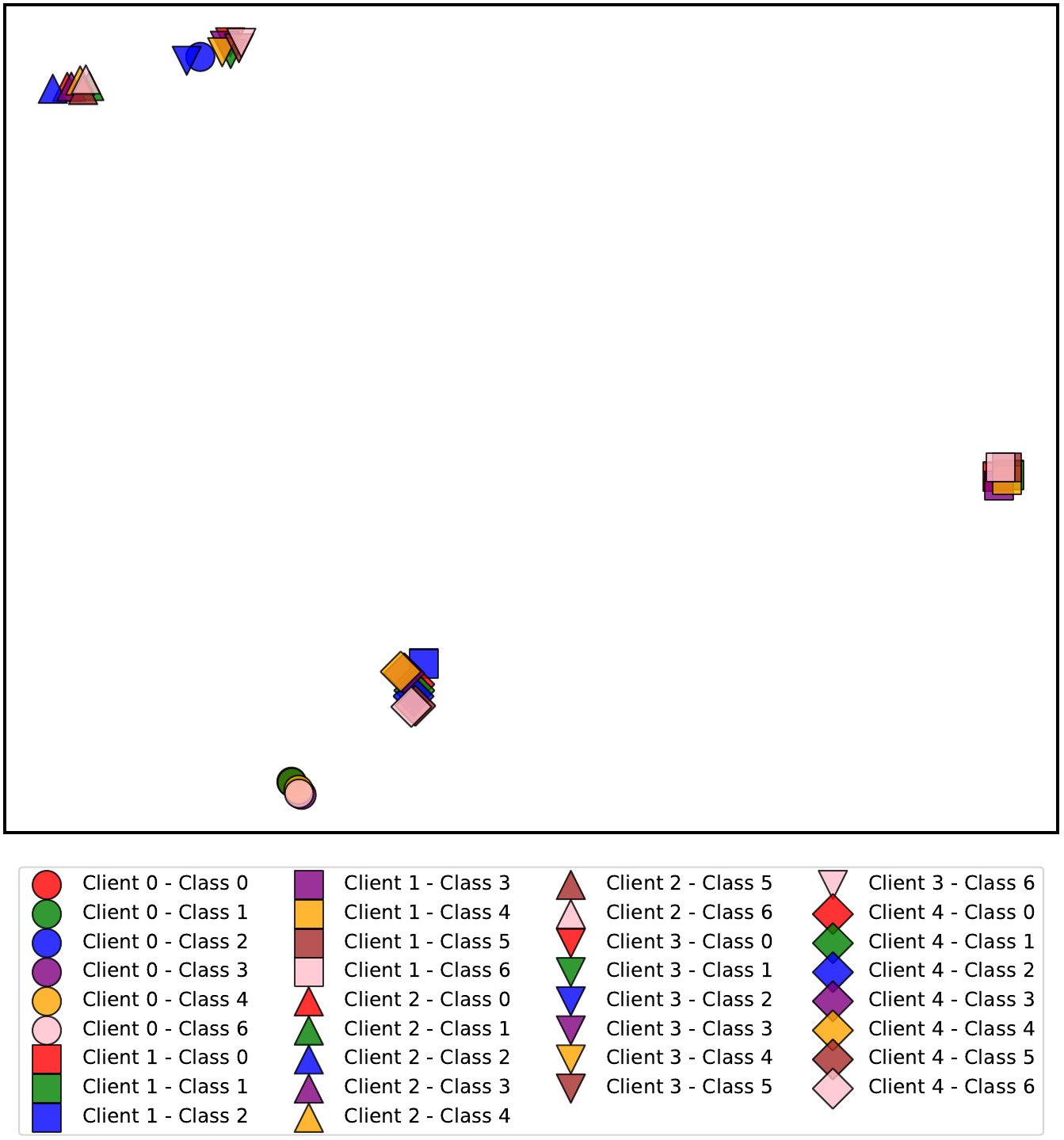}
  }
  \subfigure[Client6]{
    \includegraphics[scale=0.23]{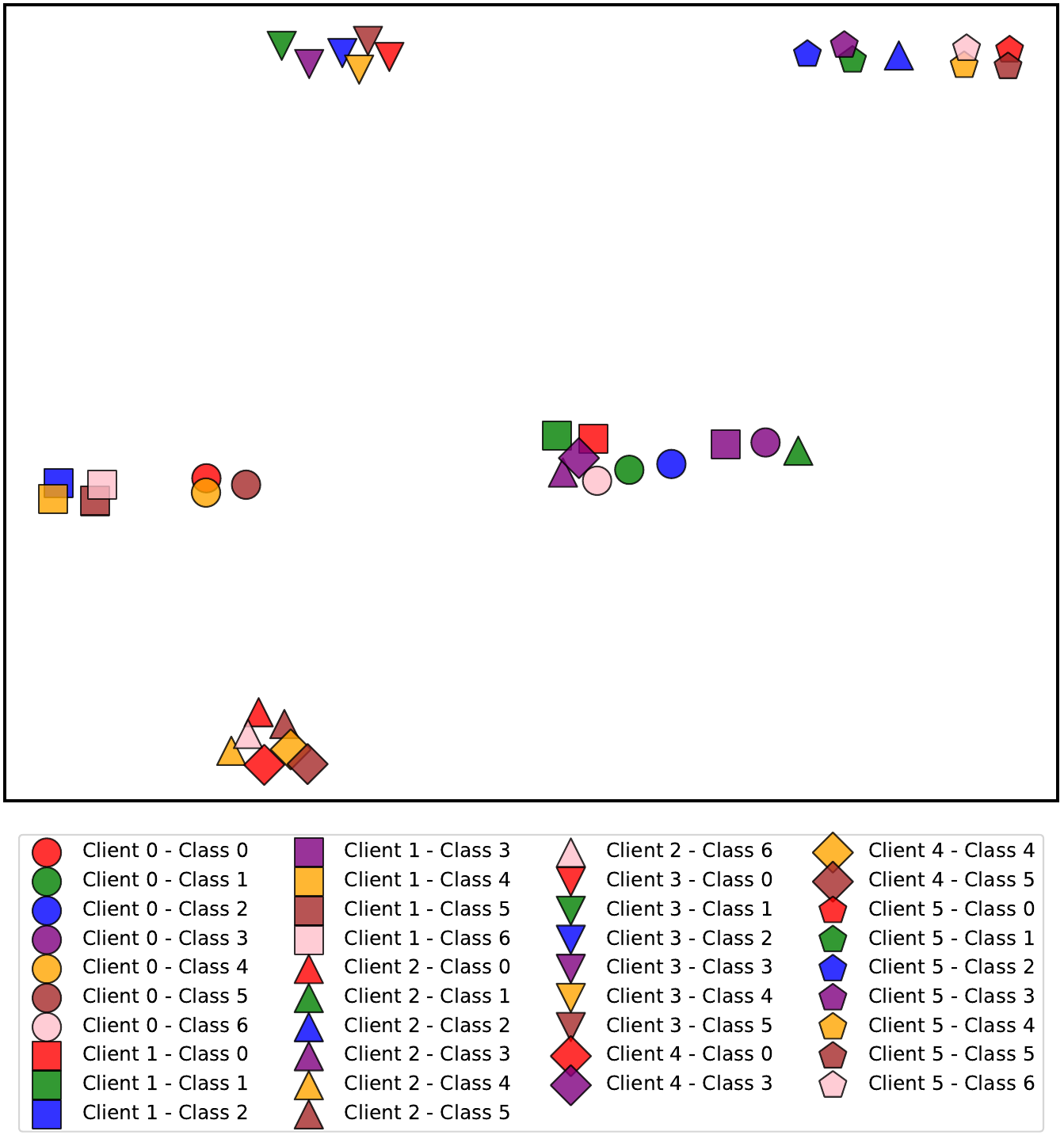}
  }
  \subfigure[Client7]{
    \includegraphics[scale=0.23]{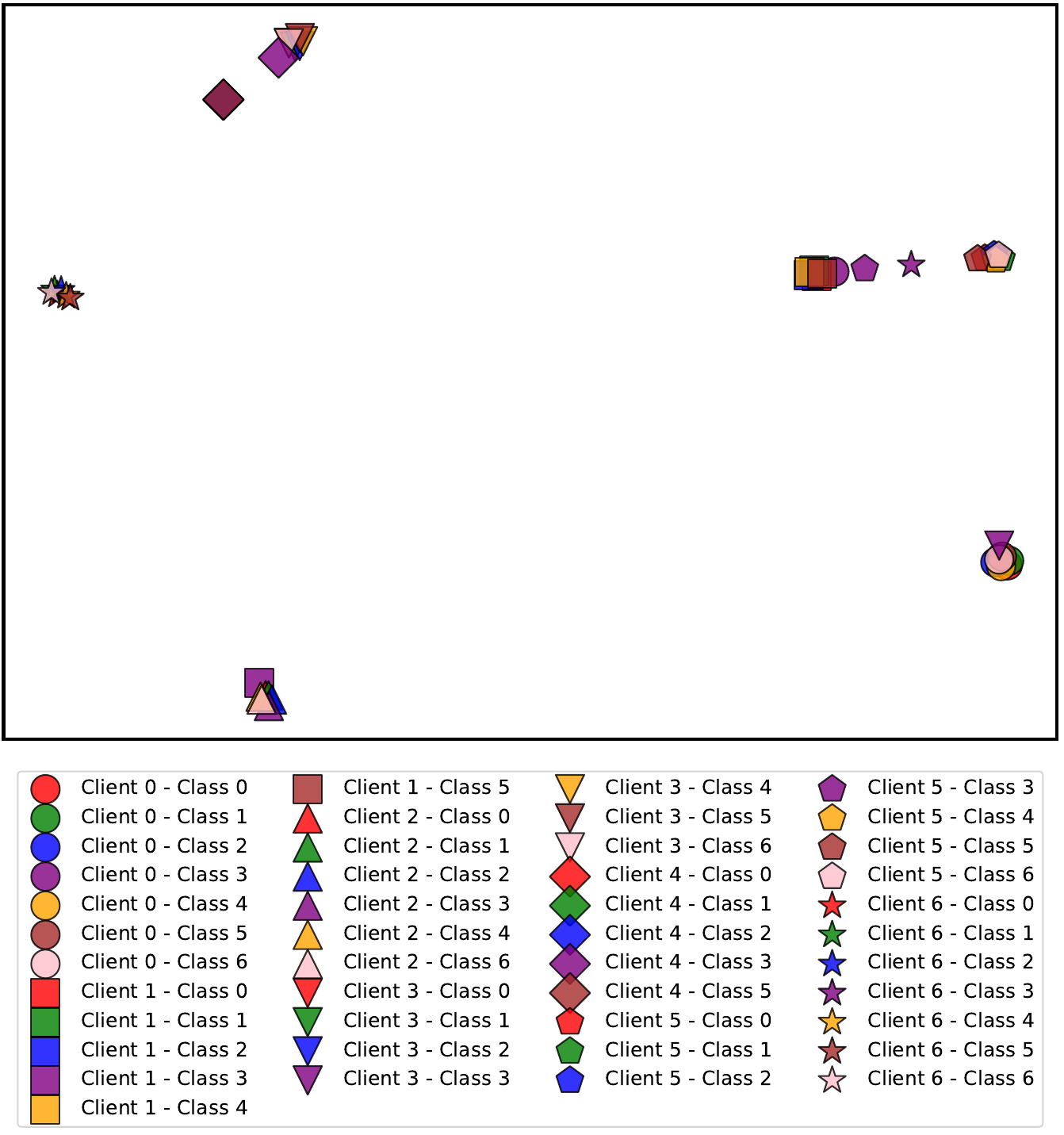}
  }
  \subfigure[Client8]{
    \includegraphics[scale=0.23]{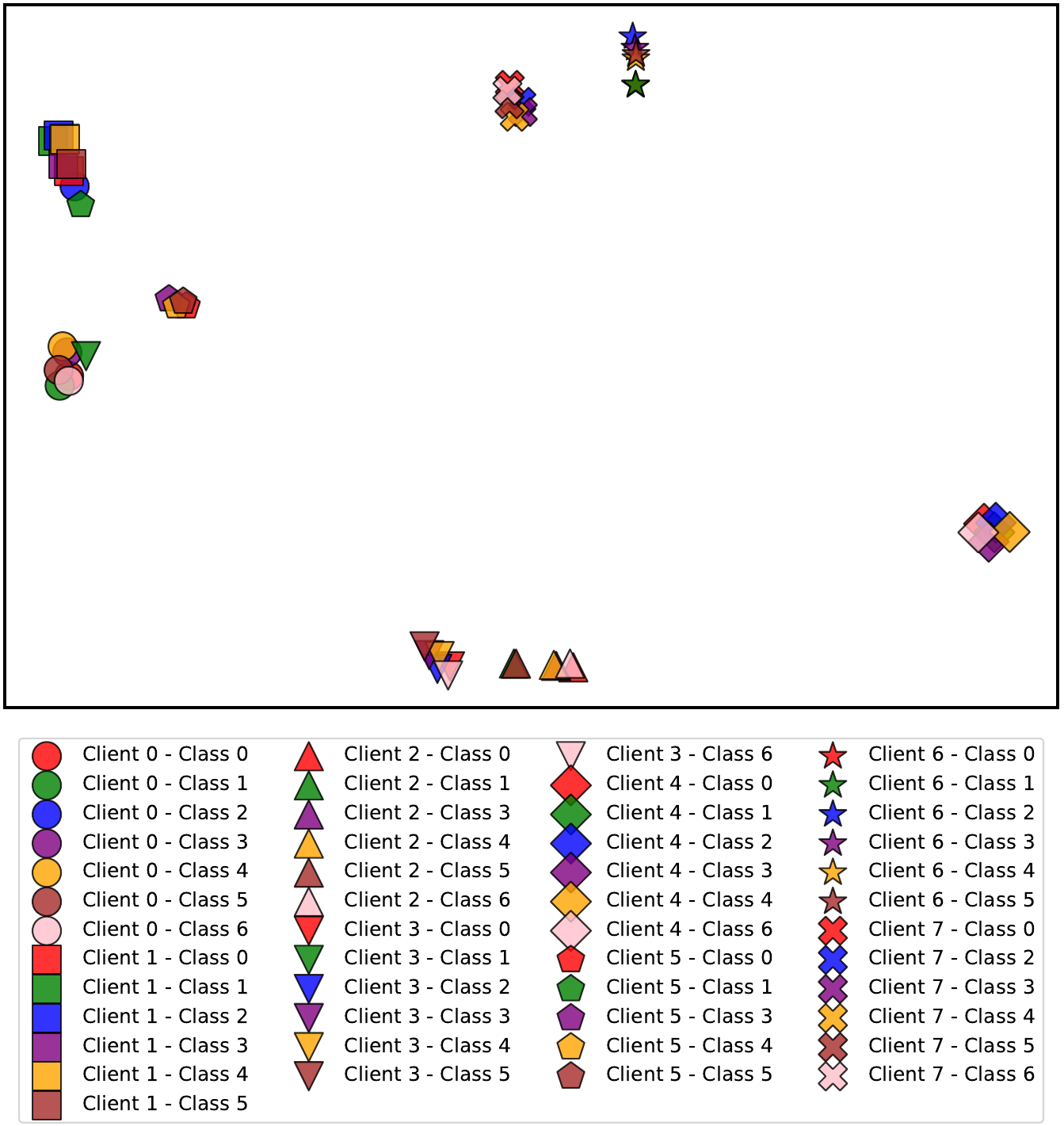}
  }
  \caption{The visualization of each client prototype on Cora generated by \ours.}
  \label{fig_prototype_visualization}
\end{figure}

To demonstrate \ours’s effectiveness in client-data identification attacks, we visualize prototypes for different clients using t-SNE on the Cora dataset based on GCN, reducing prototype dimensions to 2D (Fig~\ref{fig_prototype_visualization}). Prototypes from the same client share the same shape, while those from the same class share the same color. Results with 3–8 clients show that prototypes from the same client cluster tightly, with separability improving as client numbers decrease, simplifying classification. These trends align with results in Section~\ref{client-data-ownership_results}. \ours’s prototypes exhibit high discriminability, enabling reliable client identification by comparing the queried node’s position relative to client prototypes.
\section{Conclusion}
\label{conclusion}

This work sheds light on a previously underexplored privacy risk in federated GNNs: cross-client membership inference attacks. We propose \ours, a novel attack framework that goes beyond traditional sample-level inference by identifying the client origin of individual data samples. By exploiting structural patterns, gradient dynamics, and embedding behaviors during training, \ours effectively links nodes to their source clients. Experimental results across diverse graph datasets confirm the attack’s effectiveness under realistic FL settings. These findings emphasize the urgency of developing privacy-preserving mechanisms that are robust not only to data-level inference but also to client-level attribution threats in FedGNNs. The \textbf{limitations} of \ours are discussed in Appendix~\ref{limitations}.

\bibliographystyle{abbrvnat}
\bibliography{ccmia}

\begin{thebibliography}{62}
\providecommand{\natexlab}[1]{#1}
\providecommand{\url}[1]{\texttt{#1}}
\expandafter\ifx\csname urlstyle\endcsname\relax
  \providecommand{\doi}[1]{doi: #1}\else
  \providecommand{\doi}{doi: \begingroup \urlstyle{rm}\Url}\fi

\bibitem[Anand~Sinha et~al.(2024)Anand~Sinha, Liu, Du, and Shen]{2024gnngradientinversion}
D.~Anand~Sinha, Y.~Liu, R.~Du, and Y.~Shen.
\newblock Gradient inversion attack on graph neural networks.
\newblock \emph{arXiv e-prints}, pages arXiv--2411, 2024.

\bibitem[Bagdasaryan et~al.(2020)Bagdasaryan, Veit, Hua, Estrin, and Shmatikov]{bagdasaryan2020backdoor}
E.~Bagdasaryan, A.~Veit, Y.~Hua, D.~Estrin, and V.~Shmatikov.
\newblock How to backdoor federated learning.
\newblock In \emph{International conference on artificial intelligence and statistics}, pages 2938--2948. PMLR, 2020.

\bibitem[Bai et~al.(2025)Bai, Wu, Zeng, Zhao, Qu, and Yu]{bai2025non}
J.~Bai, D.~Wu, S.~Zeng, Y.~Zhao, Y.~Qu, and S.~Yu.
\newblock Non-iid free federated learning with fuzzy optimization for consumer electronics systems.
\newblock \emph{IEEE Transactions on Consumer Electronics}, 2025.

\bibitem[Bai et~al.(2024)Bai, Hu, Ye, Li, Wang, and Xu]{bai2024membership}
L.~Bai, H.~Hu, Q.~Ye, H.~Li, L.~Wang, and J.~Xu.
\newblock Membership inference attacks and defenses in federated learning: A survey.
\newblock \emph{ACM Computing Surveys}, 57\penalty0 (4):\penalty0 1--35, 2024.

\bibitem[Behnia et~al.(2022)Behnia, Ebrahimi, Pacheco, and Padmanabhan]{2022llmdp}
R.~Behnia, M.~R. Ebrahimi, J.~Pacheco, and B.~Padmanabhan.
\newblock Ew-tune: A framework for privately fine-tuning large language models with differential privacy.
\newblock In \emph{2022 IEEE International Conference on Data Mining Workshops (ICDMW)}, pages 560--566. IEEE, 2022.

\bibitem[Bhagoji et~al.(2019)Bhagoji, Chakraborty, Mittal, and Calo]{bhagoji2019analyzing}
A.~N. Bhagoji, S.~Chakraborty, P.~Mittal, and S.~Calo.
\newblock Analyzing federated learning through an adversarial lens.
\newblock In \emph{International conference on machine learning}, pages 634--643. PMLR, 2019.

\bibitem[Bojchevski and G{\"u}nnemann(2018)]{2018dblp}
A.~Bojchevski and S.~G{\"u}nnemann.
\newblock Deep gaussian embedding of graphs: Unsupervised inductive learning via ranking.
\newblock 2018.

\bibitem[Chatzikokolakis et~al.(2013)Chatzikokolakis, Andr{\'e}s, Bordenabe, and Palamidessi]{2013dx}
K.~Chatzikokolakis, M.~E. Andr{\'e}s, N.~E. Bordenabe, and C.~Palamidessi.
\newblock Broadening the scope of differential privacy using metrics.
\newblock In \emph{international symposium on privacy enhancing technologies symposium}, pages 82--102. Springer, 2013.

\bibitem[Chen et~al.(2024)Chen, Wang, Zhong, Ying, Tang, and Pan]{2024hpmia}
S.~Chen, W.~Wang, Y.~Zhong, Z.~Ying, W.~Tang, and Z.~Pan.
\newblock Hp-mia: A novel membership inference attack scheme for high membership prediction precision.
\newblock \emph{Computers \& Security}, 136:\penalty0 103571, 2024.

\bibitem[Chen et~al.(2025)Chen, He, Qu, and Zhang]{chen2025leapgnn}
W.~Chen, S.~He, H.~Qu, and X.~Zhang.
\newblock $\{$LeapGNN$\}$: Accelerating distributed $\{$GNN$\}$ training leveraging $\{$Feature-Centric$\}$ model migration.
\newblock In \emph{23rd USENIX Conference on File and Storage Technologies (FAST 25)}, pages 255--270, 2025.

\bibitem[Conti et~al.(2022)Conti, Li, Picek, and Xu]{conti2022label}
M.~Conti, J.~Li, S.~Picek, and J.~Xu.
\newblock Label-only membership inference attack against node-level graph neural networks.
\newblock In \emph{Proceedings of the 15th ACM Workshop on Artificial Intelligence and Security}, pages 1--12, 2022.

\bibitem[Gu et~al.(2022)Gu, Bai, and Xu]{2022csmia}
Y.~Gu, Y.~Bai, and S.~Xu.
\newblock Cs-mia: Membership inference attack based on prediction confidence series in federated learning.
\newblock \emph{Journal of Information Security and Applications}, 67:\penalty0 103201, 2022.

\bibitem[Hamilton et~al.(2017)Hamilton, Ying, and Leskovec]{2017graphsage}
W.~Hamilton, Z.~Ying, and J.~Leskovec.
\newblock Inductive representation learning on large graphs.
\newblock \emph{Advances in neural information processing systems}, 30, 2017.

\bibitem[He et~al.(2020)He, Bastani, Jagwani, Park, Abbar, Alizadeh, Balakrishnan, Chawla, Madden, and Sadeghi]{he2020roadtagger}
S.~He, F.~Bastani, S.~Jagwani, E.~Park, S.~Abbar, M.~Alizadeh, H.~Balakrishnan, S.~Chawla, S.~Madden, and M.~A. Sadeghi.
\newblock Roadtagger: Robust road attribute inference with graph neural networks.
\newblock In \emph{Proceedings of the AAAI Conference on Artificial Intelligence}, volume~34, pages 10965--10972, 2020.

\bibitem[He et~al.(2021{\natexlab{a}})He, Wen, Wu, Backes, Shen, and Zhang]{2021nodemia}
X.~He, R.~Wen, Y.~Wu, M.~Backes, Y.~Shen, and Y.~Zhang.
\newblock Node-level membership inference attacks against graph neural networks.
\newblock \emph{arXiv preprint arXiv:2102.05429}, 2021{\natexlab{a}}.

\bibitem[He et~al.(2021{\natexlab{b}})He, Wen, Wu, Backes, Shen, and Zhang]{he2021node}
X.~He, R.~Wen, Y.~Wu, M.~Backes, Y.~Shen, and Y.~Zhang.
\newblock Node-level membership inference attacks against graph neural networks.
\newblock \emph{arXiv preprint arXiv:2102.05429}, 2021{\natexlab{b}}.

\bibitem[Hu et~al.(2021)Hu, Gong, and Guo]{hu2021federated}
R.~Hu, Y.~Gong, and Y.~Guo.
\newblock Federated learning with sparsification-amplified privacy and adaptive optimization.
\newblock In \emph{Proceedings of the Thirtieth International Joint Conference on Artificial Intelligence}, 2021.

\bibitem[Karimireddy et~al.(2020)Karimireddy, Kale, Mohri, Reddi, Stich, and Suresh]{2020scaffold}
S.~P. Karimireddy, S.~Kale, M.~Mohri, S.~Reddi, S.~Stich, and A.~T. Suresh.
\newblock Scaffold: Stochastic controlled averaging for federated learning.
\newblock In \emph{International conference on machine learning}, pages 5132--5143. PMLR, 2020.

\bibitem[Karypis and Kumar(1998)]{1998metis}
G.~Karypis and V.~Kumar.
\newblock A fast and high quality multilevel scheme for partitioning irregular graphs.
\newblock \emph{SIAM J. Sci. Comput.}, 20\penalty0 (1):\penalty0 359–392, Dec. 1998.
\newblock ISSN 1064-8275.

\bibitem[Kipf and Welling(2017)]{2017gcn}
T.~N. Kipf and M.~Welling.
\newblock Semi-supervised classification with graph convolutional networks.
\newblock 2017.

\bibitem[Li et~al.(2020)Li, Sahu, Zaheer, Sanjabi, Talwalkar, and Smith]{2020fedprox}
T.~Li, A.~K. Sahu, M.~Zaheer, M.~Sanjabi, A.~Talwalkar, and V.~Smith.
\newblock Federated optimization in heterogeneous networks.
\newblock \emph{Proceedings of Machine learning and systems}, 2:\penalty0 429--450, 2020.

\bibitem[Li et~al.(2022)Li, Liu, He, Yu, Backes, and Zhang]{li2022auditing}
Z.~Li, Y.~Liu, X.~He, N.~Yu, M.~Backes, and Y.~Zhang.
\newblock Auditing membership leakages of multi-exit networks.
\newblock In \emph{Proceedings of the 2022 ACM SIGSAC Conference on Computer and Communications Security}, pages 1917--1931, 2022.

\bibitem[Lin et~al.(2020)Lin, Kong, Stich, and Jaggi]{2020feddf}
T.~Lin, L.~Kong, S.~U. Stich, and M.~Jaggi.
\newblock Ensemble distillation for robust model fusion in federated learning.
\newblock \emph{Advances in neural information processing systems}, 33:\penalty0 2351--2363, 2020.

\bibitem[Liu et~al.(2025)Liu, Chen, Xue, Guo, and Xu]{liu2025piafgnn}
J.~Liu, B.~Chen, B.~Xue, M.~Guo, and Y.~Xu.
\newblock Piafgnn: Property inference attacks against federated graph neural networks.
\newblock \emph{Computers, Materials \& Continua}, 82\penalty0 (2), 2025.

\bibitem[Liu et~al.(2024)Liu, Xing, Deng, Li, Guan, and Yu]{liu2024federated}
R.~Liu, P.~Xing, Z.~Deng, A.~Li, C.~Guan, and H.~Yu.
\newblock Federated graph neural networks: Overview, techniques, and challenges.
\newblock \emph{IEEE transactions on neural networks and learning systems}, 2024.

\bibitem[Liu et~al.(2023)Liu, Jiang, and Zhu]{liu2023subject}
Y.~Liu, P.~Jiang, and L.~Zhu.
\newblock Subject-level membership inference attack via data augmentation and model discrepancy.
\newblock \emph{IEEE Transactions on Information Forensics and Security}, 18:\penalty0 5848--5859, 2023.

\bibitem[Liu et~al.(2022)Liu, Zhang, Chen, Lin, and Li]{liu2022membership}
Z.~Liu, X.~Zhang, C.~Chen, S.~Lin, and J.~Li.
\newblock Membership inference attacks against robust graph neural network.
\newblock In \emph{International Symposium on Cyberspace Safety and Security}, pages 259--273. Springer, 2022.

\bibitem[Ma et~al.(2025)Ma, Liu, Yan, Cai, Song, Wang, Li, and Cheng]{ma2025adaptive}
K.~Ma, R.~Liu, X.~Yan, Z.~Cai, X.~Song, M.~Wang, Y.~Li, and J.~Cheng.
\newblock Adaptive parallel training for graph neural networks.
\newblock In \emph{Proceedings of the 30th ACM SIGPLAN Annual Symposium on Principles and Practice of Parallel Programming}, pages 29--42, 2025.

\bibitem[McMahan et~al.(2017)McMahan, Moore, Ramage, Hampson, and y~Arcas]{2017fedavg}
B.~McMahan, E.~Moore, D.~Ramage, S.~Hampson, and B.~A. y~Arcas.
\newblock Communication-efficient learning of deep networks from decentralized data.
\newblock In \emph{Artificial intelligence and statistics}, pages 1273--1282. PMLR, 2017.

\bibitem[Melis et~al.(2019)Melis, Song, De~Cristofaro, and Shmatikov]{melis2019exploiting}
L.~Melis, C.~Song, E.~De~Cristofaro, and V.~Shmatikov.
\newblock Exploiting unintended feature leakage in collaborative learning.
\newblock In \emph{2019 IEEE symposium on security and privacy (SP)}, pages 691--706. IEEE, 2019.

\bibitem[Nasr et~al.(2019)Nasr, Shokri, and Houmansadr]{nasr2019comprehensive}
M.~Nasr, R.~Shokri, and A.~Houmansadr.
\newblock Comprehensive privacy analysis of deep learning: Passive and active white-box inference attacks against centralized and federated learning.
\newblock In \emph{2019 IEEE symposium on security and privacy (SP)}, pages 739--753. IEEE, 2019.

\bibitem[Olatunji et~al.(2021{\natexlab{a}})Olatunji, Nejdl, and Khosla]{2021gnnmia}
I.~E. Olatunji, W.~Nejdl, and M.~Khosla.
\newblock Membership inference attack on graph neural networks.
\newblock In \emph{2021 Third IEEE International Conference on Trust, Privacy and Security in Intelligent Systems and Applications (TPS-ISA)}, pages 11--20. IEEE, 2021{\natexlab{a}}.

\bibitem[Olatunji et~al.(2021{\natexlab{b}})Olatunji, Nejdl, and Khosla]{olatunji2021membership}
I.~E. Olatunji, W.~Nejdl, and M.~Khosla.
\newblock Membership inference attack on graph neural networks.
\newblock In \emph{2021 Third IEEE International Conference on Trust, Privacy and Security in Intelligent Systems and Applications (TPS-ISA)}, pages 11--20. IEEE, 2021{\natexlab{b}}.

\bibitem[Park et~al.(2023)Park, Han, Wu, Kim, Zhu, Xie, and Cha]{2023feddefender}
S.~Park, S.~Han, F.~Wu, S.~Kim, B.~Zhu, X.~Xie, and M.~Cha.
\newblock Feddefender: Client-side attack-tolerant federated learning.
\newblock In \emph{Proceedings of the 29th ACM SIGKDD conference on knowledge discovery and data mining}, pages 1850--1861, 2023.

\bibitem[Rao et~al.(2024)Rao, Zhang, Wu, Zhu, Sun, and Chen]{rao2024privacy}
B.~Rao, J.~Zhang, D.~Wu, C.~Zhu, X.~Sun, and B.~Chen.
\newblock Privacy inference attack and defense in centralized and federated learning: A comprehensive survey.
\newblock \emph{IEEE Transactions on Artificial Intelligence}, 2024.

\bibitem[Sajadmanesh and Gatica-Perez(2021)]{sajadmanesh2021locally}
S.~Sajadmanesh and D.~Gatica-Perez.
\newblock Locally private graph neural networks.
\newblock In \emph{Proceedings of the 2021 ACM SIGSAC conference on computer and communications security}, pages 2130--2145, 2021.

\bibitem[Schmierer et~al.(2025)Schmierer, Li, and Di~Wu]{schmierer2025advancing}
T.~Schmierer, T.~Li, and Y.~L. Di~Wu.
\newblock Advancing doa assessment through federated learning: A one-shot pseudo data approach.
\newblock \emph{Neurocomputing}, 634, 2025.

\bibitem[Shchur et~al.(2018)Shchur, Mumme, Bojchevski, and G{\"u}nnemann]{2018pitfall}
O.~Shchur, M.~Mumme, A.~Bojchevski, and S.~G{\"u}nnemann.
\newblock Pitfalls of graph neural network evaluation.
\newblock In \emph{Relational Representation Learning Workshop, NeurIPS 2018}, 2018.

\bibitem[Sui et~al.(2023)Sui, Sun, Zhang, Chen, and Li]{2023ganenhance}
H.~Sui, X.~Sun, J.~Zhang, B.~Chen, and W.~Li.
\newblock Multi-level membership inference attacks in federated learning based on active gan.
\newblock \emph{Neural Computing and Applications}, 35\penalty0 (23):\penalty0 17013--17027, 2023.

\bibitem[Sun et~al.(2024)Sun, Liu, Cui, Liu, Ma, and Liu]{2024client}
Y.~Sun, Z.~Liu, J.~Cui, J.~Liu, K.~Ma, and J.~Liu.
\newblock Client-side gradient inversion attack in federated learning using secure aggregation.
\newblock \emph{IEEE Internet of Things Journal}, 2024.

\bibitem[Velickovic et~al.(2018)Velickovic, Cucurull, Casanova, Romero, Lio, and Bengio]{gat2018}
P.~Velickovic, G.~Cucurull, A.~Casanova, A.~Romero, P.~Lio, and Y.~Bengio.
\newblock Graph attention networks.
\newblock 2018.

\bibitem[Wang et~al.(2020)Wang, Liu, Liang, Joshi, and Poor]{2020fednova}
J.~Wang, Q.~Liu, H.~Liang, G.~Joshi, and H.~V. Poor.
\newblock Tackling the objective inconsistency problem in heterogeneous federated optimization.
\newblock \emph{Advances in neural information processing systems}, 33:\penalty0 7611--7623, 2020.

\bibitem[Wang et~al.(2025)Wang, Li, Han, Ye, and Zhang]{wang2025fast}
P.~Wang, S.~Li, Y.~Han, F.~Ye, and Q.~Zhang.
\newblock Fast-response edge caching scheme for graph data.
\newblock \emph{IEEE Transactions on Networking}, 2025.

\bibitem[Wang et~al.(2019)Wang, Song, Zhang, Song, Wang, and Qi]{wang2019beyond}
Z.~Wang, M.~Song, Z.~Zhang, Y.~Song, Q.~Wang, and H.~Qi.
\newblock Beyond inferring class representatives: User-level privacy leakage from federated learning.
\newblock In \emph{IEEE INFOCOM 2019-IEEE conference on computer communications}, pages 2512--2520. IEEE, 2019.

\bibitem[Wu et~al.(2021)Wu, Yang, Pan, and Yuan]{wu2021adapting}
B.~Wu, X.~Yang, S.~Pan, and X.~Yuan.
\newblock Adapting membership inference attacks to gnn for graph classification: Approaches and implications.
\newblock In \emph{2021 IEEE International Conference on Data Mining (ICDM)}, pages 1421--1426. IEEE, 2021.

\bibitem[Wu et~al.(2024)Wu, Bai, Song, Chen, Zhou, Xiang, and Sajjanhar]{wu2024fedinverse}
D.~Wu, J.~Bai, Y.~Song, J.~Chen, W.~Zhou, Y.~Xiang, and A.~Sajjanhar.
\newblock Fedinverse: Evaluating privacy leakage in federated learning.
\newblock In \emph{The twelfth international conference on learning representations}, 2024.

\bibitem[Wu et~al.(2025)Wu, Fang, Li, Yuan, Wei, Nan, and Tao]{wu2025secret}
H.~Wu, Y.~Fang, N.~Li, X.~Yuan, Z.~Wei, G.~Nan, and X.~Tao.
\newblock Secret key generation with untrusted internal eavesdropper: Token-based anti-eavesdropping.
\newblock \emph{IEEE Transactions on Information Forensics and Security}, 2025.

\bibitem[Wu et~al.(2020)Wu, Pan, Chen, Long, Zhang, and Yu]{wu2020comprehensive}
Z.~Wu, S.~Pan, F.~Chen, G.~Long, C.~Zhang, and P.~S. Yu.
\newblock A comprehensive survey on graph neural networks.
\newblock \emph{IEEE transactions on neural networks and learning systems}, 32\penalty0 (1):\penalty0 4--24, 2020.

\bibitem[Xie et~al.(2021{\natexlab{a}})Xie, Ma, Xiong, and Yang]{2021federated_noniid}
H.~Xie, J.~Ma, L.~Xiong, and C.~Yang.
\newblock Federated graph classification over non-iid graphs.
\newblock \emph{Advances in neural information processing systems}, 34:\penalty0 18839--18852, 2021{\natexlab{a}}.

\bibitem[Xie et~al.(2021{\natexlab{b}})Xie, Chen, Zhang, and Wu]{xie2021defending}
Y.~Xie, B.~Chen, J.~Zhang, and D.~Wu.
\newblock Defending against membership inference attacks in federated learning via adversarial example.
\newblock In \emph{2021 17th International Conference on Mobility, Sensing and Networking (MSN)}, pages 153--160. IEEE, 2021{\natexlab{b}}.

\bibitem[Xu et~al.(2024)Xu, Yin, Fang, and Gong]{xu2024robust}
Y.~Xu, M.~Yin, M.~Fang, and N.~Z. Gong.
\newblock Robust federated learning mitigates client-side training data distribution inference attacks.
\newblock In \emph{Companion Proceedings of the ACM Web Conference 2024}, pages 798--801, 2024.

\bibitem[Yan et~al.(2022)Yan, Li, Wang, Zhang, Sharif, Hu, and Li]{yan2022membership}
H.~Yan, S.~Li, Y.~Wang, Y.~Zhang, K.~Sharif, H.~Hu, and Y.~Li.
\newblock Membership inference attacks against deep learning models via logits distribution.
\newblock \emph{IEEE Transactions on Dependable and Secure Computing}, 20\penalty0 (5):\penalty0 3799--3808, 2022.

\bibitem[Yang et~al.(2023)Yang, Shao, Yang, Liu, Liu, Xia, Schaefer, and Fang]{2023watermarking}
W.~Yang, S.~Shao, Y.~Yang, X.~Liu, X.~Liu, Z.~Xia, G.~Schaefer, and H.~Fang.
\newblock Watermarking in secure federated learning: A verification framework based on client-side backdooring.
\newblock \emph{ACM Transactions on Intelligent Systems and Technology}, 15\penalty0 (1):\penalty0 1--25, 2023.

\bibitem[Yang et~al.(2016)Yang, Cohen, and Salakhudinov]{2016plaintoid}
Z.~Yang, W.~Cohen, and R.~Salakhudinov.
\newblock Revisiting semi-supervised learning with graph embeddings.
\newblock In \emph{ICML}, pages 40--48. PMLR, 2016.

\bibitem[Zhang et~al.(2022{\natexlab{a}})Zhang, Liu, Zhu, Ding, and Zhou]{zhang2022label}
G.~Zhang, B.~Liu, T.~Zhu, M.~Ding, and W.~Zhou.
\newblock Label-only membership inference attacks and defenses in semantic segmentation models.
\newblock \emph{IEEE Transactions on Dependable and Secure Computing}, 20\penalty0 (2):\penalty0 1435--1449, 2022{\natexlab{a}}.

\bibitem[Zhang et~al.(2021{\natexlab{a}})Zhang, Zhu, Di~Wu, Yong, and Long]{zhang2021badfss}
J.~Zhang, C.~Zhu, X.~S. Di~Wu, J.~Yong, and G.~Long.
\newblock Badfss: backdoor attacks on federated self-supervised learning.
\newblock In \emph{Proceedings of the 33rd International Joint Conference on Artificial Intelligence (IJCAI)}, 2021{\natexlab{a}}.

\bibitem[Zhang et~al.(2021{\natexlab{b}})Zhang, Liu, Huang, Wang, Lu, Liu, and Chen]{2021graphmi}
Z.~Zhang, Q.~Liu, Z.~Huang, H.~Wang, C.~Lu, C.~Liu, and E.~Chen.
\newblock Graphmi: Extracting private graph data from graph neural networks.
\newblock In \emph{Proceedings of the Thirtieth International Joint Conference on Artificial Intelligence, {IJCAI-21}}, pages 3749--3755, 8 2021{\natexlab{b}}.
\newblock Main Track.

\bibitem[Zhang et~al.(2022{\natexlab{b}})Zhang, Chen, Backes, Shen, and Zhang]{zhang2022inference}
Z.~Zhang, M.~Chen, M.~Backes, Y.~Shen, and Y.~Zhang.
\newblock Inference attacks against graph neural networks.
\newblock In \emph{31st USENIX Security Symposium (USENIX Security 22)}, pages 4543--4560, 2022{\natexlab{b}}.

\bibitem[Zhao et~al.(2020{\natexlab{a}})Zhao, Mopuri, and Bilen]{2020idlg}
B.~Zhao, K.~R. Mopuri, and H.~Bilen.
\newblock idlg: Improved deep leakage from gradients.
\newblock \emph{arXiv preprint arXiv:2001.02610}, 2020{\natexlab{a}}.

\bibitem[Zhao et~al.(2020{\natexlab{b}})Zhao, Chen, Zhang, Wu, Teng, and Yu]{zhao2020pdgan}
Y.~Zhao, J.~Chen, J.~Zhang, D.~Wu, J.~Teng, and S.~Yu.
\newblock Pdgan: A novel poisoning defense method in federated learning using generative adversarial network.
\newblock In \emph{Algorithms and Architectures for Parallel Processing: 19th International Conference, ICA3PP 2019, Melbourne, VIC, Australia, December 9--11, 2019, Proceedings, Part I 19}, pages 595--609. Springer, 2020{\natexlab{b}}.

\bibitem[Zhu et~al.(2025)Zhu, Reganti, Huang, Dickens, Rao, Subbian, and Koutra]{zhu2025simplifying}
J.~Zhu, A.~Reganti, E.~W. Huang, C.~Dickens, N.~Rao, K.~Subbian, and D.~Koutra.
\newblock Simplifying distributed neural network training on massive graphs: Randomized partitions improve model aggregation.
\newblock \emph{ACM Transactions on Knowledge Discovery from Data}, 19\penalty0 (1):\penalty0 1--26, 2025.

\bibitem[Zhu et~al.(2019)Zhu, Liu, and Han]{zhu2019deep}
L.~Zhu, Z.~Liu, and S.~Han.
\newblock Deep leakage from gradients.
\newblock \emph{Advances in neural information processing systems}, 32, 2019.

\end{thebibliography}

\appendix
\section{Technical Appendices and Supplementary Material}

\subsection{Motivation}
\label{motivation}
Most existing GNN-specific attacks are confined to centralized settings, where the assumption of full graph access significantly simplifies the attacker's task. However, the advent of federated GNNs introduces enhanced privacy protection mechanisms that render traditional centralized GNN MIA methods ineffective. This limitation arises because federated frameworks distribute disjoint and heterogeneous subgraphs among clients. Attackers, treated as equal participants within this framework, face significant constraints: they are unable to access similar shadow subgraphs from other clients and lack the posterior probabilities of other clients' GNN models, which are critical for training conventional MIA classifiers. These challenges give rise to the first question:
\textit{Is it feasible to perform cross-client membership inference without any knowledge of the GNN models employed by other clients and without access to their subgraphs?}

While prior research on MIAs in federated learning has primarily focused on sample-level membership inference within training datasets, the specific problem of node-level ownership inference in federated GNNs remains underexplored. Node-level MIAs differ fundamentally from graph-level attacks, which bear similarities to traditional sample-level attacks (\textit{e.g.}, those targeting images or text). The granular nature of nodes as integral components of a graph introduces unique complexities not present in graph-level analyses.

As detailed in Appendix~\ref{dataset_statistics}, the subgraph structures held by individual clients in federated GNNs are typically non-i.i.d., reflecting the inherent heterogeneity induced by graph partitioning algorithms. These algorithms often result in imbalanced class distributions across clients~\cite{2021federated_noniid}. This observation motivates the second research question:
\textit{Can an attacker, operating as an equal client within the federated framework, perform node-level client ownership inference?}

We have solved these problems with our proposed \ours in Section~\ref{methodology}
\subsection{Algorithms}
We present the algorithmic workflows for Gradient Inversion and Prototype-Based Client Matching in Subsection~\ref{client-data_indentification}, detailed in Algorithm~\ref{grad_inversion_alg} and Algorithm~\ref{prot_match_alg}, respectively.
\begin{algorithm}
    \SetKwInput{KwInput}{Input}
    \SetKwInput{KwOutput}{Output}
    \KwInput{
      Global GNN $\mathcal{F}(X, A; W)$;
      Intercepted client gradients $\nabla_W \mathcal{L}$;
      Ground‑truth labels $Y$.
    }
    \KwOutput{Reconstructed client data $(\hat X, \hat A)$.}
    Initialize dummy node features $\hat X^{(1)}$ and adjacency matrix $\hat A^{(1)}$\;
    \For {epoch $e = 1$ \KwTo $T$}
        {
            Compute synthetic gradient: $\nabla_W \hat{\mathcal{L}} = \partial \mathcal{L}(F(\hat{X}^e, \hat{A}^e, W), \text{Y}) / \partial W$\;
            Evaluate gradient‑fitting loss $\hat{\mathbb{L}}^{(e)}$ (Eq.~\ref{grad_loss_final})\;
            Gradient descent for dummy node features: $\hat{X}^{e + 1} = \hat{X}^e - \eta_x \nabla_{\hat{X}^e} \hat{\mathbb{L}}^e$\;
            Gradient descent for dummy adjacency: $\hat{A}^{e+1} = \text{proj}_{[0,1]}(\hat{A}^e - \eta \nabla_{\hat{A}^e} \hat{\mathbb{L}}^e)$
        }
    Obtain $\hat{A}$ by sampling $\hat{A}^{e+1}$ (Eq.~\ref{sample_edges})\;
    return $\hat{X} = \hat{X}^{e+1}, \hat{A}$
    \caption{Workflow of Gradient Inversion.}
    \label{grad_inversion_alg}
\end{algorithm}

\begin{algorithm}
\caption{Workflow of Prototype-based Client Matching}
\label{prot_match_alg}

\begin{multicols}{2}
\KwIn{Global GNN model $\mathcal{F}(X, A; W)$, reconstructed subgraphs $\{(\hat{X}_k,\hat{A}_k)\}_{k=1}^K$, real graph $(X,A)$}
\KwOut{Client assignment $\{\hat{k}_i\}_{i=1}^N$ for each node $x_i\in X$}

\BlankLine
\For{$k \leftarrow 1$ \KwTo $K$}{
    $\hat{\mathcal{E}}_k \leftarrow \mathcal{F}_1(\hat{X}_k, \hat{A}_k)$\;
    \For{each class $c \in \mathcal{C}_k$}{
        $\mu_c^{(k)} \leftarrow \dfrac{1}{|\mathcal{I}^k_c|}\sum_{j \in \mathcal{I}^k_c}\hat{\mathcal{E}}_k^j$
    }
    $\mathcal{P}_k \leftarrow \{\mu_c^{(k)} \mid c\in\mathcal{C}_k\}$\quad // Eq.~\ref{obtain_prototype}
}
\BlankLine
$\mathcal{E} \leftarrow \mathcal{F}_1(X, A)$\;
\For{$i \leftarrow 1$ \KwTo $N$}{
    \For{$k \leftarrow 1$ \KwTo $K$}{
        \eIf{$y_i \in \mathcal{C}_k$}{
            $d_{i,k} \leftarrow 1 - \displaystyle\frac{\mathcal{E}_i \cdot \mu^{(k)}_{y_i}}{\|\mathcal{E}_i\|\;\|\mu^{(k)}_{y_i}\|}$ \quad // Eq.~\ref{compute_distance}\;
        }{
            $d_{i,k} \leftarrow \infty$\;
        }
    }
    $\hat{k}_i \leftarrow \arg\min_{k} d_{i,k}$
}
\Return{$\{\hat{k}_i\}_{i=1}^N$}

\end{multicols}
\end{algorithm}

\subsection{Complexity}
\label{complexity}
We analyse the time complexity and space complexity of the membership inference and client-data identification tasks of \ours, respectively.
\subsubsection{Membership Inference}

The complexity of membership inference primarily arises from two stages: global GNN inference and MLP classification.

For an $L$-layer GNN with $N$ nodes and feature dimensions of $D$,  the feature transformation and neighborhood aggregation steps have a time complexity of $O\left(L(ND^2 + |E|D)\right)$. Memory usage includes storing the adjacency matrix ($O(|E|)$), node embeddings ($O(ND)$), and model weights ($O(LD^2)$), resulting in a space complexity of $O\left(|E| + ND + LD^2\right)$.

For a $J$-layer MLP with width $O(D)$, the computation cost per node is $O(JD^2)$. Processing $N$ nodes leads to a total time complexity of $O(NJD^2)$. The space complexity is dominated by storing parameters, requiring $O(JD^2)$ memory.

Overall, membership inference is dominated by the GNN inference. For large graphs ($|E| \gg N$), the sparse aggregation $O(L|E|D)$ becomes the primary bottleneck. If the embedding dimension $D$ is large, the dense transformation term $O(LND^2)$ also contributes significantly. The MLP, with typically small $J$, incurs relatively minor costs. Thus, the overall time and space complexities are approximately $\mathbf{O\left(L(ND^2 + |E|D)\right)}$ and $\mathbf{O\left(|E| + ND + LD^2\right)}$.

\subsubsection{Client-data Identification}
The main computational overhead of client-data identification lies in gradient inversion and GNN feature extraction. Each optimization iteration involves a forward and backward pass on an $n$-node subgraph with an $L$-layer GNN. The time complexity for a single layer is $O(N \cdot D^2 + |E| \cdot D)$, where $N$ is the number of nodes, $D$ is the embedding dimension, and $|E|$ is the number of edges. For $T$ steps of gradient descent, the total time complexity of gradient inversion is approximately: $O(T \cdot L \cdot (N \cdot D^2 + |E| \cdot D)) \approx O(T \cdot L \cdot N \cdot D^2)$, assuming $|E| = O(N)$ for sparse graphs. Memory consumption in this stage is dominated by storing the reconstructed inputs, which include the adjacency matrix ($O(N^2)$) and node features ($O(N \cdot D)$), resulting in a total space complexity of $O(N^2 + N \cdot D)$.

The feature extraction involves a single forward pass of the global GNN on the reconstructed subgraph. The time complexity is: $O(L \cdot (N \cdot D^2 + |E| \cdot D)) \approx O(L \cdot N \cdot D^2)$, with space complexity of $O(N \cdot D + |E|)$.

The prototype matching is the least computationally intensive. The $D$-dimensional embedding of the target node is compared with $K \cdot C$ prototypes, where $K$ is the number of clients and $C$ is the number of classes. The time and space complexity of this stage are both $O(K \cdot C \cdot D)$.

To summarize, the overall complexity of client-data identification is primarily determined by gradient inversion, with a time complexity of $\mathbf{O(T \cdot L \cdot N \cdot D^2)}$ and a space complexity of $\mathbf{O(N^2 + N \cdot D)}$.

\subsection{Limitations}
\label{limitations}
The \textbf{limitations} of \ours include the degradation of gradient inversion quality as the number of clients increases, indicating scalability issues in reconstructing accurate subgraph structures under large-scale federated settings. Moreover, \ours assumes class-consistent gradients and prototype separability for effective ownership inference, which may not hold under real-world non-IID client distributions.

\subsection{Federated Learning Approaches}
\label{approaches}
We evaluate the proposed \ours under various federated approaches to demonstrate its generalizability in federated GNN scenarios. The specific methods include:  
\begin{itemize}  
\item \textbf{FedAvg}~\cite{2017fedavg} is the foundational algorithm for federated learning, where clients perform local training, and the server aggregates model parameters via weighted averaging. The weights are typically proportional to the data volume of each client relative to the total dataset.  

\item \textbf{FedProx}~\cite{2020fedprox} is an extension of FedAvg that introduces a regularization term in the client loss function to enforce consistency with the global model, thereby aligning local updates with the global objective and accelerating convergence.  

\item \textbf{SCAFFOLD}~\cite{2020scaffold} incorporates control variates to correct gradient deviations caused by data heterogeneity. By coordinating global and local control variates, it mitigates client drift and improves convergence in non-IID settings.  

\item \textbf{FedDF}~\cite{2020feddf} is a robust distillation-based framework for federated model fusion. It supports heterogeneity in client models, data distributions, and neural architectures, providing flexibility and resilience in federated learning setups.  

\item \textbf{FedNova}~\cite{2020fednova} is an enhancement of FedAvg that dynamically adjusts aggregation weights by considering client-specific local iteration counts and data volumes. This approach addresses the challenges posed by heterogeneous data and varying local training steps.  
\end{itemize}

\subsection{Baselines}
\label{baselines}
We set the following strong baselines in the membership inference.
\begin{itemize}
    \item \textbf{HP-MIA}~\cite{2024hpmia} trains the membership inference model by comparing the differences in feature extraction between the server and the attacker using the data owned by the attacker. It leverages the phenomenon of model overfitting to distinguish training-set data from non-members.
    \item \textbf{GAN Based Data Enhancement}~\cite{2023ganenhance} utilizes generative adversarial networks (GANs) to augment data. By generating additional synthetic data that aligns with the attacker's data distribution, it enhances the training dataset used to train the membership inference model.
    \item \textbf{CS-MIA}~\cite{2022csmia} is a membership inference attack method designed for federated learning settings. It employs the confidence sequences from the model at each round on the attacker's local data to train the membership inference model.
\end{itemize}

\subsection{Metrics}
\label{metrics}
We explain the evaluation metrics used in this study. For membership inferences, AUC mitigates bias from varying training-set proportions. Client-data identification uses accuracy for classification of evenly divided nodes. Subgraph reconstruction via gradient inversion is evaluated using AUC for graph structure and RNMSE for node features.
\begin{itemize}
    \item \textbf{AUC (Area Under Curve)}: The area under the Reciever Operator Characteristic Curve.
    \item The \textbf{accuracy} of the client-data identificationis formalized as: \begin{equation}
        ACC_{\text{Client-data Identification}} = \frac{\sum^{N}_{i=1}(\hat{k}_i == k_i)} {N},
    \end{equation}
    where $k_i$ denotes the true client that node $i$ belongs to.
    \item The \textbf{Root Normalised Mean Squared Error(RNMSE)} is formalized as:
    \begin{equation}
        RNMSE(x_v, \hat{x}_v) = \frac{\|x_v - \hat{x}_v\|}{\|x_v\|}.
    \end{equation}
\end{itemize}

\subsection{Data Statistics}
\label{dataset_statistics}
\begin{figure}
  \centering
  \subfigure[Cora-Client1]{
    \includegraphics[scale=0.25]{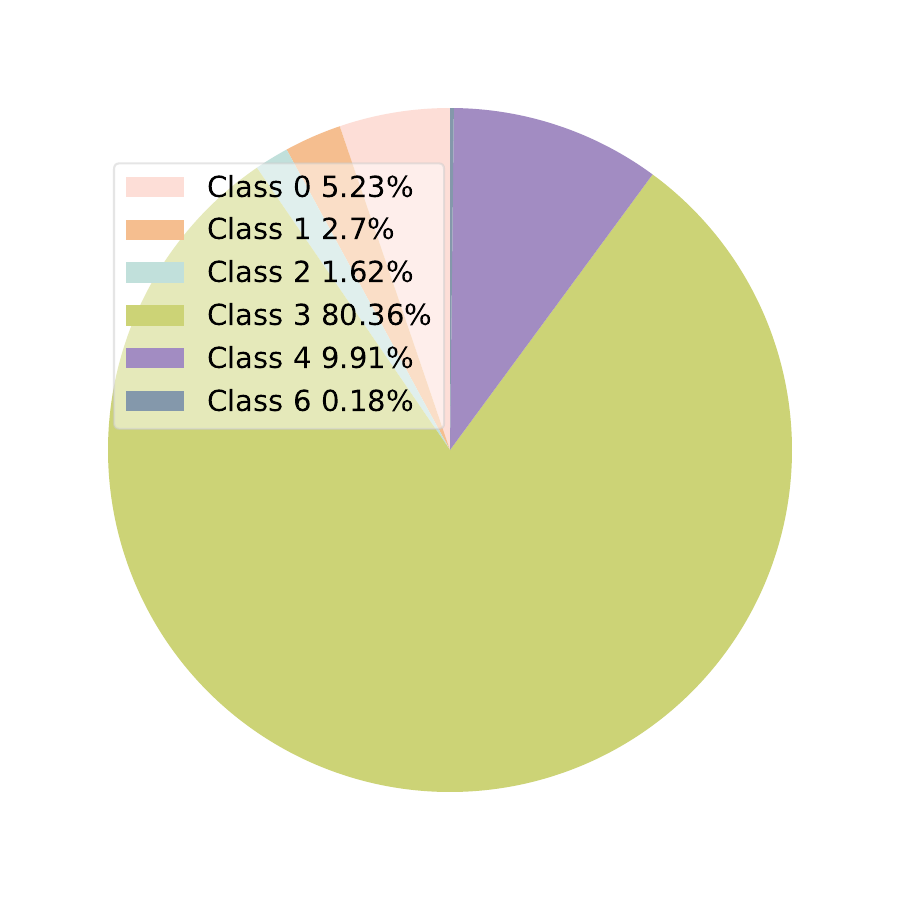}
  }
  \subfigure[Cora-Client2]{
    \includegraphics[scale=0.25]{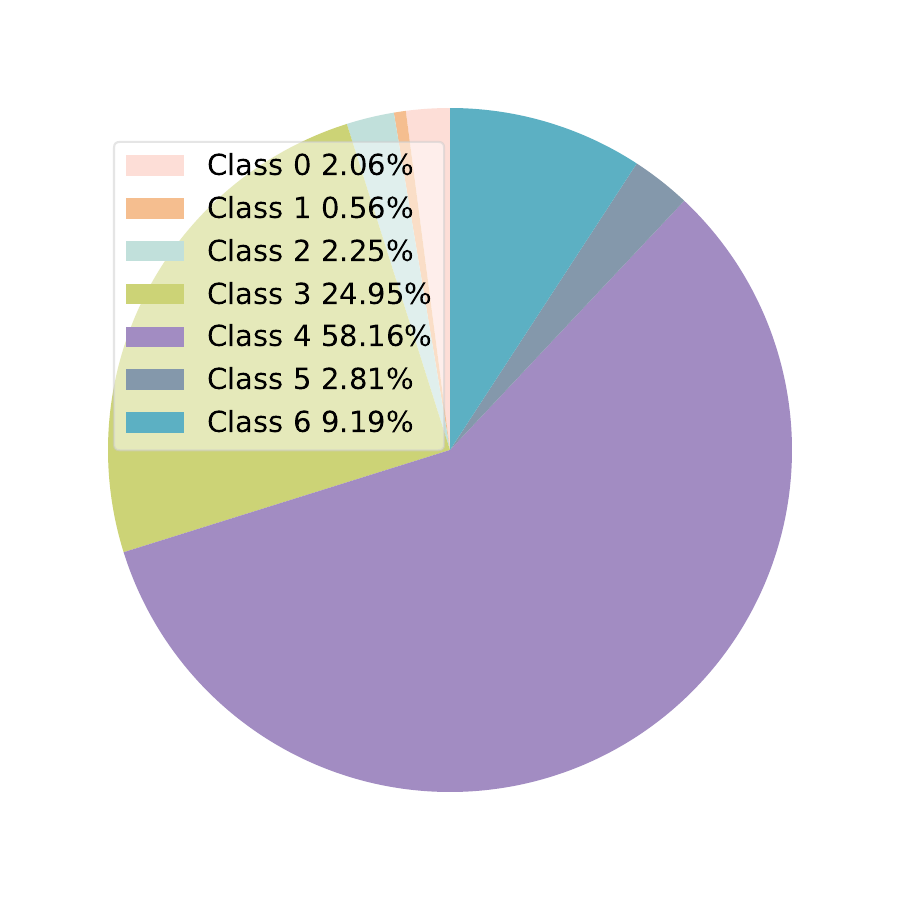}
  }
  \subfigure[Cora-Client3]{
    \includegraphics[scale=0.25]{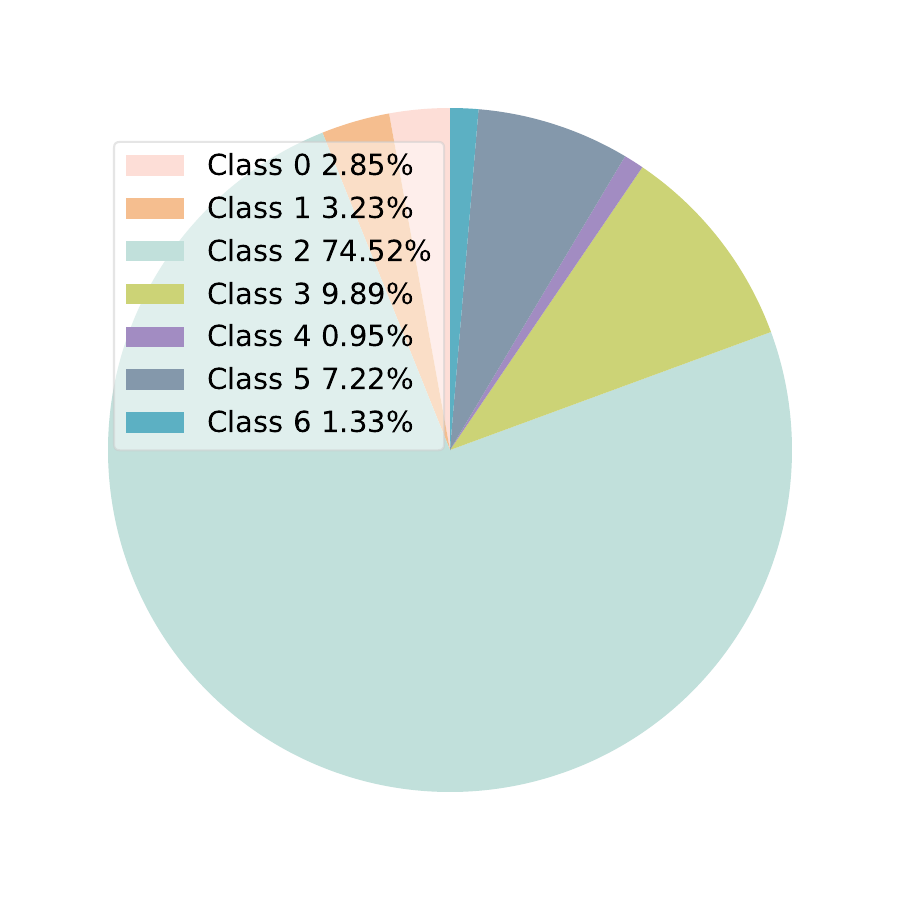}
  }
  \subfigure[Cora-Client4]{
    \includegraphics[scale=0.25]{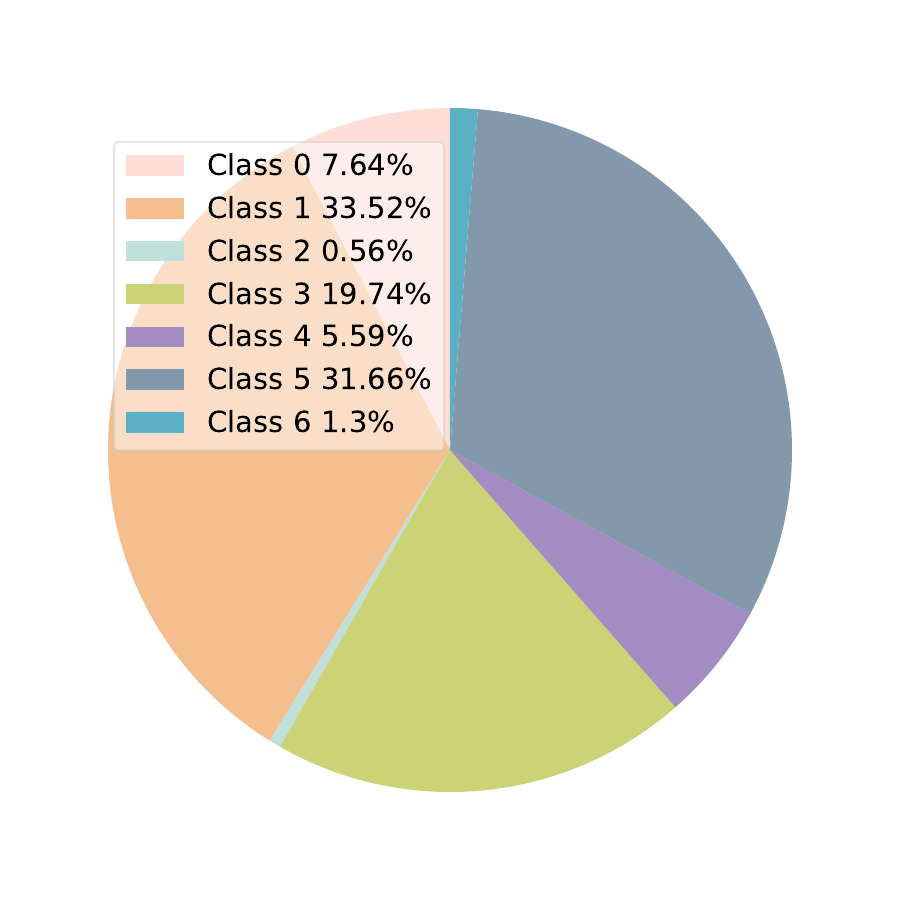}
  }
  \subfigure[Cora-Client5]{
    \includegraphics[scale=0.25]{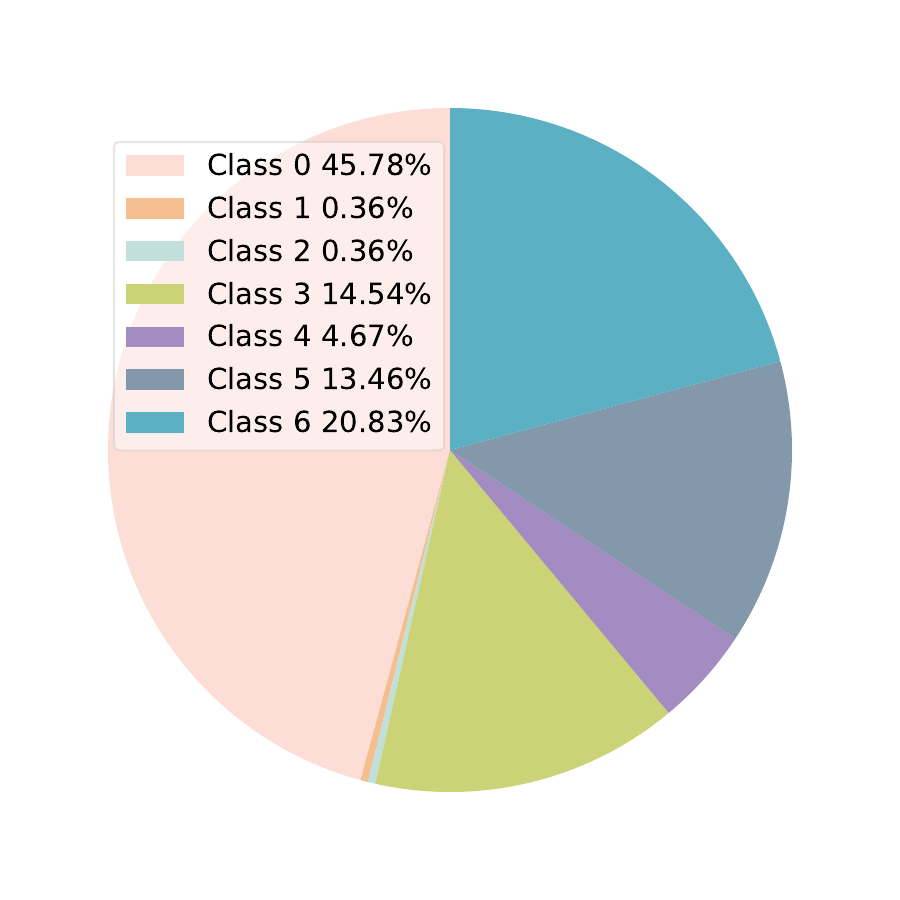}
  }
  \subfigure[Citeseer-Client1]{
    \includegraphics[scale=0.25]{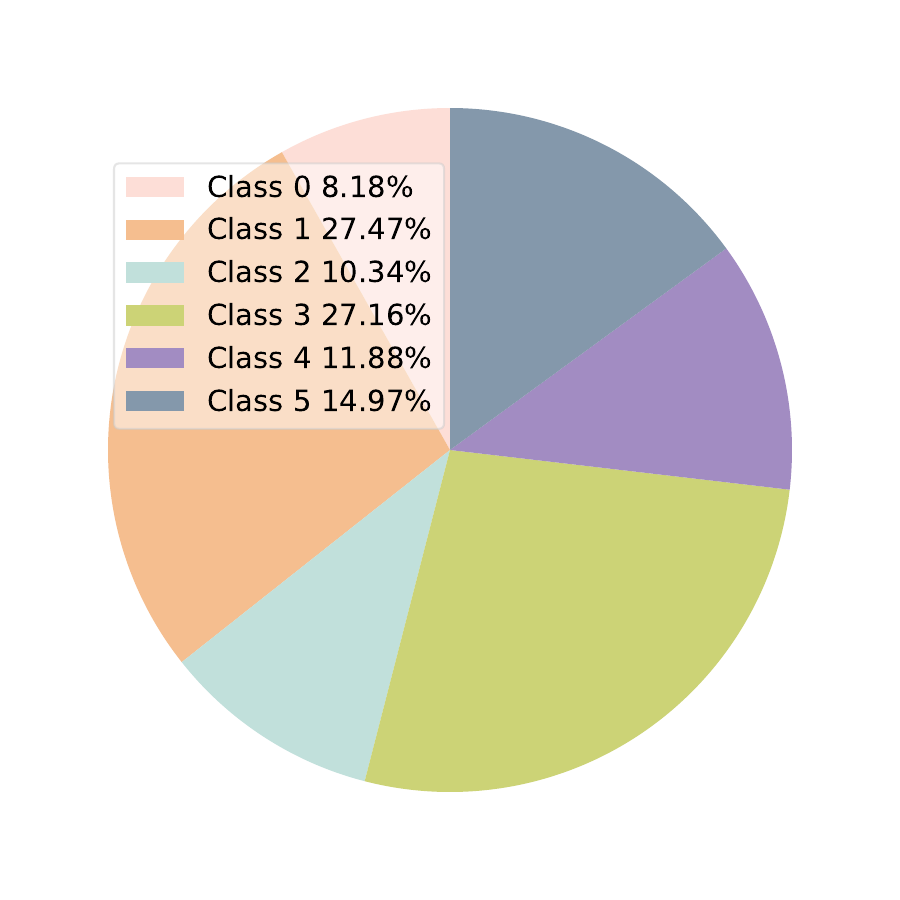}
  }
  \subfigure[Citeseer-Client2]{
    \includegraphics[scale=0.25]{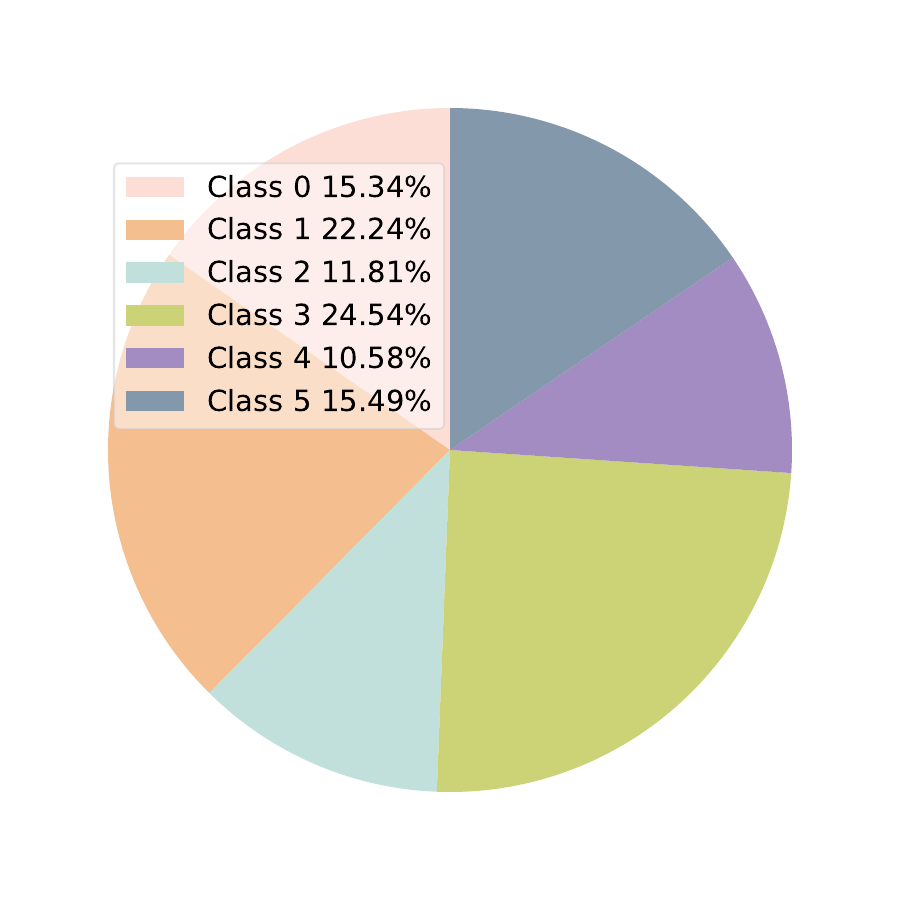}
  }
  \subfigure[Citeseer-Client3]{
    \includegraphics[scale=0.25]{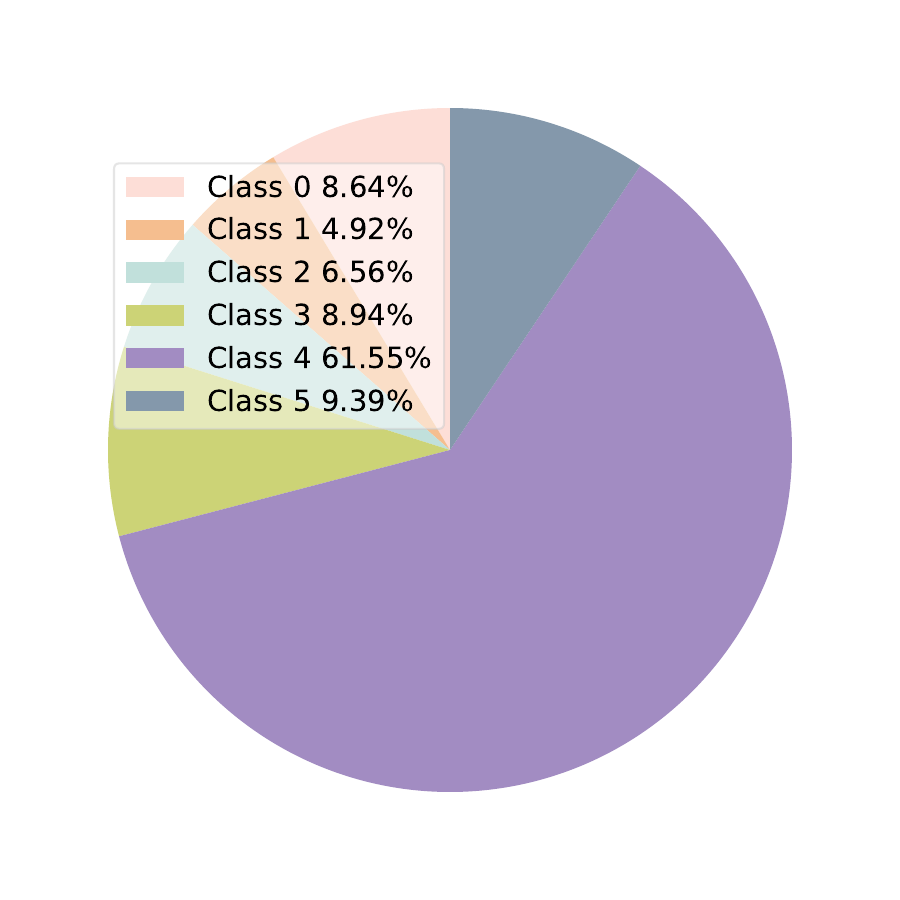}
  }
  \subfigure[Citeseer-Client4]{
    \includegraphics[scale=0.25]{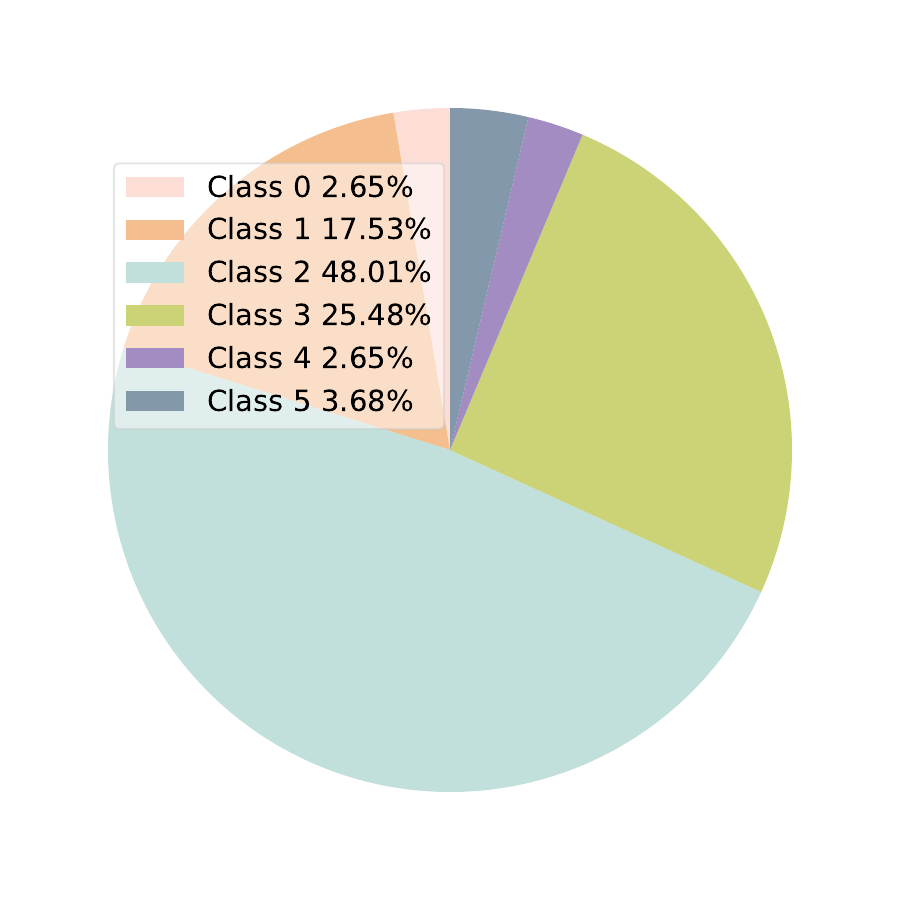}
  }
  \subfigure[Citeseer-Client5]{
    \includegraphics[scale=0.25]{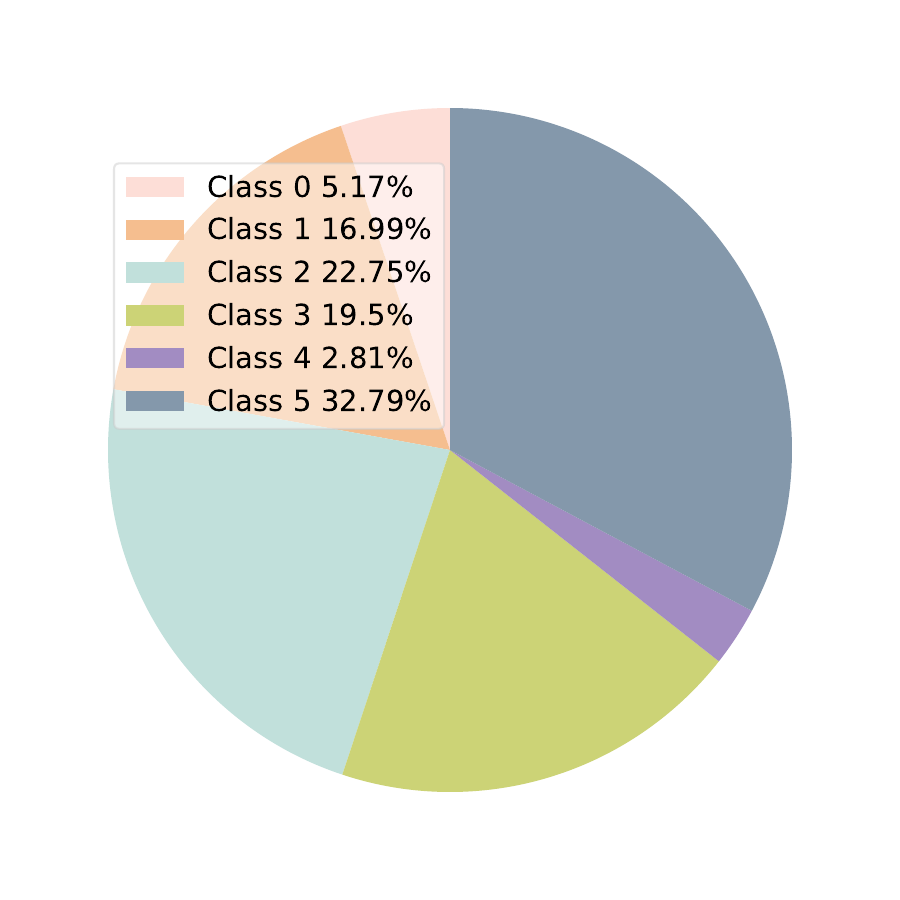}
  }
  \subfigure[PubMed-Client1]{
    \includegraphics[scale=0.25]{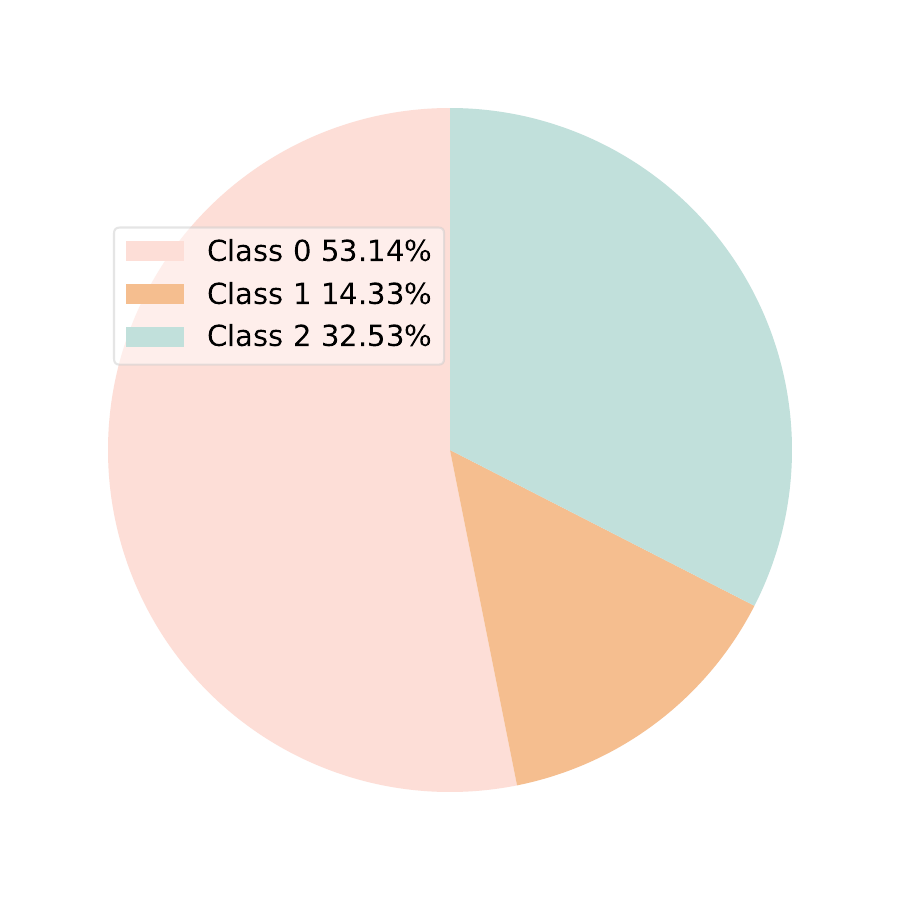}
  }
  \subfigure[PubMed-Client2]{
    \includegraphics[scale=0.25]{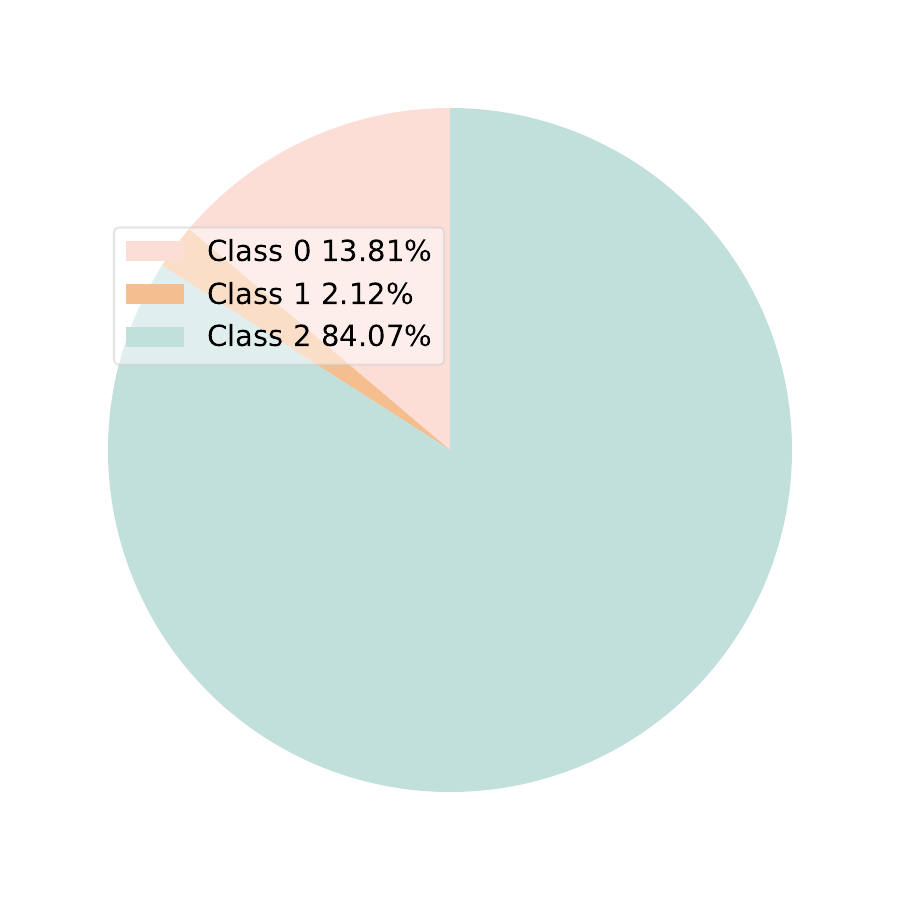}
  }
  \subfigure[PubMed-Client3]{
    \includegraphics[scale=0.25]{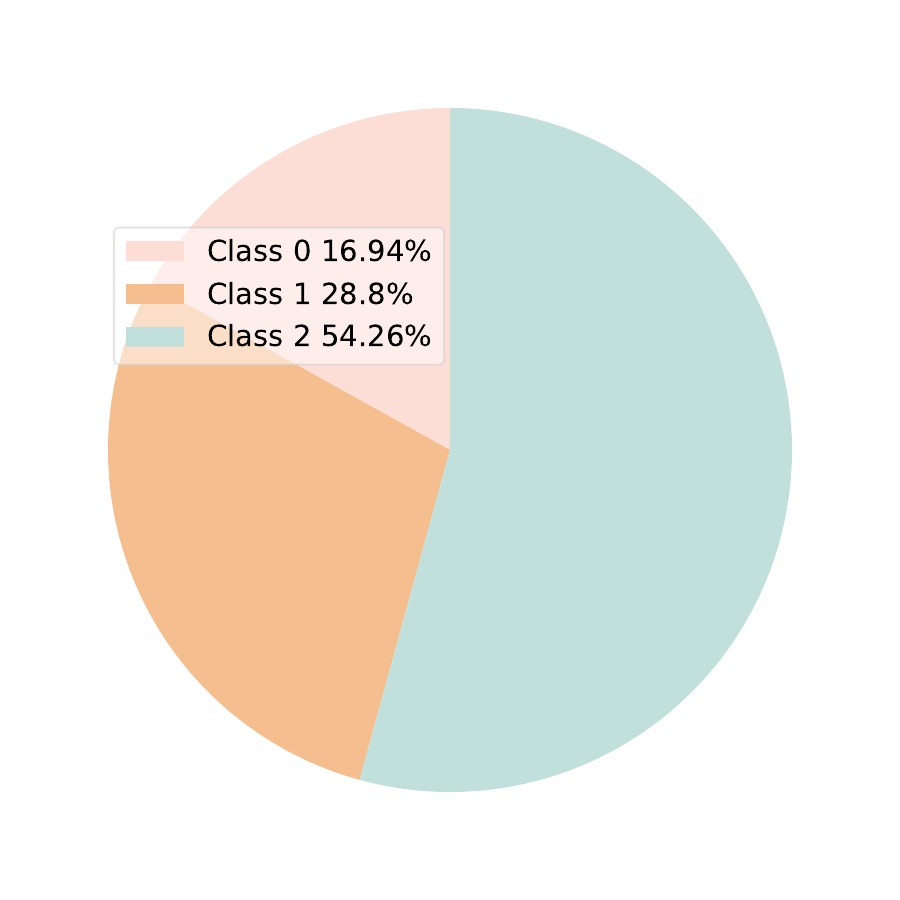}
  }
  \subfigure[PubMed-Client4]{
    \includegraphics[scale=0.25]{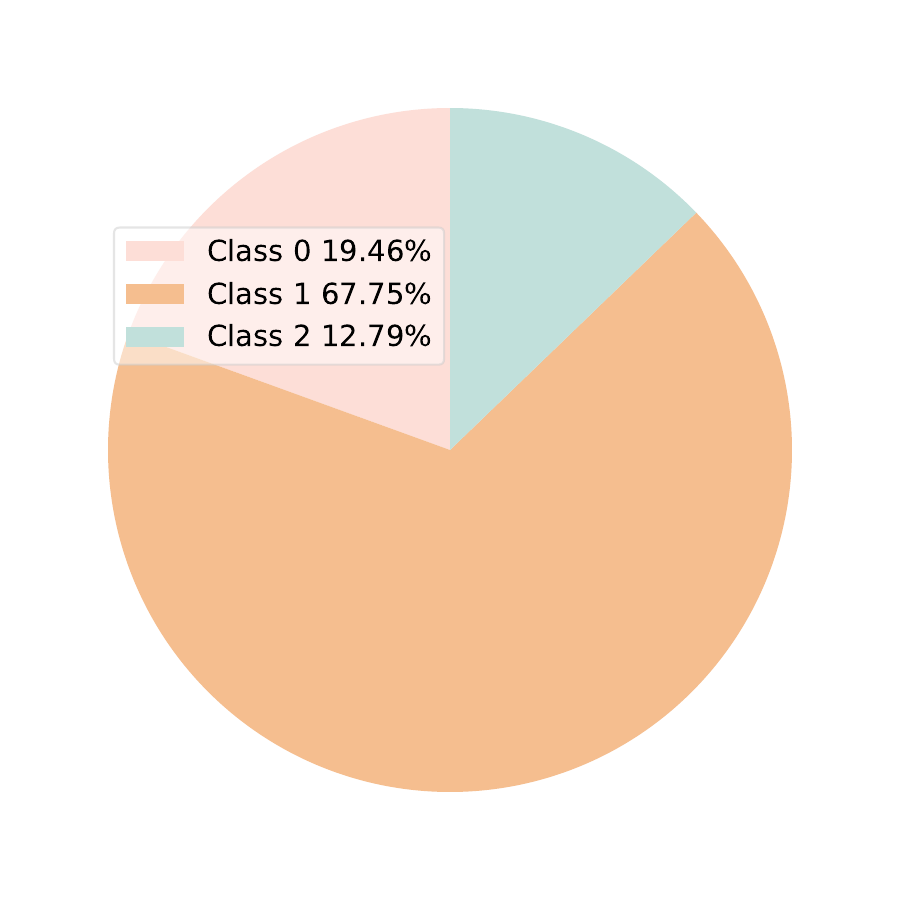}
  }
  \subfigure[PubMed-Client5]{
    \includegraphics[scale=0.25]{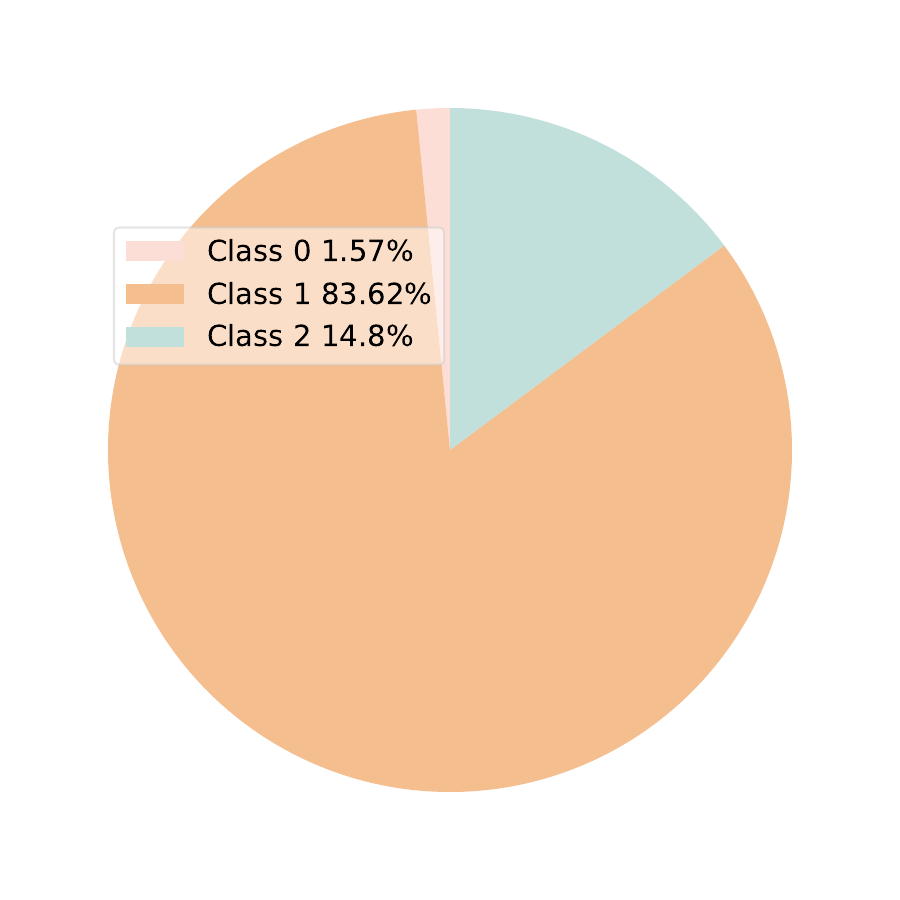}
  }
  \caption{Class distribution in different clients based on citation networks.}
  \label{fig_client_class_distribution}
\end{figure}

We report the statistics of the datasets we used, as shown in Table~\ref{tab_dataset_statistics}. For instance, we partition the target dataset into subgraphs for 5 clients using the METIS\cite{1998metis}. The class distribution of nodes for each client is illustrated in Fig~\ref{fig_client_class_distribution}. All evaluated membership inference attacks (MIAs) are conducted in heterogeneous scenarios where the data distribution is inherently unbalanced between clients.
\begin{table}
\centering
\caption{Dataset Statistics}
\begin{tabular}{ccccccc}
\toprule
Dataset   & Cora   & Citeseer  & PubMed  & CS      & Physics  & DBLP   \\\midrule
|V|       & 2,708  & 3,327     & 19,717  & 18,333  & 34,493   & 17,716  \\
|E|       & 10,556 & 9,104     & 88,648  & 163,788 & 495,924  & 105,734 \\
\# Classes & 7      & 6         & 3       & 15      &       5  & 4       \\\bottomrule
\end{tabular}
\label{tab_dataset_statistics}
\end{table}

\subsection{Varients Implementation}
\label{var_imp}
\paragraph{\ours (\textit{subgraph})}
To quantify the structural similarity between $ G_{\mathrm{inv}} $ and $ G_{\mathrm{shadow}} $, we compute the degree and clustering coefficient distributions of their nodes. Let $ \deg_G(v) $ represent the degree of node $ v $ in graph $ G $. The degree distributions $ p_1(k) $ and $ p_2(k) $ are defined as:

\begin{equation}
    p_{1}(k) = \frac{\bigl|\{v \in V_{\mathrm{inv}} : \deg_{G_{\mathrm{inv}}}(v) = k\}\bigr|}{|V_{\mathrm{inv}}|}, 
\quad
p_{2}(k) = \frac{\bigl|\{w \in V_{\mathrm{shadow}} : \deg_{G_{\mathrm{shadow}}}(w) = k\}\bigr|}{|V_{\mathrm{shadow}}|}.
\end{equation}
Similarly, the clustering coefficient distributions $ q_1(j) $ and $ q_2(j) $ are computed by dividing the clustering coefficient range $[0, 1]$ into bins $ B_j $:
\begin{equation}
    q_{1}(j) = \frac{\bigl|\{v \in V_{\mathrm{inv}} : \mathrm{CC}_{G_{\mathrm{inv}}}(v) \in B_j\}\bigr|}{|V_{\mathrm{inv}}|}, 
\quad
q_{2}(j) = \frac{\bigl|\{w \in V_{\mathrm{shadow}} : \mathrm{CC}_{G_{\mathrm{shadow}}}(w) \in B_j\}\bigr|}{|V_{\mathrm{shadow}}|}.
\end{equation}
To stabilize the calculation of KL divergence, we truncate probabilities below a threshold $ \epsilon = 10^{-3} $:
\begin{equation}
    p_{1,k}' = \max\bigl(p_{1}(k), \epsilon\bigr), \quad 
    p_{2,k}' = \max\bigl(p_{2}(k), \epsilon\bigr), \quad
    q_{1,j}' = \max\bigl(q_{1}(j), \epsilon\bigr), \quad
    q_{2,j}' = \max\bigl(q_{2}(j), \epsilon\bigr).
\end{equation}
The KL divergence for the degree and clustering coefficient distributions is then:

\begin{equation}
    D_{\mathrm{KL}}(p_1 \| p_2) = \sum_{k} p_{1,k}' \ln\frac{p_{1,k}'}{p_{2,k}'}, 
\quad
D_{\mathrm{KL}}(q_1 \| q_2) = \sum_{j} q_{1,j}' \ln\frac{q_{1,j}'}{q_{2,j}'}.
\end{equation}
The structural similarity score $ S $ between the two subgraphs is:

\begin{equation}
    S = D_{\mathrm{KL}}(p_1 \| p_2) + D_{\mathrm{KL}}(q_1 \| q_2) = \sum_{k} p_{1,k}' \ln\frac{p_{1,k}'}{p_{2,k}'} + \sum_{j} q_{1,j}' \ln\frac{q_{1,j}'}{q_{2,j}'}.
\end{equation}

\subsection{Additional performance of \ours}
\label{cc-mia_on_other_gnn}
We report the performance of additional datasets under federated GCN, with results for membership inferences and client-data identification detailed in Table~\ref{tab_training_set_mia_appendix} and Table~\ref{tab_client_ownership_mia_appendix}, respectively. Furthermore, we report the membership inference performance of federated GAT and federated GraphSAGE across all datasets, as shown in Table~\ref{tab_training_set_mia_gat} and Table~\ref{tab_training_set_mia_sage}.

\begin{table}[]
\centering
\scriptsize
\caption{Performance (AUC~\%) of federated GCN member inference attacks on other datasets.}
\begin{tabular}{ccccccc}
\toprule
Dataset                     & Shadow Dataset &FL Approaches& HP-MIA     & GAN-Based & CS-MIA  & \ours \\\midrule
\multirow{5}{*}{CS}         &   Cora         & FedAvg    &  53.46     &  50.47    & 68.24   & \textbf{81.60}\\
                            &   Citeseer     & FedProx   &  52.83     &  50.84    & 59.78   & \textbf{64.70}\\
                            &   DBLP         & SCAFFOLD  &  55.87     &  53.01    & 58.28   & \textbf{72.46}\\
                            &   Citeseer     & FedDF     &  50.13     &  55.34    & 61.47   & \textbf{78.82}\\
                            &   DBLP         & FedNova   &  53.42     &  56.12    & 58.44   & \textbf{76.27}\\\hline
\multirow{5}{*}{Physics}    &   PubMed       & FedAvg    &  54.67     &  57.04    & 59.15   & \textbf{73.00}\\
                            &   Citeseer     & FedProx   &  55.21     &  53.89    & 79.21   & \textbf{82.45}\\
                            &   DBLP         & SCAFFOLD  &  56.09     &  59.78    & 61.54   & \textbf{80.12}\\
                            &   Citeseer     & FedDF     &  57.73     &  58.56    & 73.75   & \textbf{79.88}\\
                            &   PubMed       & FedNova   &  56.92     &  57.45    & 52.63   & \textbf{75.99}\\\hline
\multirow{5}{*}{DBLP}       &   PubMed       & FedAvg    &  53.47     &  54.98    & 61.91   & \textbf{83.42}\\
                            &   PubMed       & FedProx   &  54.74     &  53.21    & 54.74   & \textbf{81.75}\\
                            &   PubMed       & SCAFFOLD  &  55.99     &  52.08    & 58.23   & \textbf{85.73}\\
                            &   PubMed       & FedDF     &  50.61     &  55.32    & 60.40   & \textbf{78.50}\\
                            &   PubMed       & FedNova   &  51.82     &  54.45    & 55.65   & \textbf{85.22}\\\bottomrule
\end{tabular}
\label{tab_training_set_mia_appendix}
\end{table}

\begin{table}
\centering
\scriptsize
\caption{Performance of federated GCN client-data identification on other datasets. Clts: Clients.}
\begin{tabular}{cccccccccc}
\toprule
                    Dataset &FL Approaches& 3-Clts & 4-Clts & 5-Clts & 6-Clts & 7-Clts & 8-Clts & 9-Clts & 10-Clts\\\midrule
\multicolumn{2}{c}{Random}&\textbf{33.33}&\textbf{25.00}&\textbf{20.00}&\textbf{16.66}&\textbf{14.29}&\textbf{12.50}&\textbf{11.11}&\textbf{10.00}\\\midrule
\multirow{5}{*}{CS}      & FedAvg    & 41.78  & 36.78  &  34.81 & 29.15  & 27.49  & 26.32  & 24.79  & 21.63 \\
                         & FedProx   & 38.26  & 34.35  &  30.16 & 30.11  & 27.92  & 23.37  & 19.66  & 19.19 \\
                         & SCAFFOLD  & 52.60  & 40.58  &  37.32 & 32.10  & 29.01  & 26.82  & 18.39  & 17.66 \\
                         & FedDF     & 48.07  & 45.57  &  37.42 & 33.40  & 28.53  & 26.05  & 25.14  & 22.98 \\
                         & FedNova   & 50.45  & 34.79  &  32.14 & 30.12  & 29.11  & 26.32  & 23.37  & 20.51 \\\hline
\multirow{5}{*}{Physics} & FedAvg    & 42.75  & 38.40  &  34.05 & 30.25  & 28.15  & 27.63  & 27.46  & 20.43 \\
                         & FedProx   & 48.25  & 42.14  &  37.89 & 31.04  & 26.03  & 24.09  & 22.54  & 19.15 \\
                         & SCAFFOLD  & 45.15  & 40.60  &  35.10 & 31.45  & 29.02  & 25.56  & 23.03  & 20.57 \\
                         & FedDF     & 40.10  & 36.42  &  32.04 & 28.00  & 24.54  & 20.36  & 18.15  & 15.74 \\
                         & FedNova   & 41.05  & 30.78  &  29.62 & 26.34  & 24.36  & 20.66  & 20.47  & 18.95 \\\hline
\multirow{5}{*}{DBLP}    & FedAvg    & 53.13  & 43.35  &  37.38 &  32.87 & 31.76  & 30.80  & 30.10  & 22.92 \\
                         & FedProx   & 72.25  & 44.42  &  42.26 & 34.94  & 34.19  & 31.99  & 28.56  & 28.77 \\
                         & SCAFFOLD  & 43.13  & 39.13  &  37.69 & 32.24  & 26.21  & 25.82  & 20.61  & 18.43 \\
                         & FedDF     & 60.58  & 46.87  &  38.48 & 37.79  & 30.17  & 27.02  & 21.53  & 20.94 \\
                         & FedNova   & 36.20  & 31.27  &  25.06 & 22.00  & 20.15  & 18.43  & 16.11  & 14.15 \\\bottomrule
\end{tabular}
\label{tab_client_ownership_mia_appendix}
\end{table}

\begin{table}
\scriptsize
\centering
\caption{Performance (AUC~\%) of federated GAT~\cite{gat2018} member inference.}
\begin{tabular}{ccccccc}
\toprule
Dataset                     & Shadow Dataset &FL Approaches& HP-MIA     & GAN-Based & CS-MIA  & \ours \\\midrule
\multirow{5}{*}{Cora}       &    Citeseer    & FedAvg    &  56.32     &  54.87    & 58.12   & \textbf{65.36}\\
                            &    CS          & FedProx   &  50.14     &  49.98    & 59.67   & \textbf{60.41}\\
                            &    CS          & SCAFFOLD  &  52.73     &  51.44    & 57.21   & \textbf{64.32}\\
                            &    PubMed      & FedDF     &  53.98     &  52.10    & 58.79   & \textbf{62.96}\\
                            &    DBLP        & FedNova   &  54.57     &  56.02    & 54.31   & \textbf{61.77}\\\hline
\multirow{5}{*}{Citeseer}   &    PubMed      & FedAvg    &  51.25     &  50.34    & 53.88   & \textbf{64.02}\\
                            &    DBLP        & FedProx   &  52.90     &  53.12    & 54.05   & \textbf{61.75}\\
                            &    DBLP        & SCAFFOLD  &  49.68     &  50.77    & 51.29   & \textbf{58.72}\\
                            &    DBLP        & FedDF     &  54.01     &  51.93    & 54.42   & \textbf{60.13}\\
                            &    DBLP        & FedNova   &  53.77     &  58.00    & 57.84   & \textbf{62.99}\\\hline
\multirow{5}{*}{PubMed}     &    DBLP        & FedAvg    &  55.12     &  50.68    & 56.45   & \textbf{63.00}\\
                            &    DBLP        & FedProx   &  54.40     &  54.37    & 55.18   & \textbf{63.93}\\
                            &    DBLP        & SCAFFOLD  &  53.21     &  54.52    & 49.87   & \textbf{62.33}\\
                            &    DBLP        & FedDF     &  56.00     &  54.66    & 57.39   & \textbf{62.30}\\
                            &    DBLP        & FedNova   &  55.93     &  52.89    & 58.71   & \textbf{63.32}\\\hline
\multirow{5}{*}{CS}         &   Cora         & FedAvg    &  53.01     &  49.12    & 55.90   & \textbf{62.08}\\
                            &   Citeseer     & FedProx   &  52.48     &  50.57    & 53.33   & \textbf{62.87}\\
                            &   Citeseer     & SCAFFOLD  &  55.04     &  52.88    & 54.17   & \textbf{61.97}\\
                            &   Cora         & FedDF     &  50.88     &  55.01    & 52.74   & \textbf{61.96}\\
                            &   Citeseer     & FedNova   &  53.19     &  56.00    & 51.62   & \textbf{60.50}\\\hline
\multirow{5}{*}{Physics}    &   DBLP         & FedAvg    &  54.29     &  57.33    & 54.88   & \textbf{63.38}\\
                            &   DBLP         & FedProx   &  55.95     &  53.60    & 56.77   & \textbf{62.23}\\
                            &   DBLP         & SCAFFOLD  &  56.47     &  59.01    & 52.18   & \textbf{61.52}\\
                            &   CS           & FedDF     &  57.11     &  58.02    & 54.01   & \textbf{61.73}\\
                            &   DBLP         & FedNova   &  56.58     &  57.20    & 50.97   & \textbf{62.36}\\\hline
\multirow{5}{*}{DBLP}       &   Cora         & FedAvg    &  53.84     &  54.33    & 52.10   & \textbf{60.21}\\
                            &   Citeseer     & FedProx   &  54.09     &  53.68    & 54.76   & \textbf{63.85}\\
                            &   Citeseer     & SCAFFOLD  &  55.23     &  52.46    & 56.49   & \textbf{63.57}\\
                            &   Citeseer     & FedDF     &  50.47     &  55.18    & 53.82   & \textbf{63.27}\\
                            &   Citeseer     & FedNova   &  51.68     &  54.72    & 55.09   & \textbf{64.28}\\\bottomrule
\end{tabular}
\label{tab_training_set_mia_gat}
\end{table}

\begin{table}
\scriptsize
\centering
\caption{Performance (AUC~\%) of federated GraphSAGE~\cite{2017graphsage} member inference.}
\begin{tabular}{ccccccc}
\toprule
Dataset                     & Shadow Dataset &FL Approaches& HP-MIA     & GAN-Based & CS-MIA  & \ours \\\midrule
\multirow{5}{*}{Cora}       &    DBLP        & FedAvg    &  55.12     &  53.78    & 57.45   & \textbf{65.82}\\
                            &    PubMed      & FedProx   &  50.97     &  49.35    & 58.76   & \textbf{71.47}\\
                            &    DBLP        & SCAFFOLD  &  52.43     &  51.21    & 56.67   & \textbf{61.35}\\
                            &    DBLP        & FedDF     &  53.84     &  52.98    & 58.47   & \textbf{63.38}\\
                            &    PubMed      & FedNova   &  53.76     &  55.32    & 53.94   & \textbf{72.17}\\\hline
\multirow{5}{*}{Citeseer}   &    PubMed      & FedAvg    &  50.81     &  50.16    & 54.02   & \textbf{71.26}\\
                            &    PubMed      & FedProx   &  52.95     &  52.41    & 53.18   & \textbf{67.83}\\
                            &    PubMed      & SCAFFOLD  &  48.73     &  49.89    & 49.67   & \textbf{63.35}\\
                            &    DBLP        & FedDF     &  54.61     &  51.65    & 53.14   & \textbf{63.60}\\
                            &    PubMed      & FedNova   &  54.02     &  56.11    & 56.72   & \textbf{64.91}\\\hline
\multirow{5}{*}{PubMed}     &    Cora        & FedAvg    &  53.84     &  50.37    & 56.12   & \textbf{62.37}\\
                            &    Cora        & FedProx   &  54.04     &  53.18    & 55.24   & \textbf{60.57}\\
                            &    Cora        & SCAFFOLD  &  52.98     &  53.76    & 50.14   & \textbf{61.98}\\
                            &    Cora        & FedDF     &  55.71     &  54.11    & 56.72   & \textbf{61.64}\\
                            &    Cora        & FedNova   &  54.89     &  53.24    & 57.94   & \textbf{60.99}\\\hline
\multirow{5}{*}{CS}         &   DBLP         & FedAvg    &  51.23     &  49.78    & 54.12   & \textbf{78.74}\\
                            &   DBLP         & FedProx   &  51.87     &  49.45    & 52.87   & \textbf{79.25}\\
                            &   DBLP         & SCAFFOLD  &  55.78     &  53.12    & 54.04   & \textbf{79.77}\\
                            &   DBLP         & FedDF     &  50.62     &  54.29    & 53.18   & \textbf{77.58}\\
                            &   DBLP         & FedNova   &  54.21     &  55.81    & 50.12   & \textbf{77.80}\\\hline
\multirow{5}{*}{Physics}    &   PubMed       & FedAvg    &  54.88     &  57.11    & 53.92   & \textbf{63.84}\\
                            &   PubMed       & FedProx   &  55.14     &  54.29    & 56.32   & \textbf{65.61}\\
                            &   PubMed       & SCAFFOLD  &  55.98     &  58.72    & 52.34   & \textbf{65.73}\\
                            &   PubMed       & FedDF     &  56.41     &  57.29    & 53.47   & \textbf{65.61}\\
                            &   PubMed       & FedNova   &  55.97     &  56.43    & 50.87   & \textbf{58.40}\\\hline
\multirow{5}{*}{DBLP}       &   PubMed       & FedAvg    &  53.12     &  53.58    & 52.78   & \textbf{69.66}\\
                            &   PubMed       & FedProx   &  54.58     &  52.98    & 54.92   & \textbf{71.39}\\
                            &   PubMed       & SCAFFOLD  &  55.31     &  53.12    & 57.84   & \textbf{73.50}\\
                            &   PubMed       & FedDF     &  50.34     &  54.62    & 54.11   & \textbf{78.56}\\
                            &   Citeseer     & FedNova   &  52.21     &  53.72    & 55.34   & \textbf{71.71}\\
\bottomrule
\end{tabular}
\label{tab_training_set_mia_sage}
\end{table}

\subsection{Convergence Analysis}
\begin{figure}
  \centering
  \subfigure[Cora]{
    \includegraphics[scale=0.27]{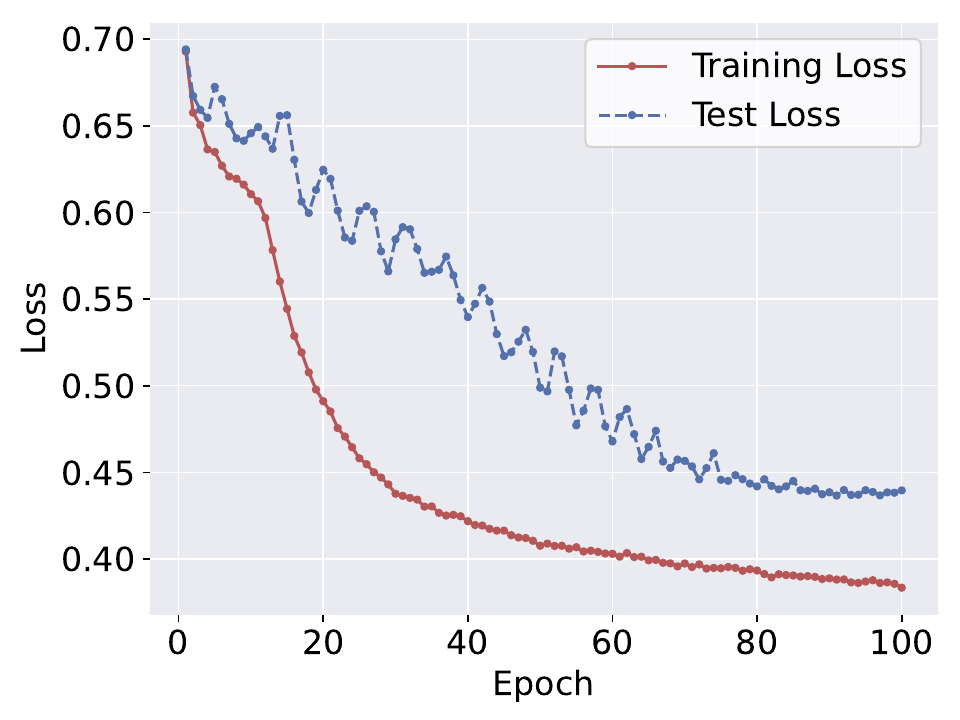}
  }
  \subfigure[Citeseer]{
    \includegraphics[scale=0.27]{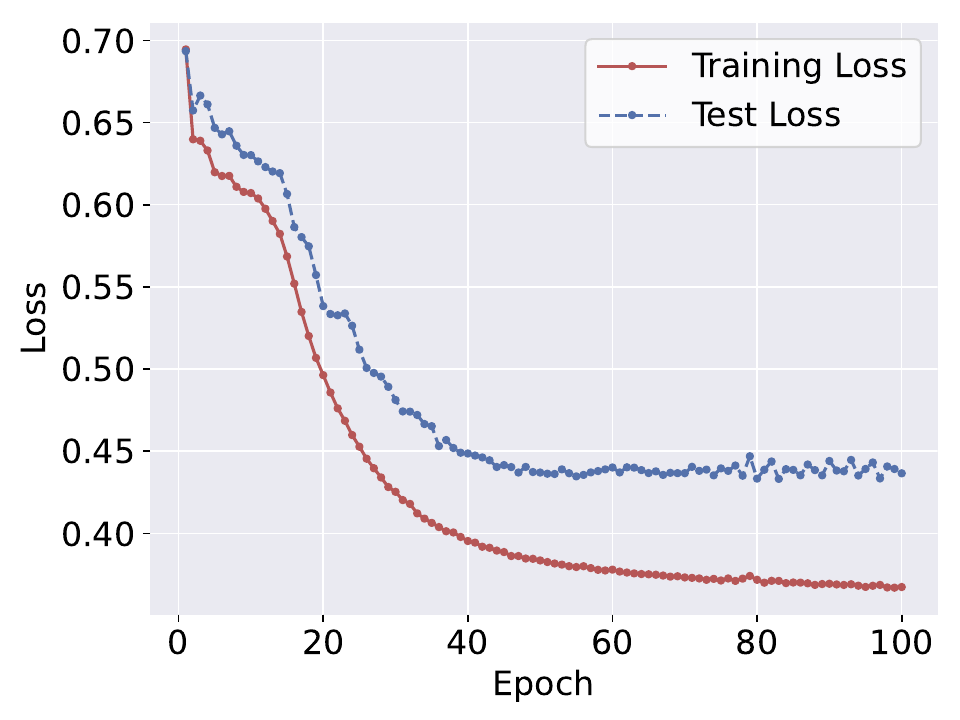}
  }
  \subfigure[PubMed]{
    \includegraphics[scale=0.27]{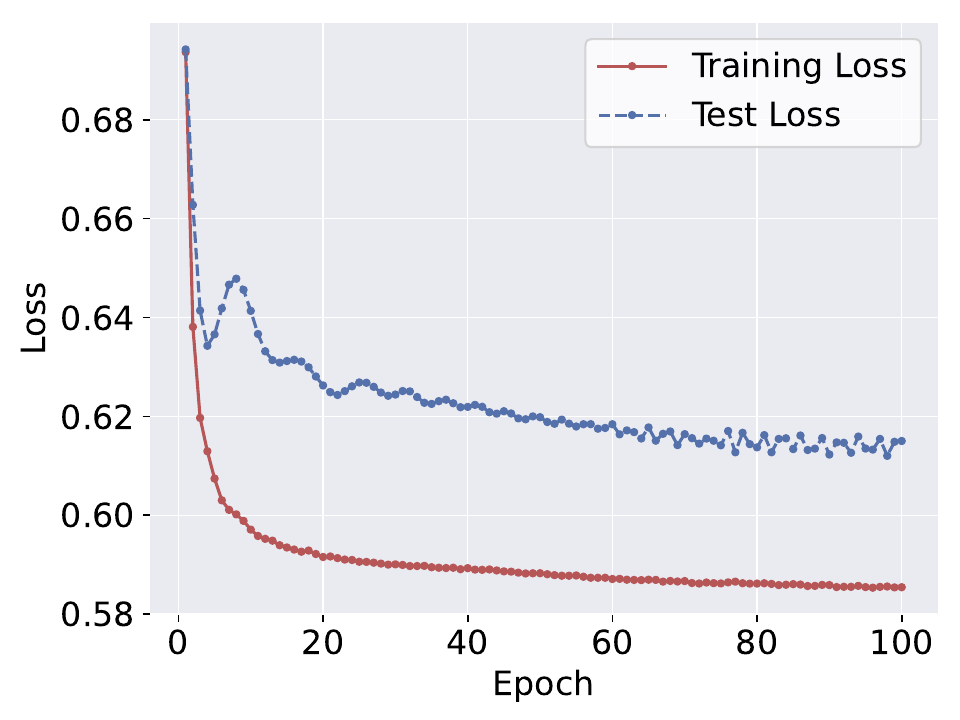}
  }
  \subfigure[CS]{
    \includegraphics[scale=0.27]{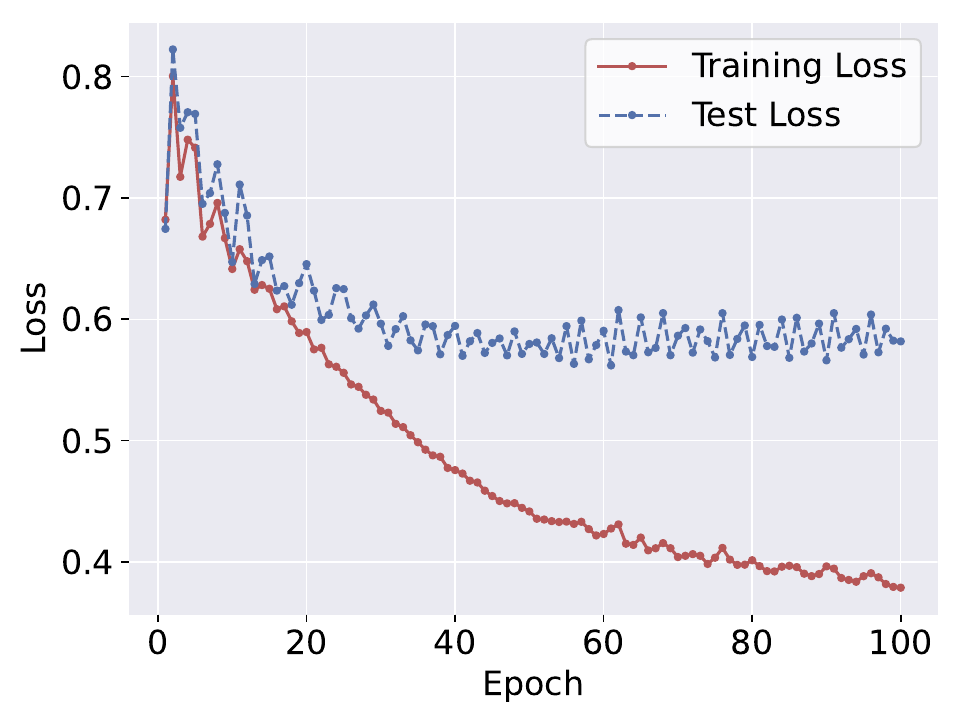}
  }
  \subfigure[Physics]{
    \includegraphics[scale=0.27]{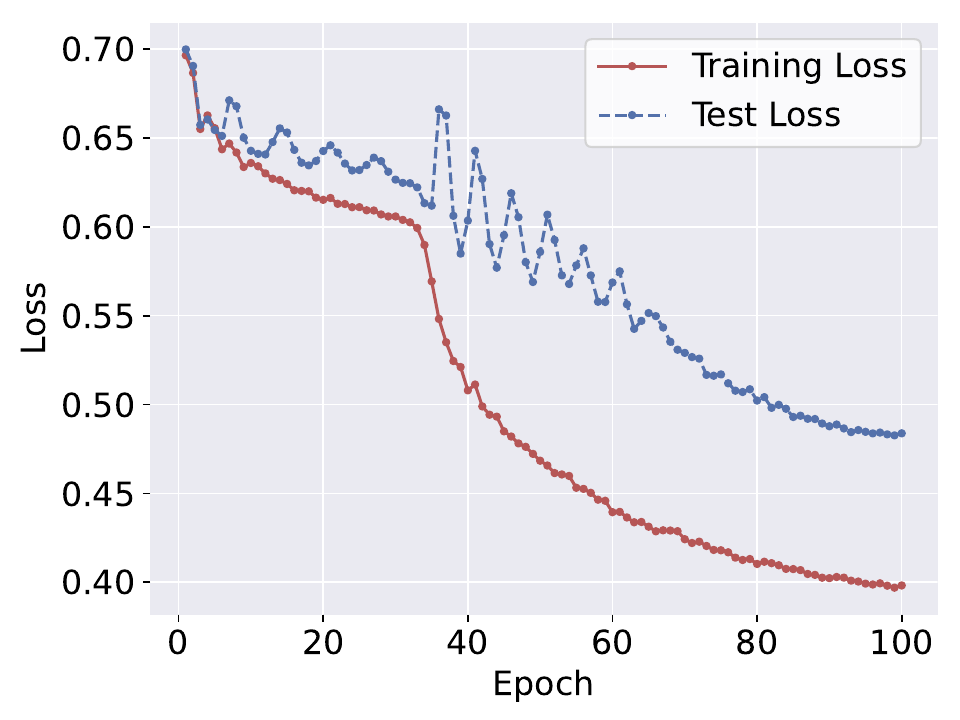}
  }
  \subfigure[DBLP]{
    \includegraphics[scale=0.27]{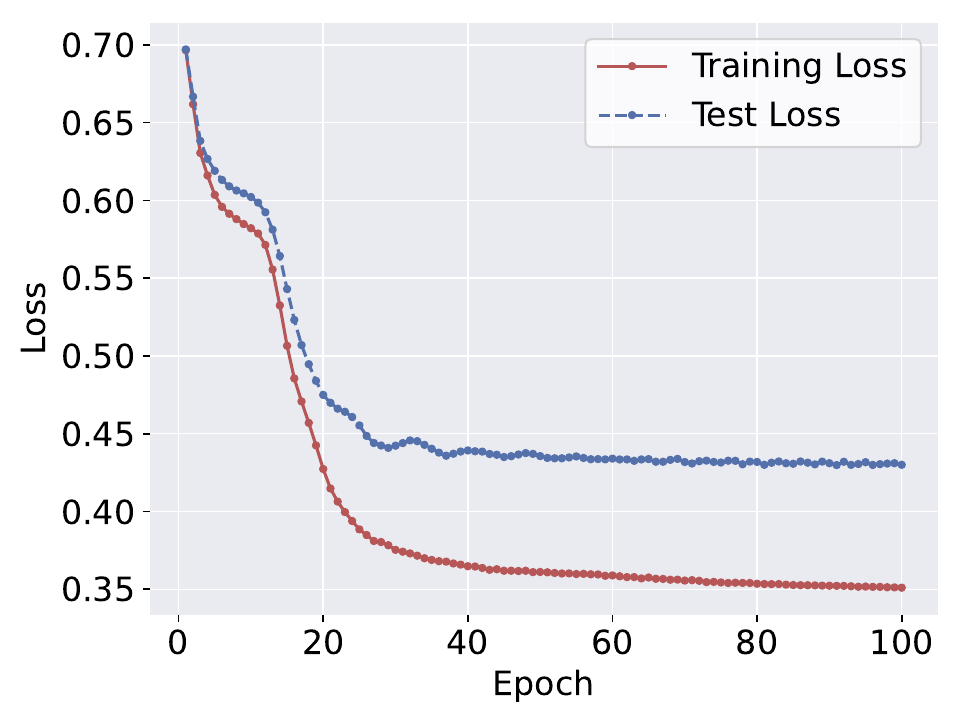}
  }  
  \caption{The convergence of \ours under membership inference.}
  \label{fig_cc-mia_loss}
\end{figure}

To validate the effectiveness of \ours, we conducted convergence evaluations for both membership inference and client-data identification. Specifically, for membership inference, we plot the loss curves of the attack model under FedAVG, using the best-matched shadow dataset from Table~\ref{tab_cc-mia}, as illustrated in Fig~\ref{fig_cc-mia_loss}. \ours achieves convergence for most datasets within 100 epochs.

\begin{figure}[h]
  \centering
  \subfigure[Cora]{
    \includegraphics[scale=0.27]{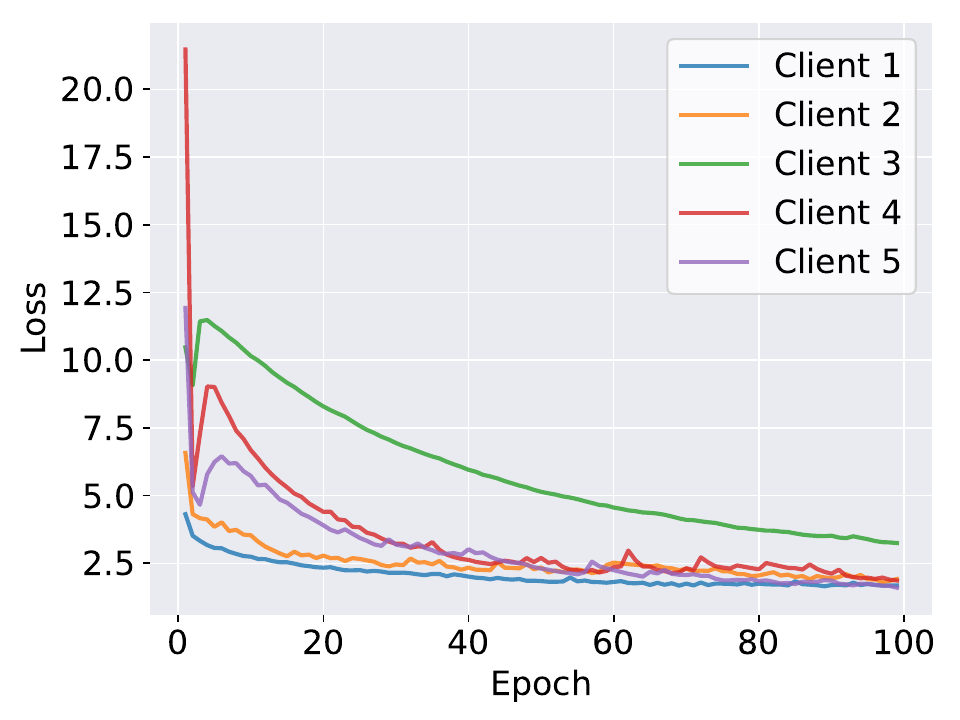}
  }
  \subfigure[Citeseer]{
    \includegraphics[scale=0.27]{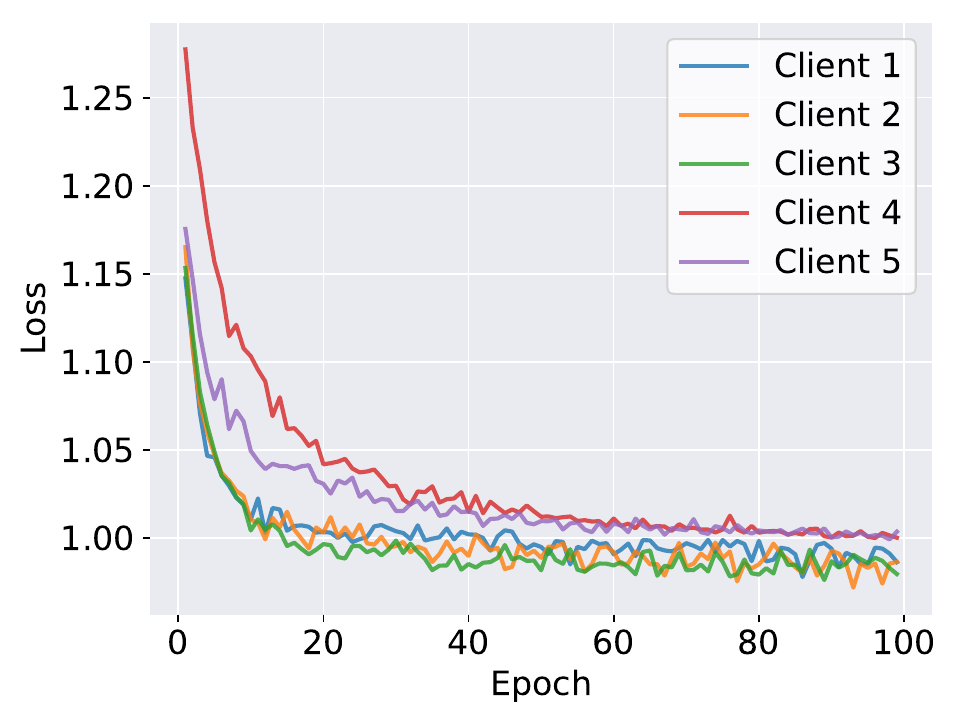}
  }
  \subfigure[PubMed]{
    \includegraphics[scale=0.27]{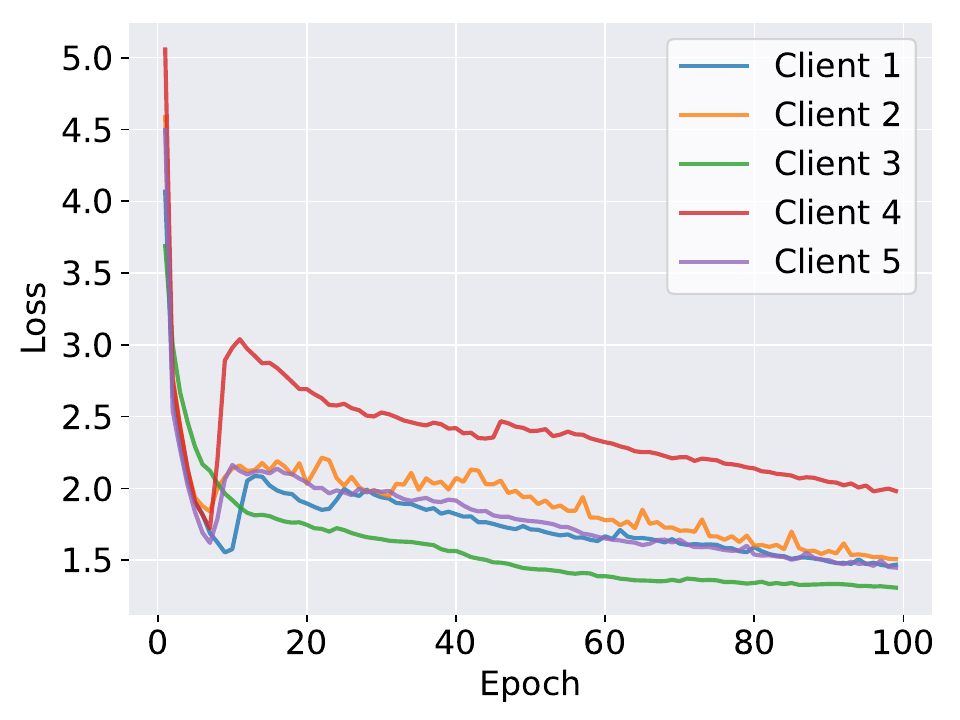}
  }
  \subfigure[CS]{
    \includegraphics[scale=0.27]{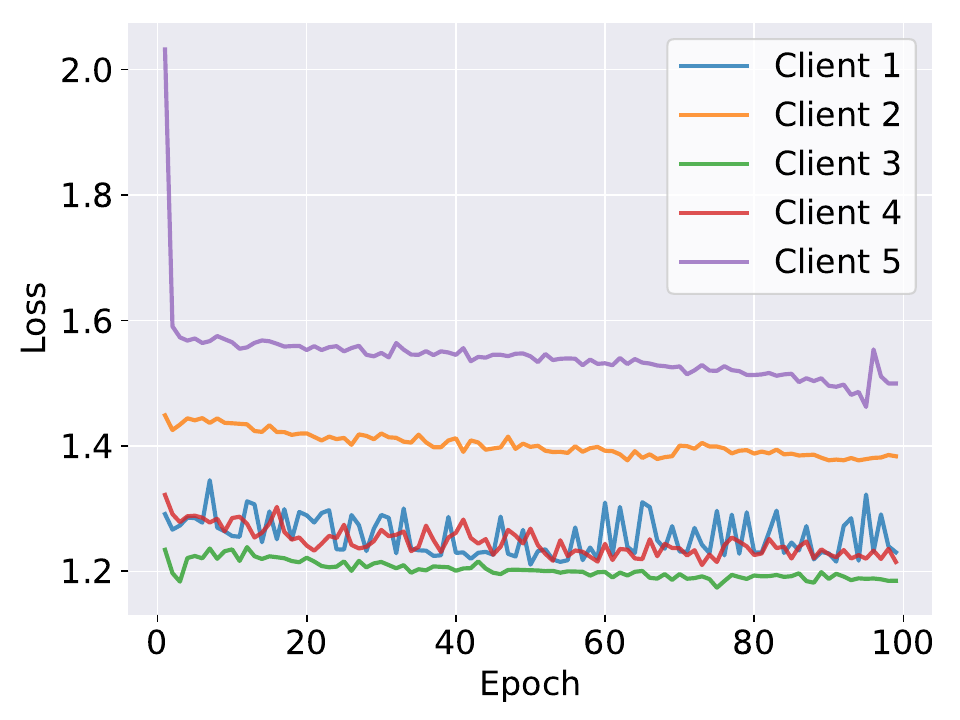}
  }
  \subfigure[Physics]{
    \includegraphics[scale=0.27]{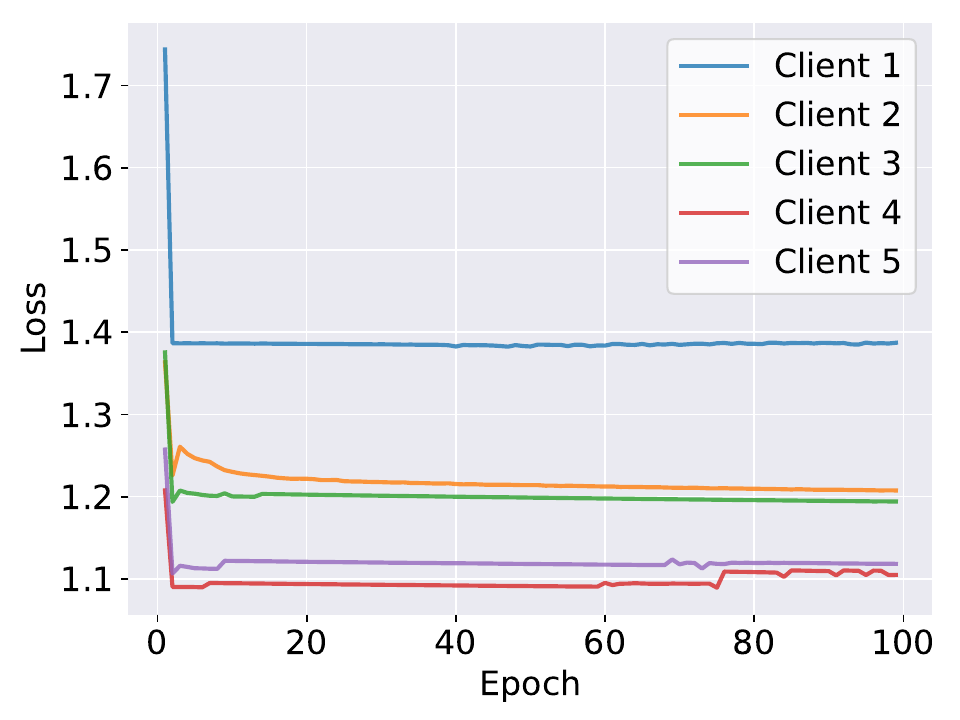}
  }
  \subfigure[DBLP]{
    \includegraphics[scale=0.27]{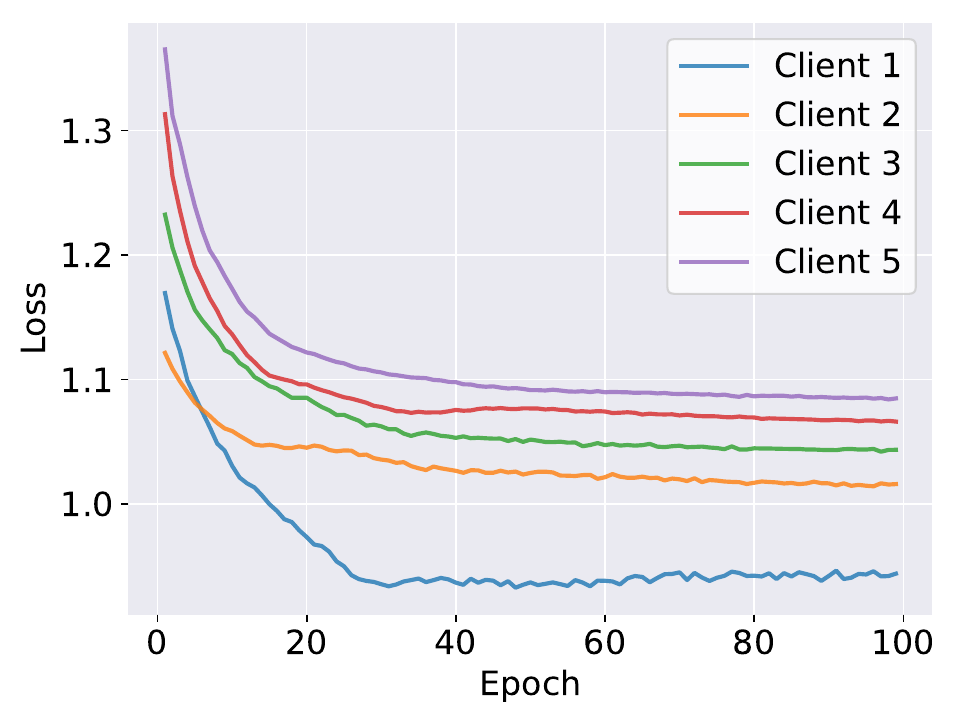}
  }
  \caption{The convergence of \ours under client-data identification.}
  \label{fig_cc-mia_identification_loss}
\end{figure}

For client-data identification, we examined the convergence of gradient inversion by plotting the loss curves for reconstructing node features and graph structures, as shown in Fig~\ref {fig_cc-mia_identification_loss}. Using a client number of 5 and FedAVG aggregation as an example, we tested gradient inversion with GCN across various datasets. The convergence trend is stable for Cora and PubMed after early oscillation. For CS and Physics, loss sharply declines during the initial epochs before stabilizing in subsequent iterations. In contrast, the DBLP dataset exhibits a gradual convergence, achieving stability after approximately 40 epochs. Overall, the gradient inversion process for client-membership MIA consistently approximates the original graph, ensuring subsequent prototype construction.

\subsection{Cost}
\label{cost}
We evaluated the time and space overhead of \ours across different datasets. Specifically, all experiments were conducted on an Nvidia RTX 3090 Ti GPU, with the number of iterations for both classifier training and gradient inversion set to 100, and the number of clients fixed at 3. The computational overhead was recorded for three components: (1) the attacker's classifier training, which corresponds to the membership inference classification phase in \ours, and (2) the gradient inversion and (3) prototype matching, which constitute the two-step process for client-data identification. The results are summarized in Table~\ref{tab_cost}.

Since the parameter size of the classifier is fixed, its space complexity remains constant regardless of the dataset size. However, the time complexity increases with the number of nodes and edge density. As elaborated in Appendix~\ref{complexity}, the runtime of the client-data identification is dominated by the gradient inversion step. Nevertheless, due to the extensive distance computations and storage requirements in the prototype matching step, its space overhead is the largest among the three components.

\begin{table}
\caption{Time and space cost of \ours on different datasets.}
\begin{tabular}{cccccccc}
\toprule
\multicolumn{2}{c}{Dataset}                    & Cora   & Citeseer & PubMed  & CS      & Physics & DBLP \\
\midrule
\multirow{3}{*}{Time (S)}   & Train Classifier  & 0.07   & 0.08     & 0.21    & 0.20    & 0.37    & 0.19   \\
                            & Grad Inversion   & 0.88   & 1.08     & 52.25   & 51.05   & 208.47  & 45.31   \\
                            & Prototype Match & 1.15   & 1.41     & 8.60    & 8.09    & 14.65   &  7.43    \\
\midrule
\multirow{3}{*}{Space (MB)} & Train Classifier & 3.02   & 3.02     & 3.02    & 3.02    & 3.02    & 3.02  \\
                            & Grad Inversion   & 6.26   & 9.24     & 317.89  & 268.58  & 951.32  & 275.65   \\
                            & Prototype Match & 48.09  & 96.45    & 1560.12 & 1793.83 & 5713.85 & 1343.37     \\
\bottomrule
\end{tabular}
\label{tab_cost}
\end{table}

\subsection{Potential Defense}
\label{defense}
Perturbation mechanisms are a common defense against inference attacks. In centralized settings, perturbations are typically applied to GNN model outputs to disrupt adversaries' ability to infer posterior probabilities, hindering MIAs. However, in \ours, where attackers leverage gradient information from shadow datasets and target clients via the global model, output perturbations are unusable since clients cannot modify the global model directly.

An alternative defense is to apply perturbations to client node features, disrupting the attacker's ability to exploit gradients derived from these features. While effective, this defense introduces a trade-off, as perturbations degrade the performance of the federated GNN aggregated.  Specifically, GCN~\cite{2017gcn} and FedAvg~\cite{2017fedavg} are applied to verify the efficiency of potential defenses.

Differential Privacy (DP) is a well-established perturbation-based mechanism that provides a robust framework for protecting data confidentiality. By introducing carefully calibrated noise during the training or inference, DP significantly reduces the risk of extracting sensitive information from the perturbed data~\cite{2022llmdp}. Building on this, we define $d_\chi$-Privacy as follows:

\begin{theorem}[$d_\chi$-Privacy]
Let $X$ denote the input domain, $Y$ the output domain, and $d_\chi$ a distance metric over $X$. A randomized mechanism $M: X \to Y$ satisfies $(\eta d_\chi)$-privacy if, for any two inputs $x, x' \in X$ and any subset $S \subseteq Y$, the following inequality holds:
\begin{equation}
\frac{\Pr[M(x) \in S]}{\Pr[M(x') \in S]} \leq e^{\eta d_\chi(x, x')},
\end{equation}
where $\eta \geq 0$ represents the privacy budget, balancing the trade-off between privacy and utility.
\end{theorem}

In the context of graph data, we define $X$ as the original node features and $X'$ as their corresponding perturbed features, which are introduced to mitigate the effectiveness of \ours.

\begin{figure}[h]
  \centering
  \subfigure[Cora]{
    \includegraphics[scale=0.27]{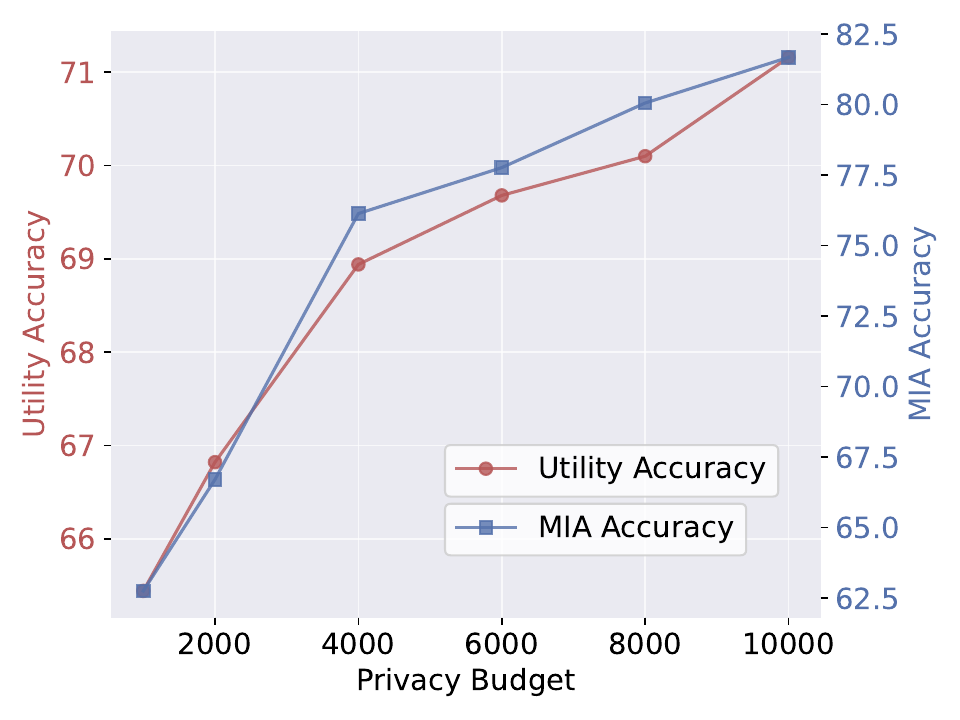}
  }
  \subfigure[Citeseer]{
    \includegraphics[scale=0.27]{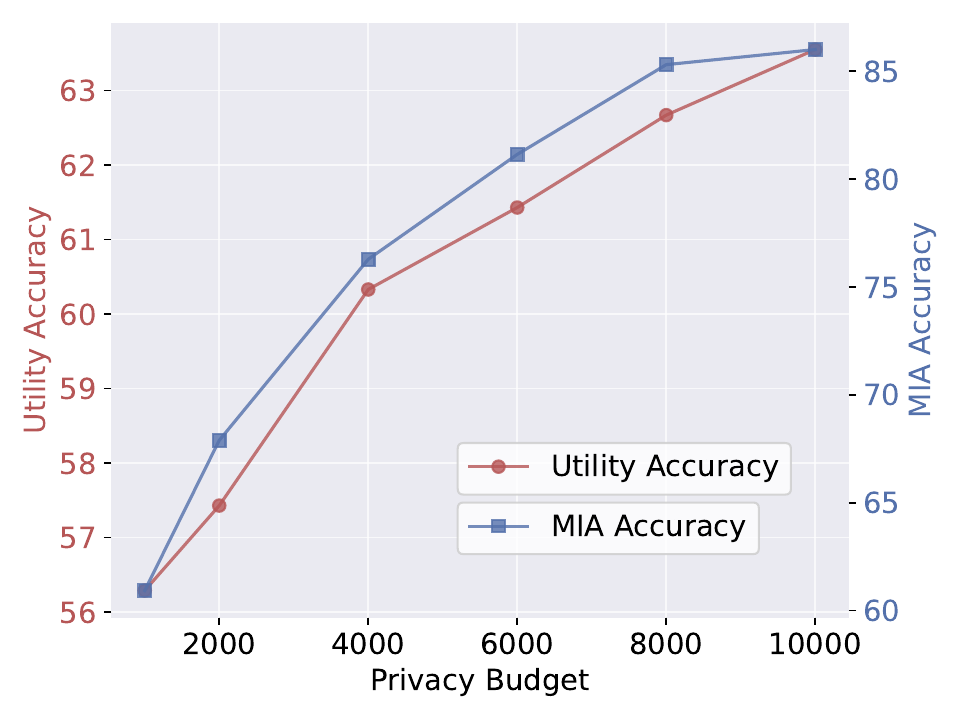}
  }
  \subfigure[PubMed]{
    \includegraphics[scale=0.27]{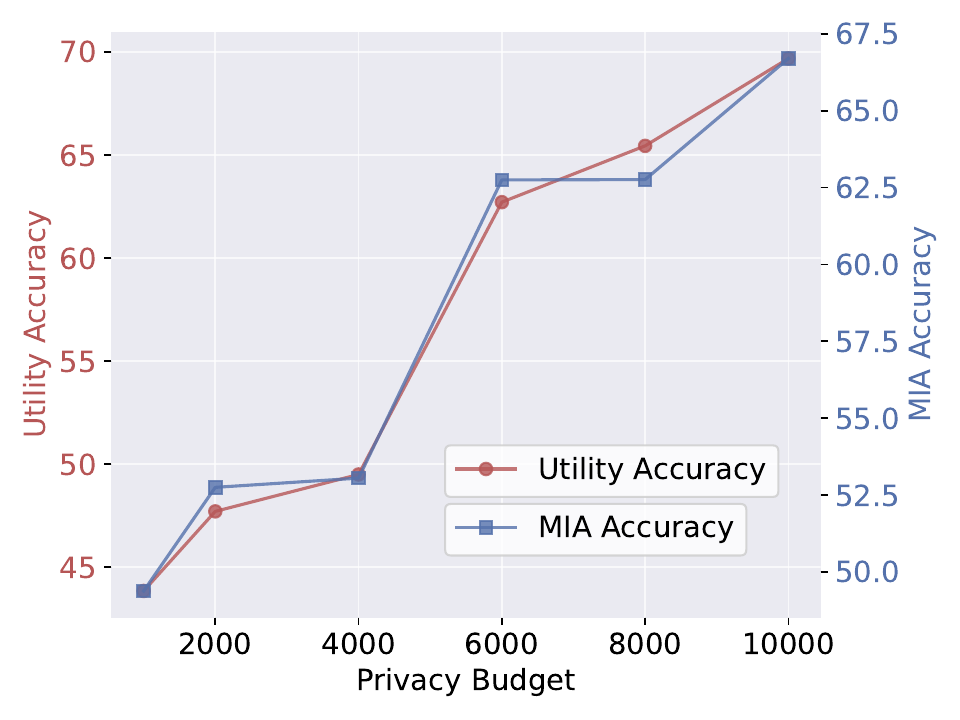}
  }
  \subfigure[CS]{
    \includegraphics[scale=0.27]{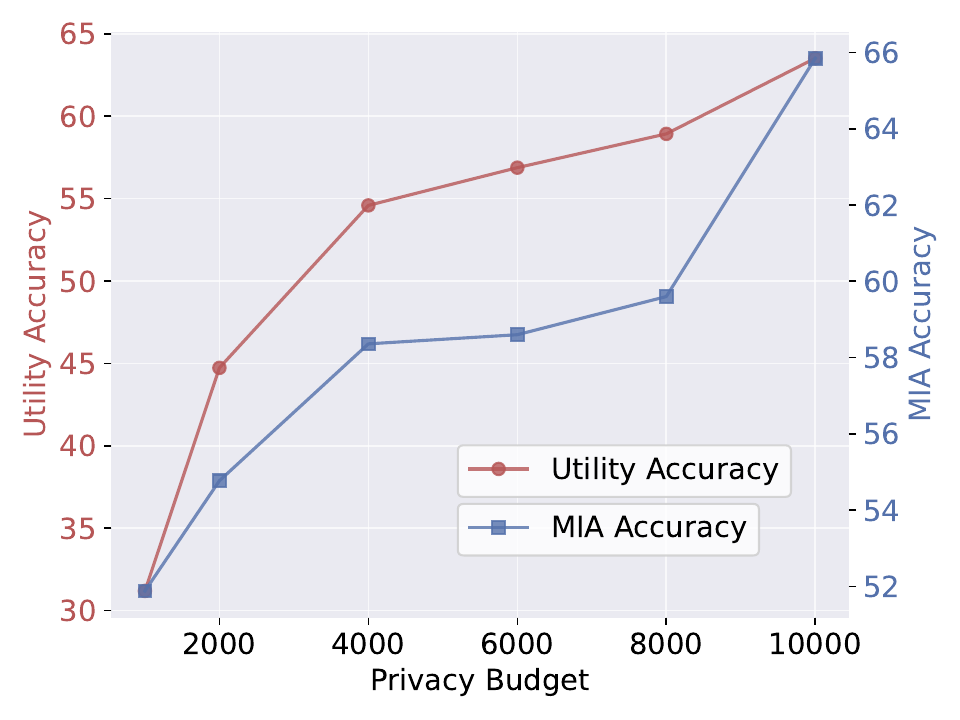}
  }
  \subfigure[Physics]{
    \includegraphics[scale=0.27]{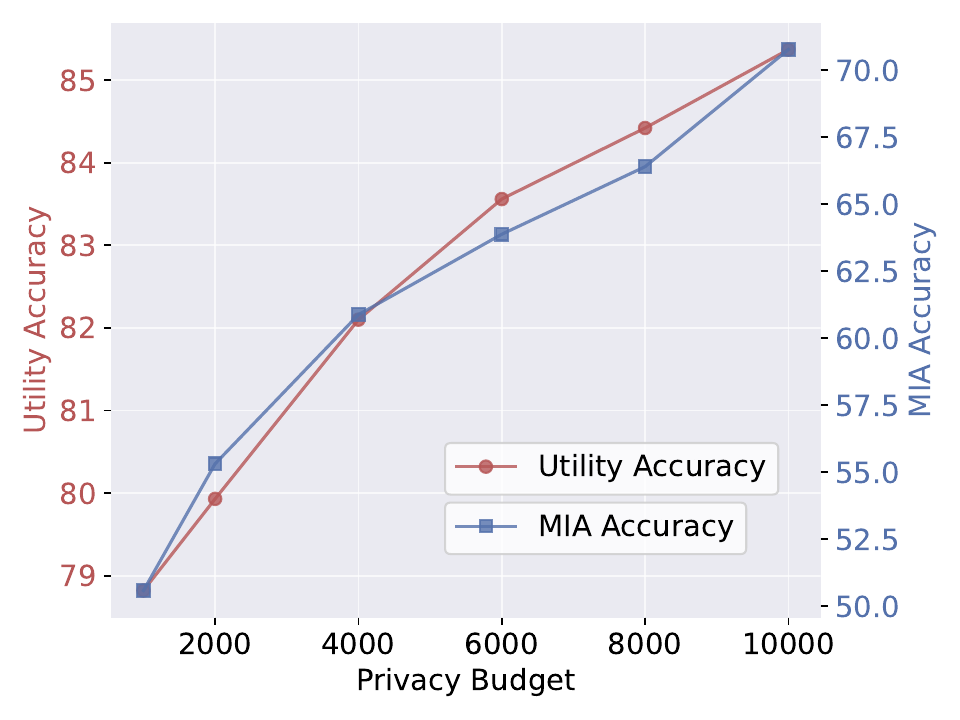}
  }
  \subfigure[DBLP]{
    \includegraphics[scale=0.27]{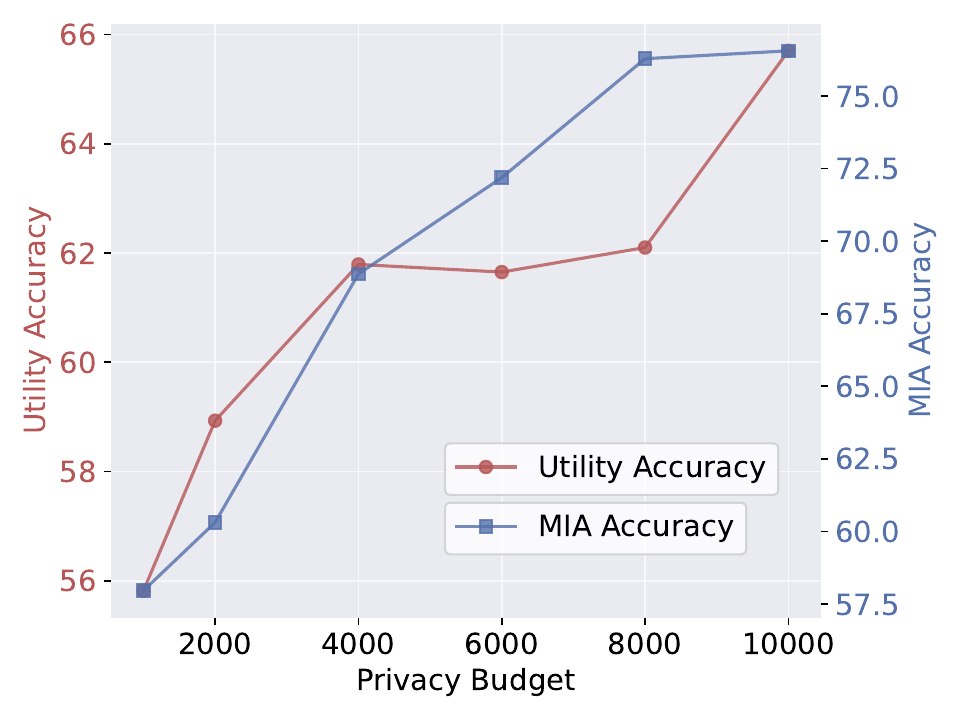}
  }
  \caption{The potential defense against \ours for training-set inference.}
  \label{defense_training_set}
\end{figure}

\begin{figure}[h]
  \centering
  \subfigure[3-Client]{
    \includegraphics[scale=0.27]{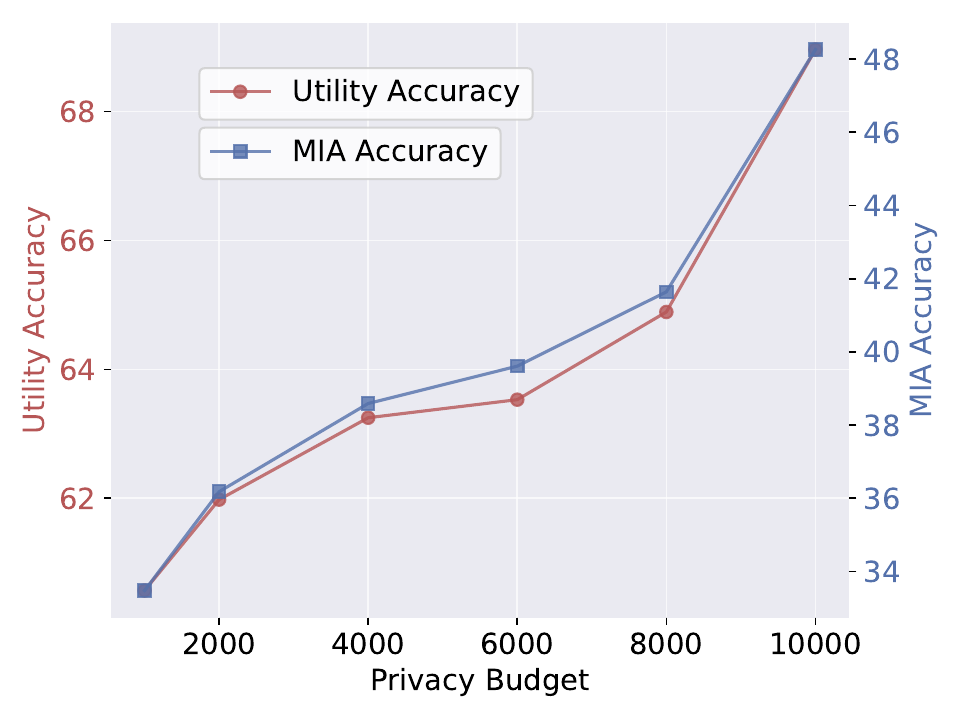}
  }
  \subfigure[4-Client]{
    \includegraphics[scale=0.27]{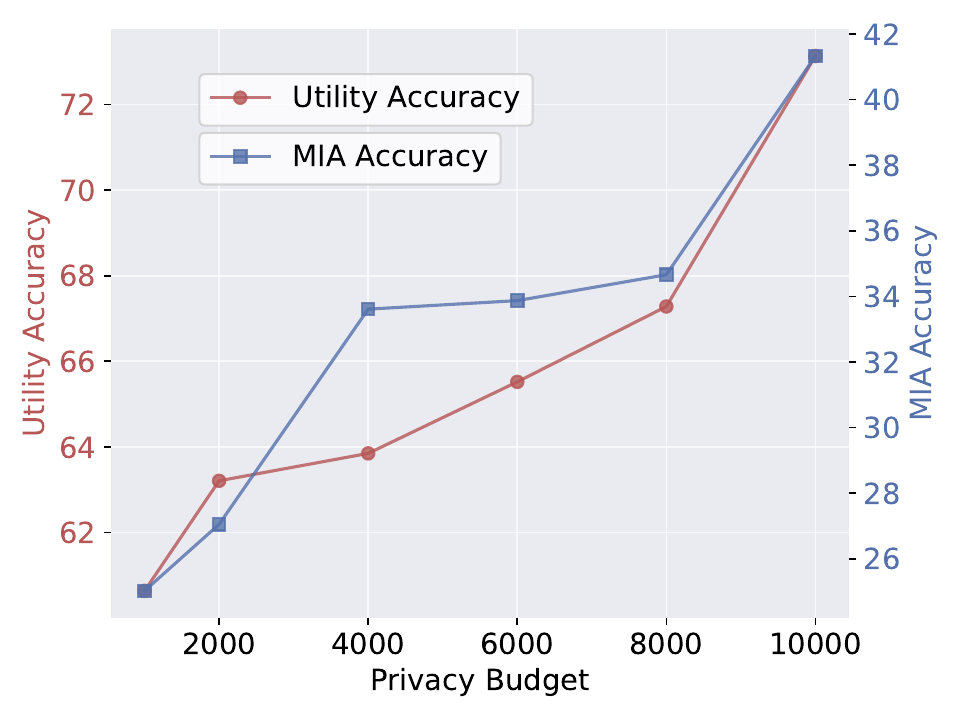}
  }
  \subfigure[5-Client]{
    \includegraphics[scale=0.27]{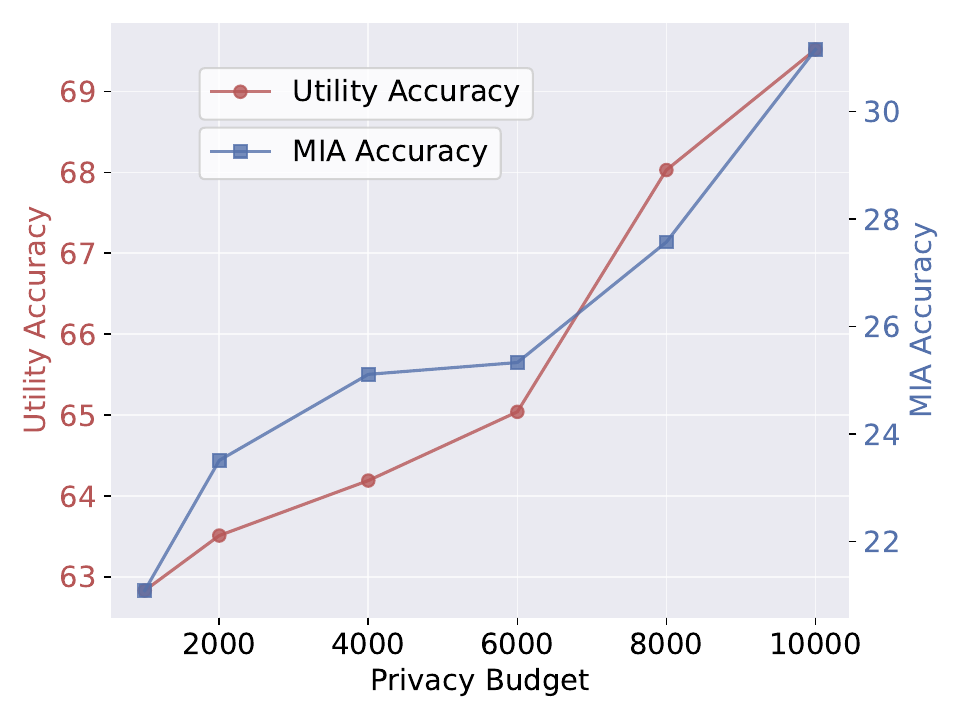}
  }
  \subfigure[6-Client]{
    \includegraphics[scale=0.27]{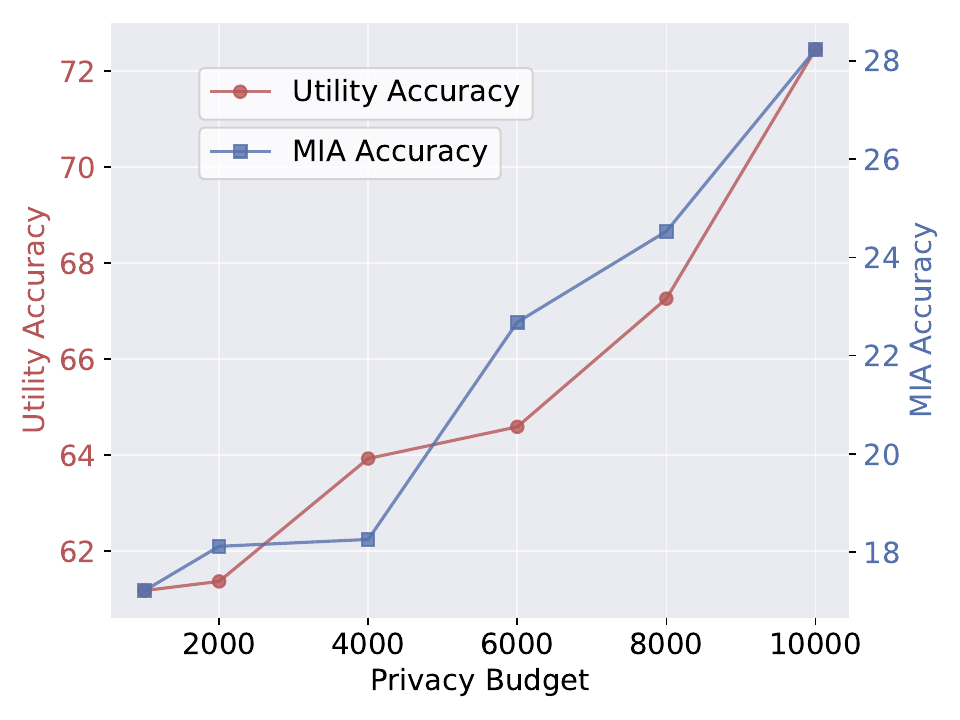}
  }
  \subfigure[7-Client]{
    \includegraphics[scale=0.27]{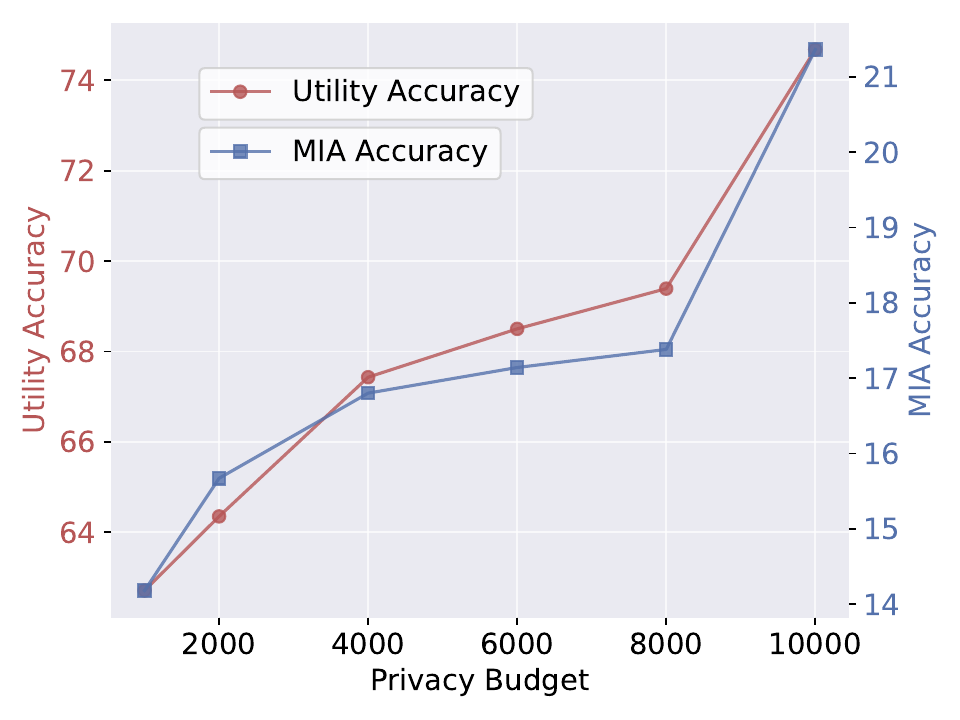}
  }
  \subfigure[8-Client]{
    \includegraphics[scale=0.27]{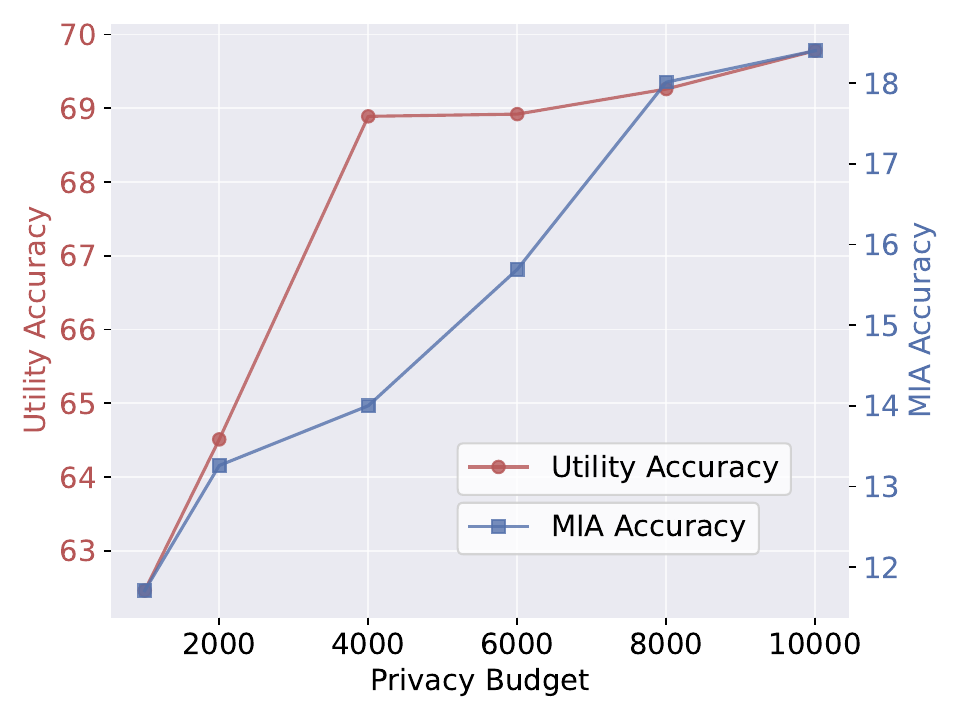}
  }
  \caption{The potential defense against \ours for client-data identification.}
  \label{defense_client_ownership}
\end{figure}

The trade-off of defense is illustrated in Fig~\ref{defense_training_set} and Fig~\ref{defense_client_ownership}.

Applying $d_{\chi}$-based~\cite{2013dx} perturbations with varying intensities results in a noticeable degradation in the performance of the global GNN. Under the training-set inference scenario, even when the model's performance is severely compromised, \ours remains highly effective in conducting inference attacks. For client-data identification, the attack success rate can only be reduced to near-random levels when the privacy budget is extremely small (\textit{i.e.}, the noise intensity is very high). However, this comes at the unavoidable cost of significant damage to the overall utility and performance of the GNN.

\subsection{Detailed Related Works}
\label{detailed_related_works}
\subsubsection{Membership Inference in Federated Learning}

Despite its privacy-preserving design, FL remains vulnerable to various inference attacks. Prior studies have shown that adversaries can exploit shared model updates to reconstruct sensitive training data \cite{zhu2019deep}, infer statistical properties of other clients' datasets \cite{melis2019exploiting}, or even synthesize representative inputs \cite{wang2019beyond}. Among these, membership inference attacks (MIAs) pose a fundamental privacy threat by determining whether a specific data sample was used in training. MIAs have significant implications: they can lead to privacy breaches (\textit{e.g.}, revealing a patient’s diagnosis by confirming their data was used in a medical model), support compliance auditing (\textit{e.g.}, verifying data deletion under “right to be forgotten” laws), and serve as precursors to more advanced threats such as model extraction. A growing body of work has explored MIAs in the FL setting, beginning with gradient-based attacks that leverage leakage from updates, hidden layer activations, and loss signals \cite{nasr2019comprehensive}. Recently, new MIAs have focused on methods of stealing hyperparameters \cite{li2022auditing}, shadow training \cite{zhang2022label}, learning logits distribution \cite{yan2022membership}, and feature construction \cite{liu2023subject}, \textit{etc}. These attacks have since been extended to various domains, including classification, regression, and recommendation, using both shared model parameters and trends in model outputs over training rounds.

\subsubsection{Membership Inference in Graph Neural Networks}

Membership inference attacks (MIAs) have recently been extended to graph neural networks (GNNs), with early efforts adapting the shadow training framework from traditional domains to node classification tasks \cite{olatunji2021membership}. One line of work introduced 0-hop, 2-hop, and combined attacks that exploit both node and neighbor posterior distributions to improve inference accuracy \cite{he2021node}. Label-only attacks have also been explored, relying solely on predicted labels without access to confidence scores \cite{conti2022label}. Beyond node-level inference, graph-level MIAs have been proposed using training-based and threshold-based methods to determine whether an entire graph instance was used during training \cite{wu2021adapting}. Further studies have examined MIAs under adversarial training settings \cite{liu2022membership}, as well as subgraph-level attacks that construct discriminative features to distinguish target subgraphs \cite{zhang2022inference}.

\end{document}